\documentclass{article}

\usepackage{microtype}
\usepackage{graphicx}
\usepackage{subfigure}
\usepackage{booktabs} 
\usepackage{xcolor}

\usepackage{hyperref}

\usepackage[accepted]{icml2024}

\usepackage{amsmath}
\usepackage{amssymb}
\usepackage{mathtools}
\usepackage{amsthm}

\usepackage[capitalize,noabbrev]{cleveref}

\theoremstyle{plain}

\theoremstyle{definition}

\theoremstyle{remark}

\usepackage[textsize=tiny]{todonotes}

\icmltitlerunning{Directions of Curvature as an Explanation for Loss of Plasticity}

\begin{document}

\twocolumn[

\icmltitle{Directions of Curvature as an Explanation for Loss of Plasticity}




\begin{icmlauthorlist}
\icmlauthor{Alex Lewandowski}{yyy}
\icmlauthor{Haruto Tanaka}{yyy}
\icmlauthor{Dale Schuurmans}{yyy,comp,cifar}
\icmlauthor{Marlos C. Machado}{yyy,cifar}
\end{icmlauthorlist}

\icmlaffiliation{yyy}{Department of Computing Science, University of Alberta, Edmonton, Canada}
\icmlaffiliation{comp}{Google DeepMind}
\icmlaffiliation{cifar}{Canada CIFAR AI Chair}

\icmlcorrespondingauthor{Alex Lewandowski}{lewandowski@ualberta.ca}

\icmlkeywords{Machine Learning, ICML}

\vskip 0.3in
]



\printAffiliationsAndNotice{} 

\begin{abstract}
  Loss of plasticity is a phenomenon in which neural networks lose their ability to learn from new experience.
  Despite being empirically observed in several problem settings, little is understood about the mechanisms that lead to loss of plasticity.
  In this paper, we offer a consistent explanation for loss of plasticity: Neural networks lose directions of curvature during training and that loss of plasticity can be attributed to this reduction in curvature.
  To support such a claim, we provide a systematic investigation of loss of plasticity across continual learning tasks using MNIST, CIFAR-10 and ImageNet.
  Our findings illustrate that loss of curvature directions coincides with loss of plasticity, while also showing that previous explanations are insufficient to explain loss of plasticity in all settings.
  Lastly, we show that regularizers which mitigate loss of plasticity also preserve curvature, motivating a simple distributional regularizer that proves to be effective across the problem settings we considered. \looseness=-1
\end{abstract}

\section{Introduction}

The goal of continual learning research is to develop algorithms that can
continue to learn from a dynamic data distribution
\citep{ring94_contin,thrun98_lifel}.
While current machine learning algorithms
are capable of learning from a fixed dataset and generalizing to unseen data
from the same distribution, these algorithms can struggle to adapt to changes in
the distribution over the course of learning
\citep{zilly21,abbas23_loss_plast_contin_deep_reinf_learn,lyle23_under,dohare23_maint_plast_deep_contin_learn}.
The ability to learn from new data, also referred to as plasticity, is one way
in which algorithms can adapt to such changes. It has recently been noted that
loss of plasticity---a reduced ability to learn new things
\citep{dohare23_maint_plast_deep_contin_learn,lyle23_under}---is a critical
shortcoming of neural network learning algorithms in continual learning
settings.
In particular, neural network algorithms
that are trained on a changing data distribution can lead to progressively slower learning, or completely stop
learning altogether.

Identifying the mechanisms behind neural network plasticity is an active area of research, with several components
potentially contributing to improving plasticity, or otherwise mitigating loss of plasticity.
For example, the use of stateful optimizers have been found to exacerbate loss of plasticity due to inaccurate gradient estimates after a data distribution change \citep{dohare23_maint_plast_deep_contin_learn, lyle23_under}.
Tuning other properties of the optimization process, such as the step-size \citep{ash20_warm_start_neural_networ_train,berariu2021study} and the number of updates \citep{lyle23_under}, have been found to mitigate loss of plasticity.
Saturating activation functions can also lead to loss of plasticity by limiting capacity \citep{sokar23_dorman_neuron_phenom_deep_reinf_learn}, which can be mitigated by
non-saturating activation functions \citep{abbas23_loss_plast_contin_deep_reinf_learn}.
While a full mechanistic understanding of plasticity has not yet been identified,
these experimental results suggest that optimization dynamics play an important role in sustaining neural network properties needed for plasticity.
Some properties that have been found to correlate with loss of plasticity include, a decrease in the gradient or update norm \citep{abbas23_loss_plast_contin_deep_reinf_learn},
neuron dormancy \citep{sokar23_dorman_neuron_phenom_deep_reinf_learn}, and an increase in the norm of the parameters \citep{nikishin22_primac_bias_deep_reinf_learn}.
Unfortunately, these properties do not explain loss of plasticity in all the situations that it occurs.
\looseness=-1

In this paper, we propose that loss of plasticity can be explained by a loss of curvature directions.
Our work contributes to a growing literature on the importance of curvature for understanding neural network dynamics \citep{cohen2021gradient,hochreiter97_flat,fort19_emerg}.
Within the continual learning and plasticity literature, the assertion that curvature is related to plasticity is relatively new \citep{lyle23_under}.
In contrast to the general assertion that curvature is related to plasticity, our work specifically posits that loss of curvature directions explains loss of plasticity.
In particular, we provide empirical evidence that supports the claim that
loss of plasticity co-occurs with a reduction in the rank of the Hessian of the training objective at the beginning of a new task.

More specifically, this work improves the understanding of loss of plasticity in continual supervised learning by:

\begin{enumerate}
  \item Surveying previous explanations for loss of plasticity. We provide counterexamples showing 
   that existing explanations are not consistent, that is, they do not explain loss of plasticity in all situations it occurs.

  \item Proposing that loss of curvature directions, measured as the reduction in the rank of the Hessian of the training objective, is a consistent explanation for loss of plasticity. We demonstrate that loss of curvature directions coincides with loss of plasticity across all factors and benchmarks that we consider.

  \item Introducing a Wasserstein regularizer
        that keeps the distribution of weights close to the initialization distribution. This Wasserstein regularizer allows the parameters to move further from initialization while preserving curvature for successive tasks.
  Learning with the Wasserstein regularizer requires fewer iterations and achieves a lower error compared to other regularizers.
\end{enumerate}

\section{Factors and Explanations for Loss of Plasticity}
\label{sec:fac}

Before defining what we mean by loss of plasticity, we outline the continual supervised learning problem setting we study.
We assume the learning algorithm operates in a minibatch setting, processing $M$ observation-target pairs, $\{x_{i},y_{i}\}_{i=1}^{M}$, and updating the neural network parameters, $\theta$, after each minibatch.
In continual supervised learning, there is a periodic and regular change every $U$ updates to the distribution generating the observations or targets.
For every $U$ updates, the neural network must minimize an objective defined over a new distribution---we refer to this new distribution as a \textit{task}.
The problem setting is designed so that the task at any point in time has the same difficulty.\footnote{A suitably initialized neural network should be able to equally minimize the objective for any of the tasks we consider.}
We are primarily interested in the error at the the end of task $K$ averaged across all observations in that task, $J_{K} = J(\theta_{UK}) = \mathbb{E}_{p_{K}}\big[\ell(f_{\theta_{UK}}(x), y)\big]$, for some loss function $\ell$, and task specific data distribution $p_{K}$.

Although loss of plasticity is an empirically observed phenomenon, the way it is measured in the literature can vary.
In this paper, we use loss of plasticity to refer to the phenomenon that $J_{K}$ increases rather than decreases as a function of $K$.
Some works evaluate learning and plasticity with the average online error over the learning trajectory within a task \citep[e.g.,][]{elsayed23_utilit_pertur_gradien_descen,dohare23_maint_plast_deep_contin_learn,kumar23_maint_plast_regen_regul}.
While the two are related, we focus on the error at the end of the task to remove the effect of the unavoidable error increase at the beginning of a subsequent task. If we were to consider the large initial error, we might infer loss of plasticity in the average online error even if the error at the end of a task is constant (see Appendix~\ref{appendix:online_error}).
Because the error at the end of a task increases as more tasks are seen, this means that the neural network is struggling to learn from the new experience given by the subsequent task.

\subsection{Factors That Can Contribute to Loss of Plasticity}
\label{sec:factors}

Given a concrete notion of plasticity,
we reiterate that the underlying mechanisms leading to loss of plasticity have been so-far elusive.
This is partly because multiple factors can potentially contribute to, or mitigate, loss of plasticity.
In this section, we summarize some of these potential factors before surveying previous explanations for the underlying mechanism behind loss of plasticity.

  \textbf{Optimizer}\hspace{2mm}
        Optimizers that were designed and tuned for stationary distributions can exacerbate loss of plasticity in non-stationary settings.
        For instance, the work by \citet{lyle23_under} showed empirically that Adam \citep{kingma14_adam} can be unstable on a subsequent task due to its momentum and scaling from a previous task.

  \textbf{Step-size}\hspace{2mm}
        In addition to the optimizer,
        the step-size is a crucial factor in both contributing to and mitigating loss of plasticity.
        The study by \citet{berariu2021study}, for example, suggests that loss of plasticity is preventable by amplifying the randomness of gradients with a larger step-size.
        These findings extend to other hyper-parameters of the optimizer. Properly tuned hyper-parameters for Adam, for example, can mitigate loss of plasticity which leads to policy collapse in reinforcement learning \citep{dohare23_overc_polic_collap_deep_reinf_learn,lyle23_under}. \looseness=-1

  \textbf{Update budget}\hspace{2mm}
      Continual supervised learning experiments, including those below, use a fixed number of update steps per task \citep[e.g.,][]{abbas23_loss_plast_contin_deep_reinf_learn, elsayed23_utilit_pertur_gradien_descen, javed2019meta}.
      Despite the fact that the individual tasks themselves are of the same difficulty, the neural network might not be able to escape its task-specific initialization within the pre-determined update budget.
        \citet{lyle23_under} show that, as the number of update steps increase in a first task, learning slows down on a subsequent task, requiring even more update steps on the subsequent task to reach the same training error.

  \textbf{Activation function}\hspace{2mm}
        One major factor that can contribute or mitigate loss of plasticity is the activation function.
        Work by \citet{abbas23_loss_plast_contin_deep_reinf_learn} suggests that, in the reinforcement learning setting, loss of plasticity occurs because of an increasing portion of hidden units being set to zero by \texttt{ReLU} activations \citep{fukushima75_cognit,nair10_rectif_linear_units_improv_restr_boltz_machin}.
        The authors then show that \texttt{CReLU} \citep{shang16_under} prevents saturation, mitigating loss of plasticity almost entirely.
        However, other works have shown that loss of plasticity can still occur with non-saturating activation functions \citep{dohare21_contin_backp, dohare23_maint_plast_deep_contin_learn} such as \texttt{leaky-ReLU} \citep{xu15_empir}.

  \textbf{Properties of the objective function and the regularizer}\hspace{2mm}
        The objective function being optimized greatly influences the optimization landscape and, hence, plasticity \citep{ lyle21_under_preven_capac_loss_reinf_learn,lyle23_under,ziyin23_symmet_leads_struc_const_learn}.
        Regularization is one modification to the objective function that helps mitigate loss of plasticity.
        For example, when weight decay is properly tuned, it can help mitigate loss of plasticity \citep{dohare23_maint_plast_deep_contin_learn}.
        Another regularizer that mitigates loss of plasticity is regenerative regularization, which regularizes towards the parameter initialization \citep{kumar23_maint_plast_regen_regul}.

\subsection{Previous Explanations for Loss of Plasticity}

Not only are there several factors that could possibly contribute to loss of plasticity, there are also several explanations for this phenomenon. We survey the recent explanations of loss of plasticity below. In the next section, we present results showing that none of these explanations are sufficient to explain loss of plasticity across all problem settings we consider.

  \textbf{Decreasing update/gradient norm}\hspace{2mm} The simplest explanation for loss of plasticity is that the update norm goes to zero. This would mean that the parameters of the neural network stop changing, eliminating all plasticity. This tends to occur with a decrease in the norm of the features for particular layers \citep{abbas23_loss_plast_contin_deep_reinf_learn,nikishin22_primac_bias_deep_reinf_learn}.

  \textbf{Dormant Neurons}\hspace{2mm} Another explanation for loss of plasticity is a steady decrease in the proportion of active neurons, namely, the dormant neuron phenomenon \citep{sokar23_dorman_neuron_phenom_deep_reinf_learn}.
  It is hypothesized that a decrease in the number of active neurons also decreases a neural network's expressivity, potentially leading to loss of plasticity.

   \textbf{Decreasing representation rank}\hspace{2mm} Related to the effective capacity of a neural network, lower representation rank suggests that fewer features are being represented by the neural network \citep{kumar20_implic_under_param_inhib_data}.
  It has been observed that decreasing representation rank is sometimes correlated with loss of plasticity \citep{lyle23_under,kumar23_maint_plast_regen_regul,dohare23_maint_plast_deep_contin_learn}.

  \textbf{Increasing parameter norm}\hspace{2mm} An increasing parameter norm is sometimes associated with loss of plasticity in both continual supervised and continual reinforcement learning \citep{nikishin22_primac_bias_deep_reinf_learn,dohare23_maint_plast_deep_contin_learn}, but it is not necessarily a cause \citep{lyle23_under}. It is not clear why the parameter norms increase and lead to loss of plasticity, perhaps suggesting a slow divergence in the training dynamics. \looseness=-1

\section{Counterexamples for Previous Explanations}
\label{sec:counter}
In this section, we investigate the explanations for loss of plasticity described in Section~\ref{sec:fac} and we provide counterexamples for them, showing that they fail to fully explain loss of plasticity. To do so, we use a linearly separable subset of the MNIST dataset \citep{lecun10_mnist}, in which the labels of each image are periodically shuffled.
While MNIST is a simple classification problem, label shuffling highlights the difficulties associated with preserving plasticity \citep[see][]{lyle23_under,kumar23_maint_plast_regen_regul}.
We focus on this problem for its simplicity, showing that even in a setting where linear function approximation is sufficient, one can find counterexamples to the previous explanations in the literature for loss of plasticity.
We emphasize that the goal here is merely to uncover simple counterexamples that refute proposed explanations for loss of plasticity, not to investigate the phenomenon more broadly.
In Section~\ref{sec:main_exp}, we extend our investigation of loss of plasticity to larger scale benchmarks.
\looseness=-1

\paragraph{Methods}
In this experiment, we vary only the activation function between \texttt{ReLU}, \texttt{leaky-ReLU}, \texttt{tanh} and the \texttt{identity}.
As noted in Section~\ref{sec:factors}, previous work has found that the activation function has a significant effect on the plasticity of the neural network.
We measure the error across all observations at the end of each task.
Each task lasts 200 epochs, which is sufficient for neural networks with any of the considered activation functions to achieve low error on the first few tasks using a random initialization.

\begin{figure}[t]
  \centering
\includegraphics[width=1.0\linewidth]{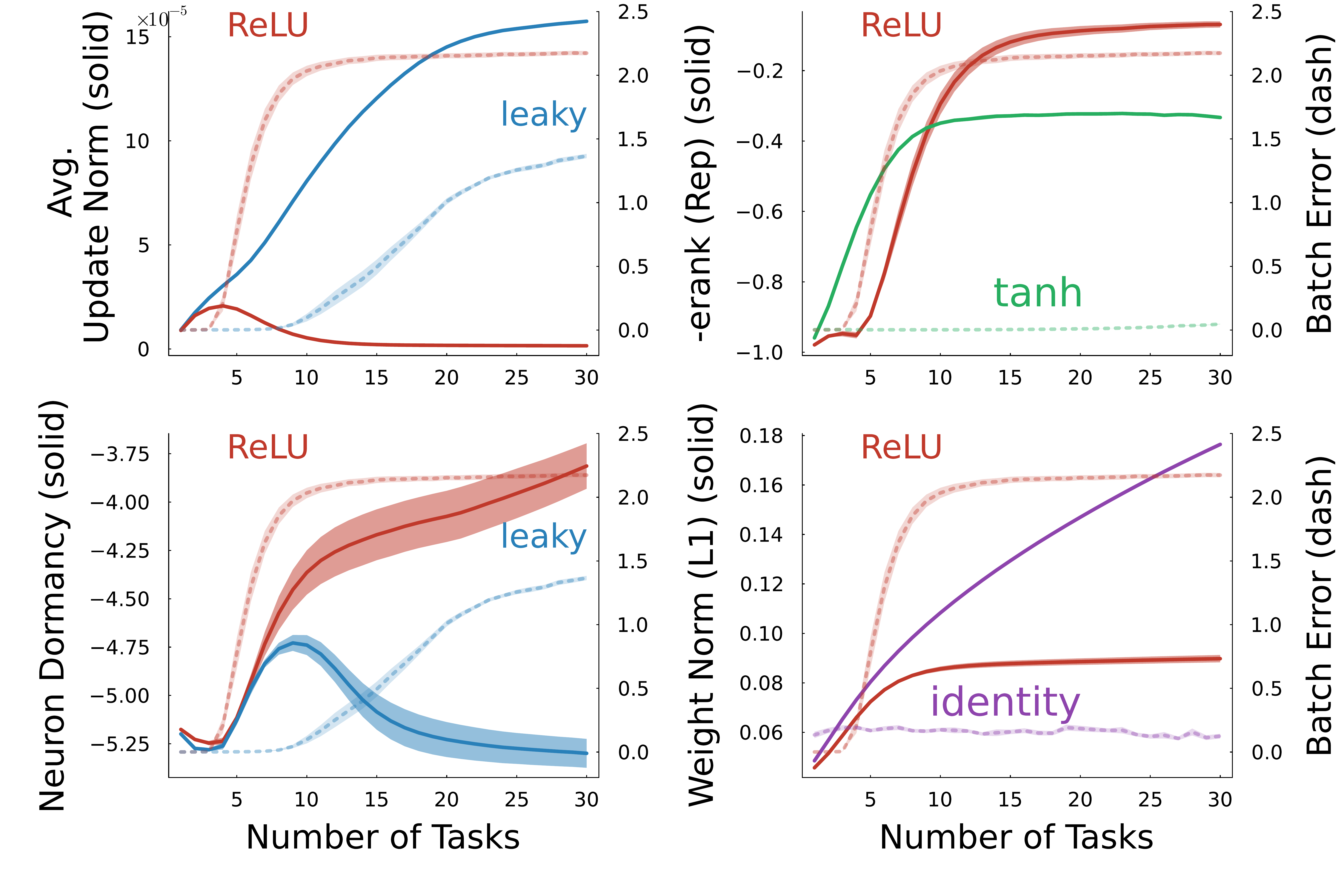}
  \caption{\textbf{Inconsistencies of previous explanations for loss of plasticity on Random Label MNIST (subset)}. The explanations on the left are not consistent because both \texttt{ReLU} and \texttt{leaky-ReLU} suffer from loss of plasticity. On the right, there is no loss of plasticity for \texttt{tanh} and \texttt{identity} but the corresponding explanations predict that they do. All results have a shaded region corresponding to a 95\% confidence interval of the mean over 30 runs.
    \looseness=-1
  }
  \label{fig:confounding}
\end{figure}
\paragraph{Results}
The main result of this experiment can be found in Figure~\ref{fig:confounding}.
Our findings show that none of the aforementioned explanations of loss of plasticity explain the phenomenon.
All non-linear activation functions can achieve low error on the first few tasks, but
for \texttt{ReLU} and \texttt{leaky-ReLU}, the error increases and eventually becomes worse than the neural network with identity activation (which is incapable of feature learning).\footnote{While the neural network with \texttt{tanh} activations does not lose plasticity in this experiment, in Section~\ref{sec:main_exp} we show that it does lose plasticity when we consider the full MNIST dataset.}
Despite some non-linear activation functions losing plasticity, the explanations on the left side of Figure~\ref{fig:confounding} fail to predict loss of plasticity consistently.
A decreasing update norm, for example, may seem like an intuitive explanation of loss of plasticity.
However, in the top-left plot, we see that the update norm consistently increases for the \texttt{leaky-ReLU} activation function, making the explanation inconsistent.
For the right side of Figure~\ref{fig:confounding}, the corresponding explanation predicts loss of plasticity for \texttt{tanh} and \texttt{identity} but we see it does not actually occur.
The rank of the representation (plotted as a negative for uniformity with other explanations), another popular candidate explanation, decreases for the \texttt{tanh} activation despite no loss of plasticity in this problem. \looseness=-1

\begin{figure}[t]
  \centering
\includegraphics[width=0.99\linewidth]{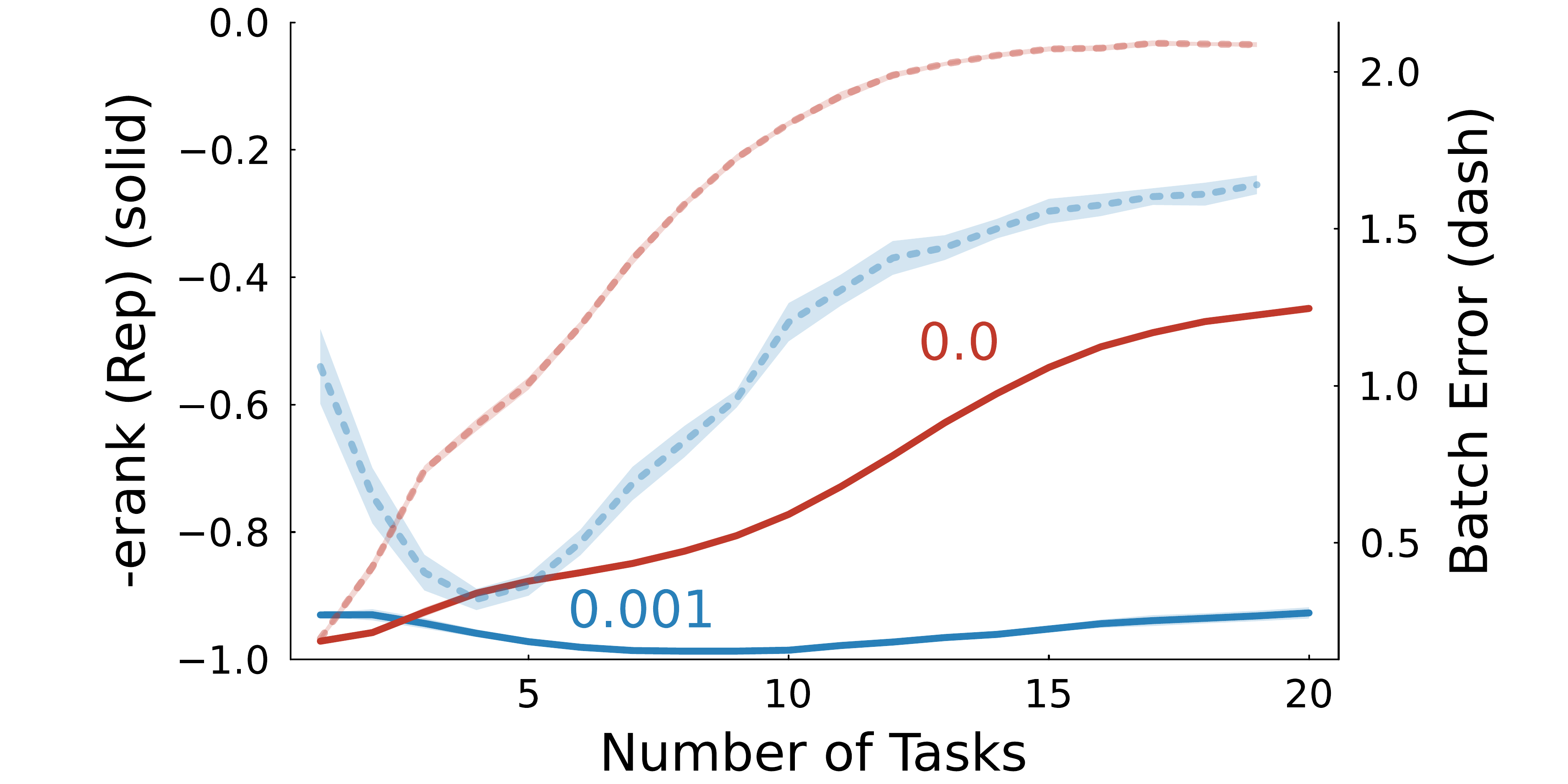}
  \caption{\textbf{Effect of feature rank regularization is maintaining plasticity.} Loss of plasticity still occurs with \texttt{leaky-ReLU} and feature rank regularization, despite the fact that the feature rank remains high. All results have a shaded region corresponding to a 95\% confidence interval of the mean over 30 runs.}
  \label{fig:confounding_featurereg}
\end{figure}

Because feature rank is such a predominant explanation for loss of plasticity, we provide an additional counter-example showing that the feature rank is also not a sufficient explanation; rather, it is a symptom of a deeper problem. We re-run the previous experiment using a regularizer, $J_{\text{feature-reg}}(\Phi) = \sigma_{1}^{2}(\Phi) - \sigma_{d}^{2}(\Phi)$, that encourages the feature representation to be full rank \citep{kumar20_implic_under_param_inhib_data}. The results, in Figure~\ref{fig:confounding_featurereg}, show that regularization increases the feature rank, but that this is not sufficient to prevent loss of plasticity. For example, take the rank of the feature representation between tasks 5 and 10; although it increases in that period, the error increases, which means plasticity is still being lost.
\looseness=-1

\paragraph{Summary}
The previous explanations are not consistent because there exists at least one activation such that the trend in the training error does not agree with the trend in the explanation (see Appendix~\ref{sec:appendix_analysis} for additional analysis).
A maybe surprising finding is that the deep linear network (a neural network with an \texttt{identity} activation function) is able to maintain a low training error across all tasks for this problem.
A deep linear network has more parameters than a linear function, but it can only represent linear functions.
This is sufficient to solve each task because the number of data points ($1280$) is smaller than the effective dimensionality of the network ($d_{in}\times d_{out} = 7840$).
The deep linear network's ability to preserve plasticity is surprising because the training dynamics of a deep linear network are non-linear and similar to a deep non-linear network \citep{saxe14_exact}.
The fact that loss of plasticity only occurs with non-linear activations suggests that the curvature introduced by the non-linearities is crucial in explaining loss of plasticity. \looseness=-1

\color{black}

\section{Measuring the Curvature of a Changing Optimization Landscape}

A missing piece in the previously proposed explanations is the curvature of the optimization landscape.
While previous work pointed out that curvature is connected to plasticity \citep{lyle23_under}, our work specifically posits that a reduction in the number of curvature directions coincides with loss of plasticity.
In Section~\ref{sec:main_exp} we show that loss of plasticity occurs when, at the start of a new task, the optimization landscape has a diminishing number of curvature directions.

The optimization landscape in continual learning is not easy to characterize because it can change without the parameters changing.
Unlike supervised learning, where the data distribution is stationary, the data distribution underlying the observations and targets will change in the continual learning setting.
Thus there can be changes in the objective, gradient and Hessian that is due to the data changing and not due to parameter changes.

Before presenting empirical evidence of the relationship between plasticity and curvature, we note that there are several notions of curvature in the literature.
The local curvature of the optimization landscape at a particular parameter $\theta$ is expressed by the Hessian of the objective function, $H_{t}(\theta) = \nabla_{\theta}^{2}J_{t}(\theta)\big|_{\theta = \theta_{t}}$.\footnote{
We omit the dependence on data in the training objective and the Hessian, instead indexing both by time.}
Different measures of curvature correspond to different functions of this Hessian matrix.
One common measure of curvature is the sharpness, given by the maximum eigenvalue of the Hessian \citep{kesker17_generalization_gap_and_sharp_minima,cohen2021gradient}.
Sharpness is coarse-grained, it only gives the magnitude of the vector of maximal curvature and it fails to characterize other directions.
Another measure, and the one that this paper investigates, is the effective rank of the Hessian matrix, which counts the effective number of directions of curvature.

We are interested in how the curvature of the optimization landscape changes when the task changes.
Of particular interest is the rank of the Hessian after a task change. If it is decreasing, then there are fewer directions of curvature to explore the parameter space and to learn on the new task.
For simplicity, and in alignment with our experiments, we assume that each task has an update budget of $U$ iterations.
Thus, the training objective on the $K$-th task is stationary for $U$ steps.
When the task changes, at $t = UK + 1$, the Hessian changes due to changes in the data---and not due to changes in the parameters.
We measure the rank at the beginning of the task by the \emph{effective rank}, $\texttt{erank}\left(H_{UK+1}(\theta)\right)$, where $\texttt{erank}(M) = \min \left\{ j \, : \, \frac{ \sum_{i=1}^{j}\sigma_{i}(M)}{\sum_{i=1}^{d}\sigma_{i}(M)} > 0.99\right\}$ is the effective rank and $\{\sigma_{i}(M)\}_{i=1}^{d}$ are the singular values arranged in decreasing order.
The effective rank specifies the number of basis vectors needed to represent 99\% of image of the matrix $M$ \citep{yang19_harnes_struc_value_based_plann_reinf_learn,kumar20_implic_under_param_inhib_data}.

\subsection{Approximating the Hessian Rank}

Neural networks typically have a large number of parameters, requiring approximations to the Hessian due to the massive computational overhead for producing the matrix.
Diagonal approximations are employed to capture curvature information relevant for optimization \citep{elsayed22_hessc,lecun89_optim}, but are full rank unless the parameter gradients become zero, which typically does not occur in classification.
There are low-rank approximations of the Hessian \citep{roux07_topmoum}, these too are problematic for our analysis because we aim to measure the rank of the Hessian and cannot presuppose that it is low-rank.
Lastly, stochastic Lanzcos methods are able to efficiently approximate the smallest and largest eigenvalues \cite{ghorbani19_inves_neural_net_optim_hessian_eigen_densit}, but they cannot efficiently estimate the middle bulk of eigenvalues which can determine the rank.
\looseness=-1

To approximate the Hessian rank, we use the an outer-product approximation of $m$ per-sample gradients, $\mathbf{H} \approx \hat{\mathbf{H}} = \sum_{i}^{m} g_{i} g_{i}^{\intercal}$, where  $g_{i} = \nabla_{\theta}J(\theta, x_{i}, y_{i})$ is the gradient with respect to a single datapoint $(x_{i}, y_{i})$. This approximation is useful for estimating the rank because if $v$ is in the nullspace of the Hessian, $\hat{\mathbf{H}}v = 0$, then it is a direction of zero curvature and orthogonal to the per-sample gradients, $g_{i}^{\intercal} v = 0$. Thus, the vector is in the nullspace of the outer-product approximation and $\hat{\mathbf{H}}v = 0$. Of course, $\texttt{rank}(\hat{\mathbf{H}}) \leq M$ and $M << d$ means that the approximation will underestimate the rank. Our interest is in the relative decrease in the rank. We will report the effective rank divided by the maximum rank because the exact number of curvature directions is not relevant for our results.

The outer-product approximation also avoids the computational demands of the singular value decomposition needed to compute the effective rank.
First, we rewrite the approximation $\hat{\mathbf{H}} = \sum_{i=1}^{M}g_{i}g_{i}^{\intercal} = \mathbf{G} \mathbf{G}^{\intercal}$, where $\mathbf{G} = [g_{1}, \dotso, g_{m}] \in \mathbb{R}^{d \times M}$ is the matrix of per-sample gradients. Then, because $\hat{\mathbf{H}}$ is a Gram matrix, we have that $\texttt{rank}( \mathbf{G}\mathbf{G}^{\intercal}) =  \texttt{rank}( \mathbf{G}^{\intercal} \mathbf{G})$. This is useful because $\mathbf{G}^{\intercal} \mathbf{G} \in \mathbb{R}^{M \times M}$ and $M$ is much smaller than $d$.

Another name for this approximation is
the empirical Fisher information matrix, and it has been argued that it should not be used as a replacement for the Hessian as a pre-conditioner in second-order optimization because it is not guaranteed to capture the curvature information of the Hessian \citep{kunstner19_limit_fisher}.
Recent work studying neural network generalization, however, argues that the inner product of the per-example gradients can be useful in understanding neural network generalization and learning dynamics \citep{fort19_stiff,lyle22_learn_dynam_gener_deep_reinf_learn}.
The matrix of gradient inner products, equivalently $\mathbf{G}^{\intercal} \mathbf{G}$, was also used to assess gradient covariance in continual learning \citep{lyle23_under}.
Thus, the relative rank of the gradient outer-products provides a reasonable approximation to the relative rank of the Hessian, which we demonstrate empirically in the next section.

\subsection{Validating the Hessian Rank Approximation}

We evaluate the approximation to the Hessian rank in a simple problem where we can efficiently calculate the full Hessian and its rank. The problem is similar to the experiments in Section~\ref{sec:counter}, except we also apply a stochastic projection matrix to the MNIST images to reduce the input dimension and overall parameter count.

\begin{figure}[t]
  \centering
  \includegraphics[width=0.99\linewidth]{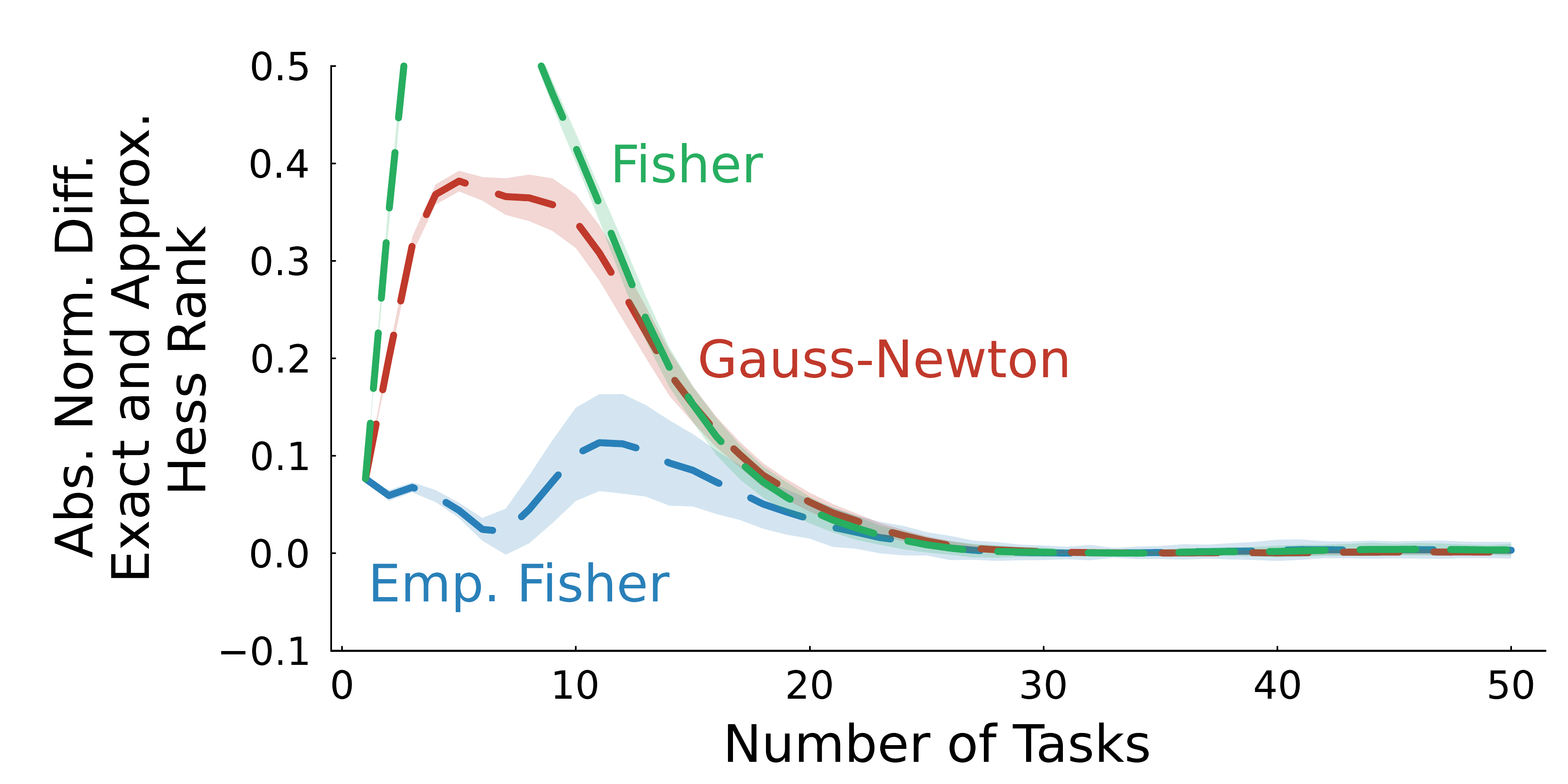}
  \caption{\textbf{Comparison between different methods for approximating the Hessian rank.} The empirical Fisher approximation to the Hessian rank is highly accurate in the first few tasks, which is when loss of plasticity occurs. When plasticity worsens in later tasks, the approximation quality marginally worsens. Overall, the empirical Fisher is an accurate and efficient approximation to the Hessian rank.
\looseness=-1
  }
  \label{fig:hessian_approx}
\end{figure}

We compare the approximation quality of the Hessian rank using three different methods: 1)~Empirical Fisher (our approach), 2)~Fisher, and, 3)~Gauss-Newton.
We measure the rank of the exact Hessian and the rank of the Hessian approximation at the beginning of each new task. Next, we normalize each rank by its corresponding maximum possible rank.
To measure the approximation quality, we plot the absolute difference between the relative effective Hessian ranks.
Our results in Figure~\ref{fig:hessian_approx} show that the proposed empirical Fisher approximation to the Hessian rank is particularly accurate in estimating the rank in the first few tasks, which is when loss of plasticity occurs.
As plasticity degrades in later tasks, the approximation quality worsens but still accurately represents the overall trend of the true Hessian rank. 

Comparisons for other neural networks, further details, and figures demonstrating the dynamics of the Hessian approximation can be found in Appendix~\ref{appendix:hessian_approx}.
We use this Hessian rank approximation to explain loss of plasticity in continual supervised learning in the rest of our experiments.

\section{Preserving Curvature with Regularization}

In the previous section, we claimed that loss of curvature may explain loss of plasticity.
Regularization is commonly used to improving the conditioning of matrices \citep{benning18_moder_regul_method_inver_probl}.
This does not immediately imply that regularization preserves plasticity because we are interested in minimizing the unregularized objective, and preserving the rank of the Hessian with respect to the unregularized objective.
Our central claim is that regularization also preserves the rank of the unregularized Hessian, and allows neural networks to preserve plasticity.\footnote{All measurements of the Hessian rank are with respect to the unregularized objective.}

If curvature is lost over the course of learning, then one solution to this problem is to regularize towards the curvature present at initialization.
While explicit Hessian regularization would be computationally costly, previous work has found that even weight decay can mitigate loss of plasticity \citep{dohare21_contin_backp,lyle21_under_preven_capac_loss_reinf_learn,kumar23_maint_plast_regen_regul}, without attributing this benefit to preserving directions of curvature.
These methods, however, do more than just prevent loss of curvature, they also prevent parameters from growing large (subject to the regularization parameter's strength).
Weight decay, for example, mitigate loss of plasticity but also prevent the parameters from deviating far from the origin.
The restriction that weight decay imposes on the update requires careful tuning of the regularization strength as we show in Section~\ref{sec:main_exp} and Appendix~\ref{appendix:reg_hyperparam}.

We propose a new regularizer that is simple and that gives the parameters more leeway for moving from the initialization, while preserving the desirable plasticity and curvature properties of the initialization.
Our regularizer penalizes the distribution of parameters if it is far from the distribution of the randomly initialized parameters.
At initialization, the parameters at layer $l$ are sampled i.i.d. $\mathbf{\theta}_{i,j} \sim p^{(l,0)}(\theta)$ according to some pre-determined distribution, such as the Glorot initialization \citep{glorot10_under}.
 The distribution of parameters at iteration $t$ during training and for any particular layer, denoted by $p^{(l,t)}$, is no longer known (the parameters may not be independent nor identically distributed).
However, it is still possible to regularize the empirical distribution towards the initialization distribution by using the empirical Wasserstein metric \citep{bobkov19_one_kantor}.
We denote the flattened parameter matrix for layer $l$ at time $t$ by $\mathbf{\bar{\theta}}^{(l,t)}$. The squared Wasserstein-2 distance between the distribution of parameters at initialization and the current parameter distribution is defined as,
$$ \mathcal{W}_2^2\left(p^{(l,0)},p^{(l,t)}\right) = \sum_{i=1}^{d} \left(\mathbf{\bar{\theta}}_{(i)}^{(l,t)} - \mathbf{\bar{\theta}}_{(i)}^{(l,0)}\right)^{2}.$$
The order statistics of the parameter is denoted by  $\theta_{(i)}^{(l,t)}$ and represents the $i$-th smallest parameter at time $t$ for layer $l$.
In the above equation, we are taking the L2 difference between the order statistics of each layer's parameters at initialization and at iteration $t$ during training.
The \textit{Wasserstein regularizer} uses the empirical Wasserstein distance for each layer of the neural network.

A recent alternative, regenerative regularization, regularizes the neural
network parameters towards their initialization
\citep{kumar23_maint_plast_regen_regul}.
The regenerative regularizer mitigates loss of plasticity, but it also prevents the neural network parameters from deviating far from the initialization.
Unlike the regenerative regularizer, the Wasserstein regularizer takes the
difference of the order statistics. Thus, the regenerative regularizer is always larger because the Wasserstein regularizer takes the difference in the sorted values, $\sum_{i=1}^{d} \left(\mathbf{\bar{\theta}}_{(i)}^{(l,t)} - \mathbf{\bar{\theta}}_{(i)}^{(l,0)}\right)^{2} <
\sum_{i=1}^{d} \left(\mathbf{\bar{\theta}}_{i}^{(l,t)} - \mathbf{\bar{\theta}}_{i}^{(l,0)}\right)^{2}$.
As we show in Appendix~\ref{appendix:dist}, the
Wasserstein regularizer
allows the network parameters to deviate further from the initialization. This means that
learning with the Wasserstein regularizer requires fewer iterations while achieving a lower
error compare to other regularizers (see inter-task learning curves, Appendix~\ref{appendix:intertask}).

\section{Experiments: Effect of Curvature and Regularization in Plasticity Benchmarks}
\label{sec:main_exp}

We now validate our claim that loss of curvature, as measured by the reduction in the rank of the Hessian,
explains loss of plasticity.
Our experiments use the four most common continual learning benchmarks in which loss of plasticity has been reported (see Appendix~\ref{appendix:exp_details} for further details):

\begin{itemize}
  \item Permuted MNIST: A commonly used benchmark across continual learning where the pixels are periodically permuted \citep{goodfellow13_empir_inves_catas_forget_gradien,zenke17_contin,kumar23_maint_plast_regen_regul,dohare23_maint_plast_deep_contin_learn,elsayed23_utilit_pertur_gradien_descen}.
  \item Random Label MNIST: A more difficult task change where all labels are randomized \citep{kumar23_maint_plast_regen_regul,lyle23_under,elsayed23_utilit_pertur_gradien_descen}. This problem was used in Section~\ref{sec:counter}, but in this section we use the entire MNIST dataset.
  \item Random Label CIFAR-10 \citep{Krizhevsky09learningmultiple}: An increasingly common problem setting for studying the plasticity of convolutional neural networks due to the relative complexity of images in CIFAR \citep{kumar23_maint_plast_regen_regul,lyle23_under,sokar23_dorman_neuron_phenom_deep_reinf_learn}.
  \item Continual ImageNet \cite{dohare23_maint_plast_deep_contin_learn}: A sequence of 500 binary classification tasks from the ImageNet dataset \citep{russakovsky15_imagen} where none of the classes are shared between tasks.
\end{itemize}

\begin{figure}[t]
  \centering
  \includegraphics[width=0.99\linewidth]{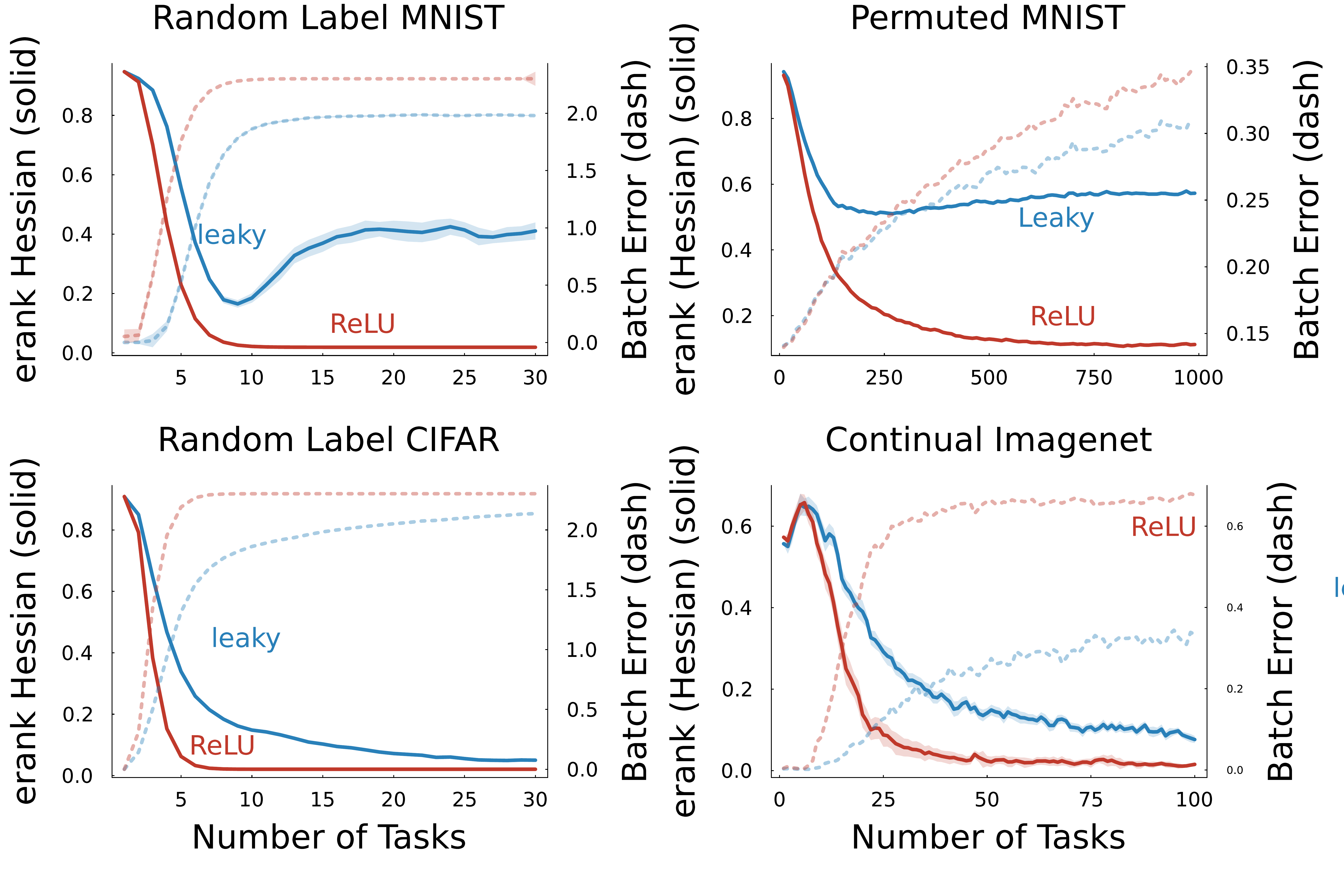}
  \caption{
    \textbf{Validating that a reduction in the directions of curvature is a consistent explanation for loss of plasticity.}
    A reduction in the directions of curvature co-occurs with loss of plasticity. \texttt{leaky-ReLU} preserves plasticity for longer but is unable to recover its directions of curvature.}
  \label{fig:unreg}
\end{figure}

To provide evidence for the claim that curvature explains loss of plasticity, we conduct an in-depth analysis of the change of curvature in continual supervised learning.
We first show that curvature is a consistent explanation across different problem settings.
Afterwards, we investigate the role of curvature on learning to find that the gradient tends to overlap with the shrinking top-subspace of the Hessian (to a degree depending on the activation function).
Lastly, we show that regularization, which has been demonstrated to be effective in mitigating loss of plasticity, also mitigates loss of curvature.
\looseness=-1

\subsection{Does Loss of Curvature Explain Loss of Plasticity?}

We present the results on the four problem settings in Figure~\ref{fig:unreg}.
This is the same setting as the results in Section~\ref{sec:counter}, but with the full MNIST dataset (see Appendix~\ref{appendix:full_act} for results on all activation functions).
Loss of curvature tends to co-occur with loss of plasticity for the non-linear activations, providing a consistent explanation of the phenomenon compared to previous explanations.

\begin{figure}[t]
  \centering
  \includegraphics[width=0.49\linewidth]{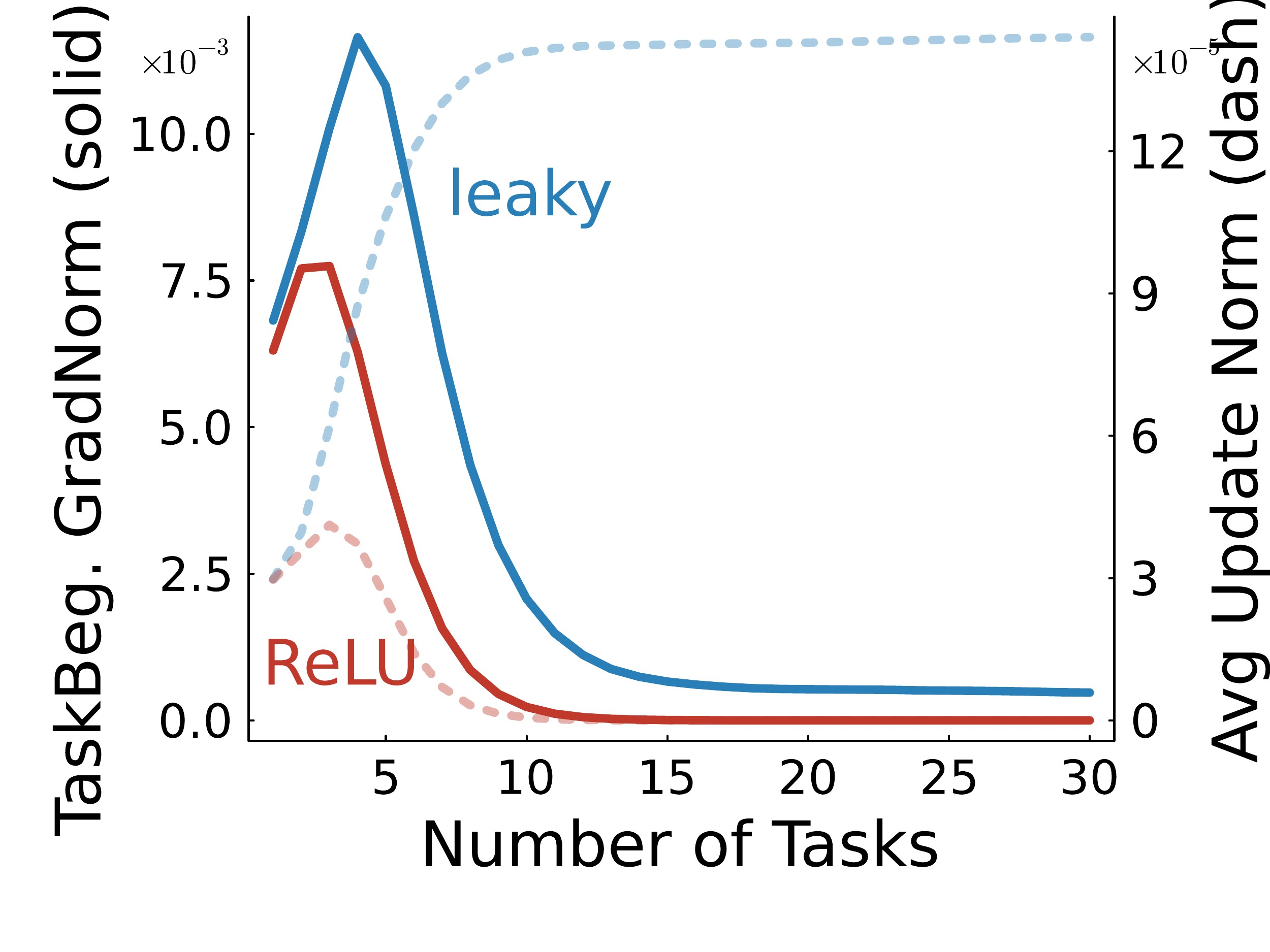}
  \includegraphics[width=0.49\linewidth]{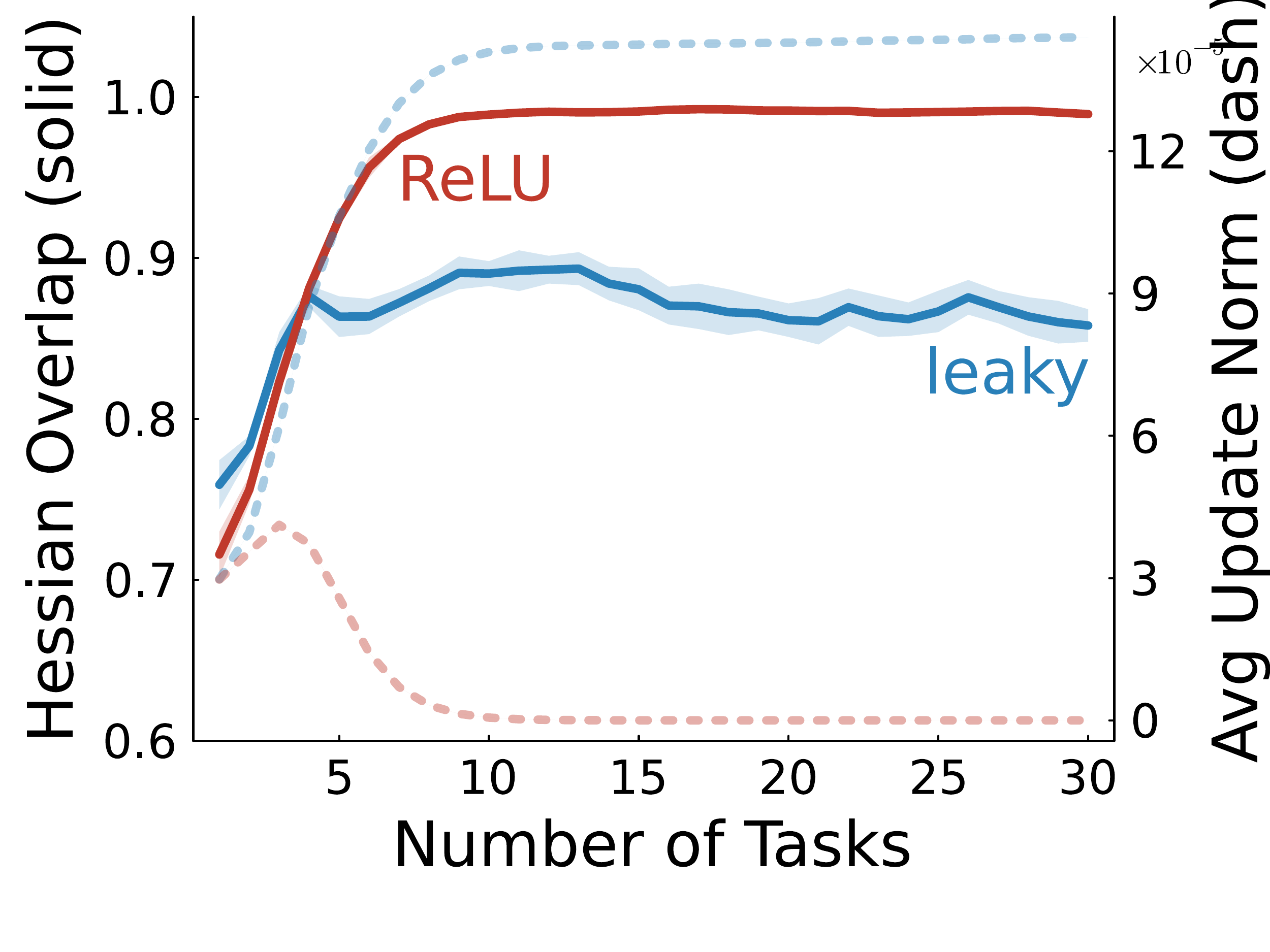}

  \caption{\textbf{Curvature explains why the average update norm increases when using \texttt{leaky-ReLU} despite loss of plasticity.}  Left: \texttt{leaky-ReLU} has an increasing average update norm despite a decrease in the gradient norm at the beginning of a task. Right: gradients with \texttt{leaky-ReLU} have less overlap with the low-rank Hessian, meaning that updates occur in more directions than with \texttt{ReLU}.
    \looseness=-1
  }
\vspace{-5mm}
  \label{fig:updatenorm}
\end{figure}

\subsection{How Does Loss of Curvature Affect Learning?}
Having demonstrated that loss of curvature co-occurs with loss of plasticity, we now investigate how loss of curvature affects the gradients and learning.
Our goal is to explain why the update norms can be increasing for \texttt{leaky-ReLU} despite loss of plasticity.
In Figure~\ref{fig:updatenorm} (Left), we see that the gradient norm at the beginning of each task is decreasing, which neither explains loss of plasticity nor the increasing update norm.
In the right plot, we measure the overlap between the gradient and the (top subspace) Hessian-gradient product at the beginning of a task given by $\frac{g^{T}Hg}{\|g\|\|Hg\|}$.\footnote{We zero out singular values smaller than the effective rank to ensure that the gradient overlaps with the top-subspace Hessian.}
This measures whether the gradient is contained in the top subspace of the Hessian \citep{gur-ari18_gradien_descen_happen_tiny_subsp}.
For \texttt{leaky-ReLU}, the gradient has less overlap with the top subspace of its Hessian. This means that updates with \texttt{leaky-ReLU} explore a higher dimensional space than than either \texttt{tanh} or \texttt{ReLU}, explaining why its average update norm is higher.

\begin{figure}[t]
  \centering
  \includegraphics[width=0.99\linewidth]{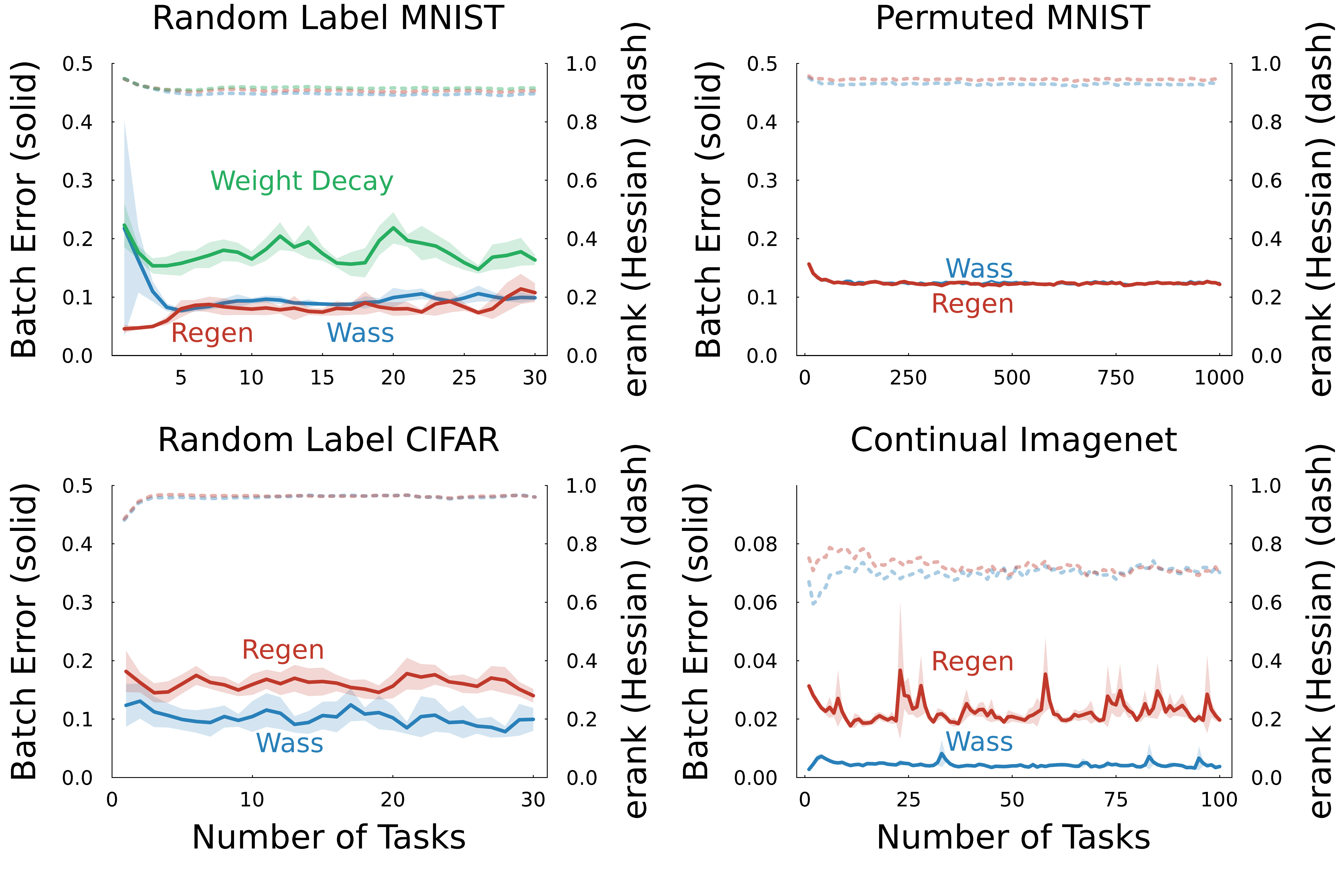}
  \caption{
    \textbf{Regularization preserves plasticity and directions of curvature.}
    Wasserstein and regenerative regularizers are effective at preserving plasticity and curvature. On harder problems (bottom), the Wasserstein regularizer achieves a lower error.
  }
  \label{fig:reg}
  \vspace{-3mm}
\end{figure}

\subsection{Can Regularization Preserve Curvature?}
We now investigate whether regularization prevents loss of plasticity and, if it does, whether it also preserves directions of curvature.
Our results for the four problem settings are summarized in Figures~\ref{fig:reg}.
We see that the Wasserstein is able to preserve plasticity, achieving similar error to the regenerative regularizer on the easier MNIST problems and achieving the lowest error on Random Label CIFAR and Continual ImageNet.
The success of the Wasserstein regularizer can be seen from two perspectives: 1)  parameters can move further from initialization (see Appendix~\ref{appendix:dist}) and 2) reduced sensitivity to the regularization strength (see Appendix~\ref{appendix:reg_hyperparam}). The inter-task learning curves reveal that learning with the Wasserstein regularizer not only achieves a lower error, but that learning can require fewer iterations (see Appendix~\ref{appendix:intertask}). Lastly, we find that the feature rank is often decreasing for the regularized neural networks, which further demonstrates its inconsistency as an explanation for loss of plasticity (see Appendix~\ref{appendix:more_feature_rank}).
\looseness=-1

\subsection{Does Scale Help Preserve Plasticity \& Curvature?}

\begin{figure}[t]
  \centering
  \includegraphics[width=0.49\linewidth]{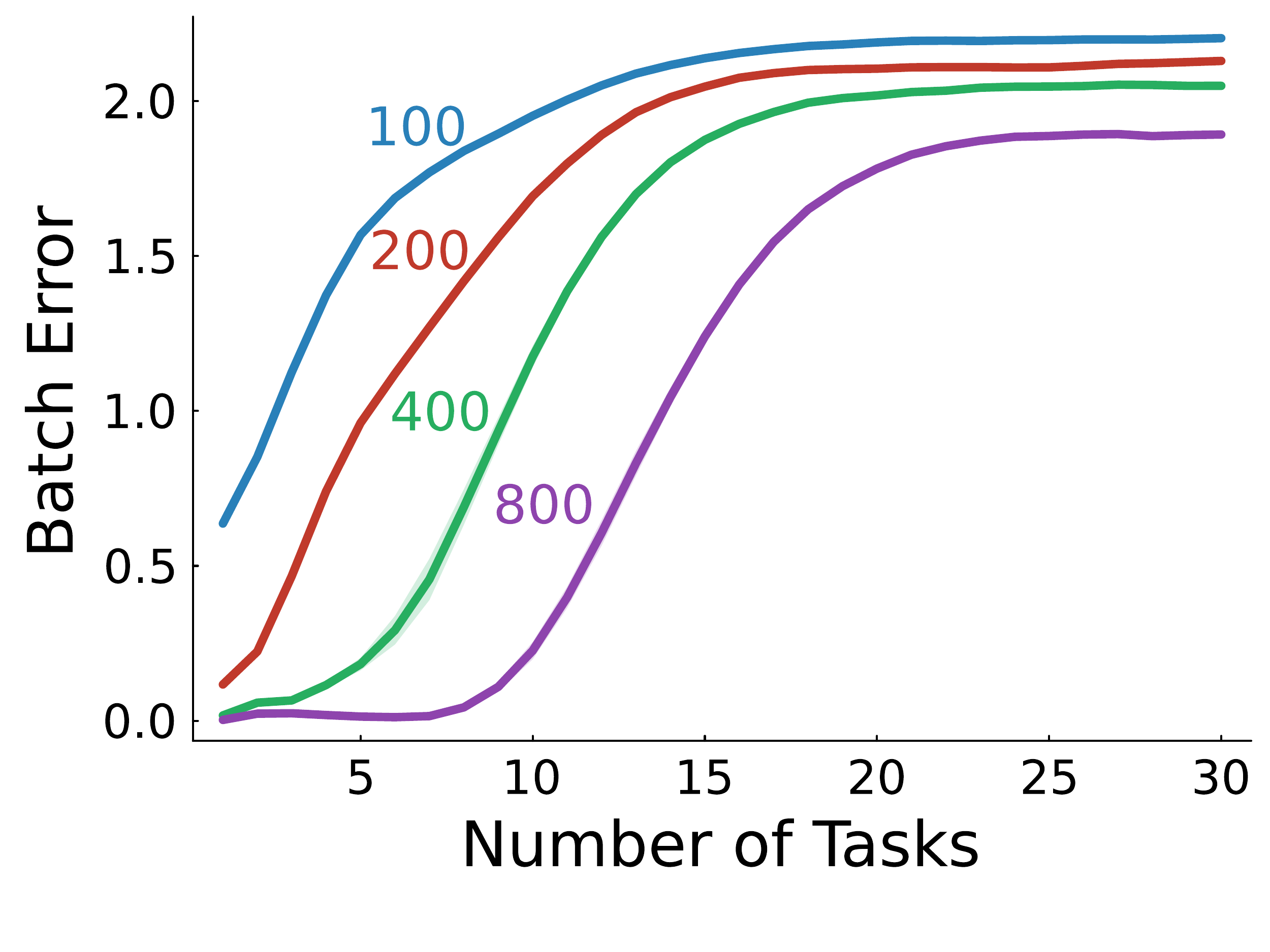}
  \includegraphics[width=0.49\linewidth]{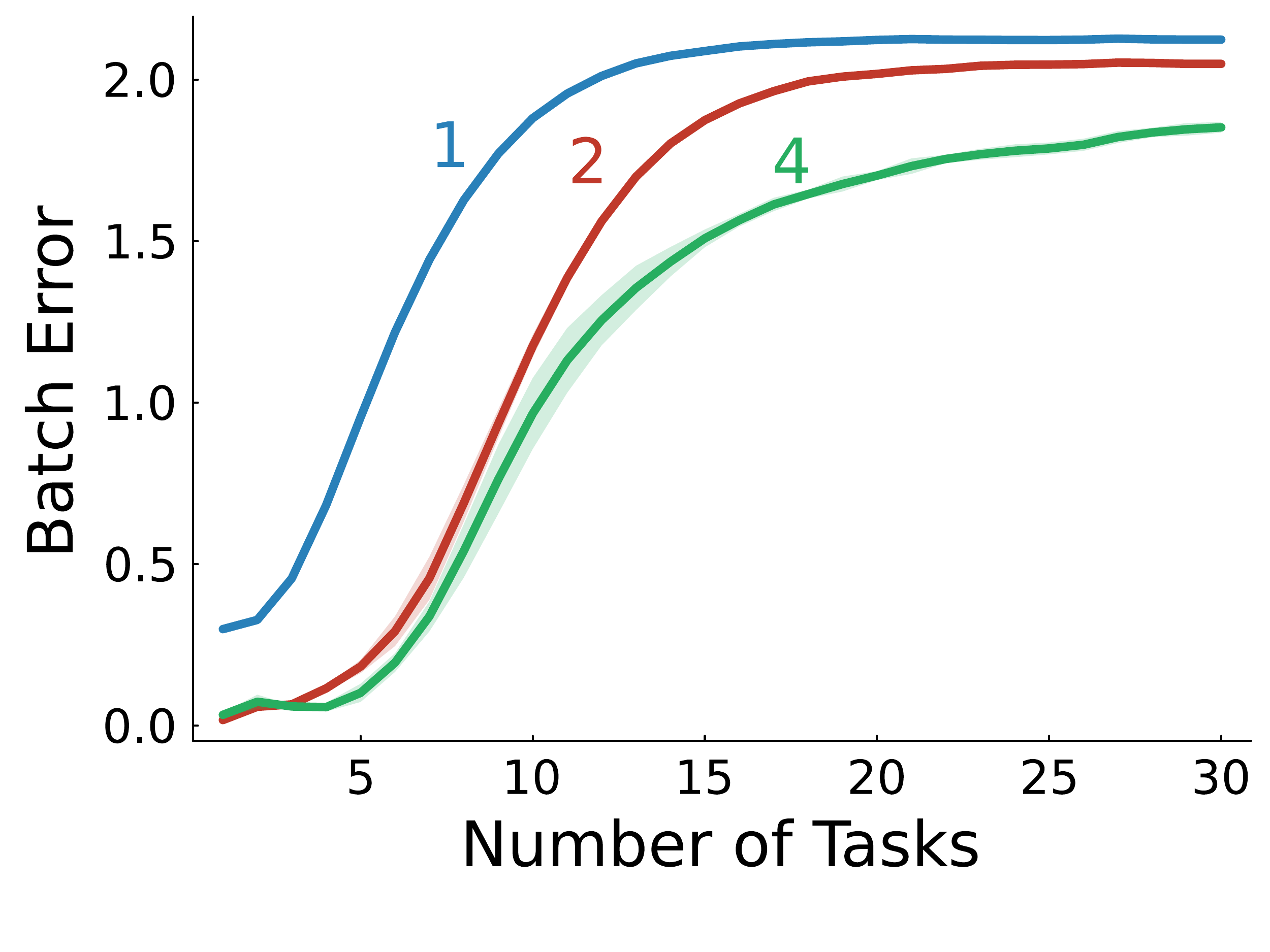}

  \caption{
    \textbf{Effect of width and depth on loss of plasticity.}
    Increasing either the width of the hidden layer or the depth (number of hidden layers) in a neural network delays loss of plasticity, and marginally lowers the error plateau, but does not eliminate loss of plasticity. Right: Varying the width, while keeping the depth fixed at 4. Left: Varying the depth, while keeping the width fixed at 800.}
  \label{fig:mnist_archablate}
  \vspace{-3mm}
\end{figure}

\begin{figure}[b]
  \centering
\includegraphics[width=0.49\linewidth]{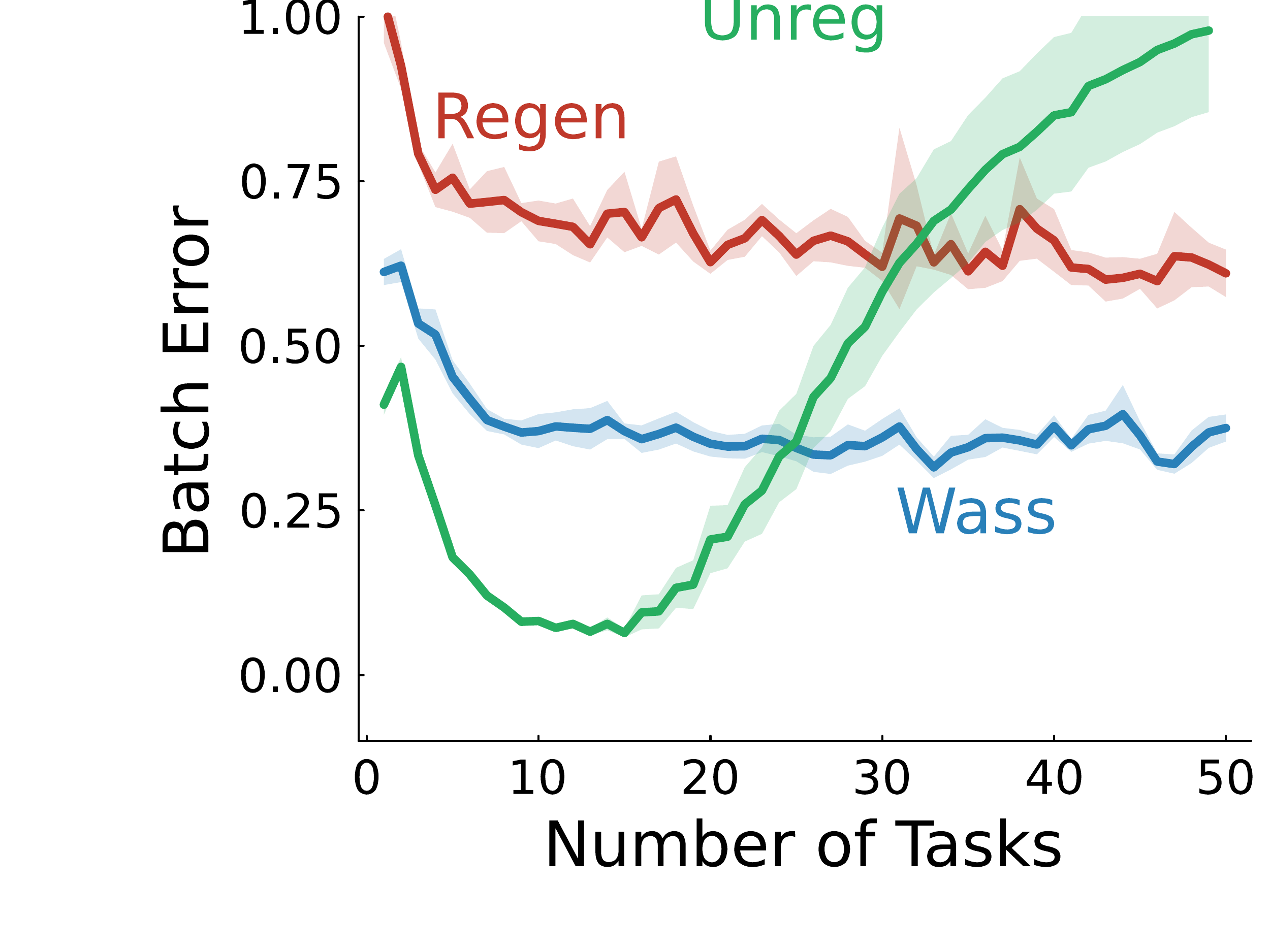}
\includegraphics[width=0.49\linewidth]{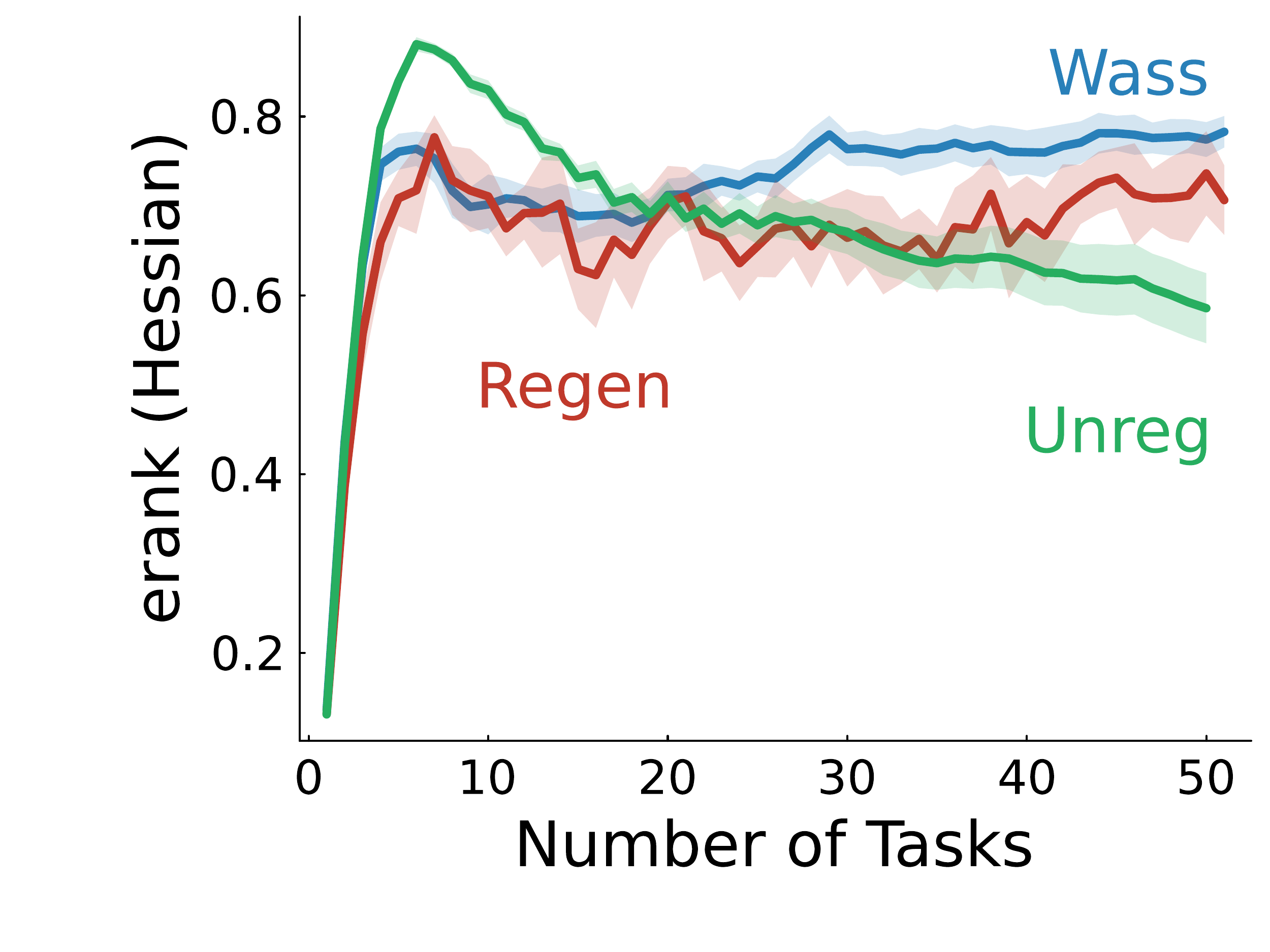}
\caption{
  \textbf{ResNet18 without regularization still suffers from loss of plasticity.}
  Despite a much higher higher parameter count and batch normalization, the ResNet is not able to maintain its initial error without regularization due to a reduction in the number of curvature directions, as measured by the rank of the Hessian.
}
  \label{fig:cifar_resnet}
\end{figure}

To investigate the role of neural network scale, we ablate different neural network widths and depths. The results in Figure~\ref{fig:mnist_archablate} show that increasing both the depth and width of the neural network only delays loss of plasticity. In Figure~\ref{fig:cifar_resnet}, we test whether loss of plasticity occurs in CIFAR-10 using a much larger network with batch normalization, ResNet18 \citep{he16_deep}. Unlike the previously considered convolutional networks, the ResNet is able to decrease the error on the first few tasks despite training for only 20 epochs. However, loss of plasticity still occurs without regularization. With regularization, the ResNet is able to achieve an error level slightly higher than the best error that the unregularized version can achieve.\looseness=-1

\section{Discussion}

We have demonstrated how loss of curvature directions is a consistent explanation for loss of plasticity when compared to previous explanations offered in the literature.
One limitation of our work is that we study an approximation to the Hessian.
Our experiments suggest that this approximation of the Hessian is enough to capture changes in the number of curvature directions, but more insight may be found from theoretical study of the entire Hessian.
Another limitation is that it is not clear what drives neural networks to lose curvature directions during training.
Understanding the dynamics of training neural networks with gradient descent, however, is an active research area even in supervised learning.
It will be increasingly pertinent to understand what drives neural network training dynamics to lose curvature directions so as to develop principled algorithms for continual learning.
\looseness=-1

Our experimental evidence demonstrates that, when loss of plasticity occurs, there is a reduction in curvature as measured by the rank of the Hessian at the beginning of subsequent tasks.
When loss of plasticity does not occur, curvature remains relatively constant.
Unlike previous explanations, this phenomenon is consistent across different datasets, non-stationarities, step-sizes, and activation functions.
Lastly, we investigated the effect of regularization on plasticity, finding that regularization tends to preserve curvature but can be sensitive to the regularization strength.
We proposed a simple distributional regularizer that proves effective in maintaining plasticity across the problem settings we consider, while maintaining curvature and being less hyperparameter sensitive.

\newpage
\section*{Acknowledgments}
We thank Shibhansh Dohare, Khurram Javed, Farzane Aminmansour and Mohamed Elsayed for early discussions about loss of plasticity. The research is supported in part by the Natural Sciences and Engineering Research Council of Canada (NSERC), the Canada CIFAR AI Chair Program, the Digital Research Alliance of Canada and Alberta Innovates Graduate Student Scholarship.

\section*{Impact Statement}
This paper presents work whose goal is to advance the field of Machine Learning. There are many potential societal consequences of our work, none which we feel must be specifically highlighted here.

\bibliography{iclr2024_conference}
\bibliographystyle{icml2024}

\appendix
\section*{Appendix}

\section{Additional analysis on counter-examples}
\label{sec:appendix_analysis}
In the body of paper, we provided a high-level analysis of Figure~\ref{fig:confounding}, and concluded that none of the previous explanations for loss of plasticity (i.e. increasing error) is consistent amongst the different activation fucntions.
Here, we aim to compliment that high-level analysis by providing detailed explanation on how each metric is inconsistent with the batch error.

\begin{enumerate}
    \item \textbf{Average Update Norm} (top-left):
    The plot measures the average L1 norm of the parameter updates, and it is predicted that a decrease in the update norm leads to loss of plasticity.
    Both \texttt{Leaky-ReLU} and \texttt{ReLU} exhibit the opposite trend in their update norm: the former is increasing its average update norm and the latter is decreasing.
    But, both activation functions have an increasing error and thus suffer from loss of plasticity.
    Hence, the update norm is an inconsistent explanation for loss of plasticity

    \item \textbf{Effective Rank of Representation} (top-right):
    The plot measures the normalized effective rank of the representation (the last hidden layer that is mapped linearly to the output space), and it is predicted that a decrease in the feature rank leads to loss of plasticity.
    For \texttt{ReLU}, the representation rank decreases along with the error increasing, which is what the effective rank explanation of plasticity predicts.
    The representation rank is inconsistent because \texttt{tanh} has an initial drop of its representation rank despite the error remaining constant.
    Hence, the representation rank is an inconsistent explanation for loss of plasticity.

    \item \textbf{Dormant Neurons} (bottom-left):
    The plot measures neuron dormancy by the negative of the entropy of the normalized absolute value of the features for each task, which captures the notion of dormancy that activations can concentrate on a small subset of features.
    It is predicted that an increase in neuron dormancy will lead to loss of plasticity
    The plot shows that the \texttt{ReLU} activation has an increase in neuron dormancy and an increasing error, which is what neuron dormancy predicts.
    But, \texttt{leaky-ReLU} experiences plasticity loss and the neuron dormancy is non-decreasing.
    Hence, the dormant neuron phenomenon is an inconsistent explanation for loss of plasticity.

    \item \textbf{Weight Norm} (bottom-right):
    The plot presents the L1 norm of the weights at the end of each task, and it is predicted that an increasing norm leads to loss of plasticity
    Both \texttt{ReLU} and \texttt{identity} provide counterexamples.
    For \texttt{ReLU}, the weight norm plateaus but loss of plasticity occurs.
    For \texttt{identity}, the weight norm increases seemingly indefinitely and yet, loss of plasticity does not occur.
    Hence, the weight norm is an inconsistent explanation for loss of plasticity.

\end{enumerate}

\section{Experimental Details}
\label{appendix:exp_details}

\subsection{Random Label MNIST}
\label{appendix:exp_details_label_mnist}
Non-stationary variant of the ordinary (stationary) supervised classification problem on MNIST dataset.
The source of non-stationarity in this problem is the periodical random shuffling of labels, irrespective of the original class labels.
The dataset consists of $51200$ uniformly sampled MNIST image-label pairs.
We iterate over the dataset for 200 epochs in the experiments in the main paper, but ablate for different number of epochs in Section~\ref{appendix:epochs}.
After 200 epochs,  the labels will be reshuffled within the same dataset, producing the new task.
Each gradient updates are performed with the batch of 256 datapoints, hence the update number of updates per epoch is 200 and the number of updates in the task is 40000.
The architecture is a 3 hidden layer feed-forward neural network with widths $(256, 256, 256)$.
We use the Adam optimizer with default hyperparameters.
We average over 30 seeds for the unregularized experiments and average over 30 seeds for the regularized experiments.
For the regularized experiments, we sweep over the regularization strength of $\{0.005, 0.001, 0.0005\}$. We use \texttt{leaky-ReLU} for all regularized experiments (except with the ResNet) due to its increased effectiveness in the continual learning setting.

\subsection{Permuted MNIST}
The overall problem framework is identical to the Random Label MNIST, except for the source of non-stationarity.
The non-stationarity is introduced by reordering the positions of pixels in each input image, while label remains the same throughout the experiment.
At the beginning of each task, the permutation of pixels are shuffled, and each input images are uniformly shuffled according to that permutation.
For the regularized experiments, we sweep over the regularization strength of $\{0.01, 0.005, 0.001, 0.0005\}$.
Other components of experiment do not vary from Random Label MNIST problem.

\subsection{Random Label CIFAR-10}
A non-stationary supervised classification problem using the CIFAR-10 dataset, similar to the Random Label MNIST problem.
Similarly in the label-shuffled MNIST problem, this problem uniformly samples $38400$ datapoints from CIFAR-10.
The architecture uses 4 convolutional layers with stride 2 and $(16, 32, 64, 128)$ filters, before flattening and using a single layer feed-forward neural network with width $(512)$.
For the regularized experiments, we sweep over the regularization strength of $\{0.01, 0.005, 0.001, 0.0005\}$.
Other components of experiment do not vary from Random Label MNIST problem.
The ResNet18 architecture \citep{he16_deep} is unchanged, using \texttt{ReLU} and batch normalization. We train the network for a reduced number of epochs (20) to demonstrate that the ResNet can initially improve its plasticity before losing plasticity. The regularized ResNet uses a regularization strength of $0.005$ which was the best regularization strength found on the smaller convolutional neural network.

\subsection{Continual ImageNet}
We use the Continual ImageNet environment introduced by \cite{dohare23_maint_plast_deep_contin_learn}.
We train the same convolutional neural network as before, but for 250 epochs.
For the regularized experiments, we sweep over the regularization strength of $\{0.01, 0.005, 0.001, 0.0005\}$.
Other components of experiment do not vary from Random Label CIFAR problem.

\section{Additional Results}

\subsection{Average Online Error Can Suggest Loss of Plasticity Even in Its Absence}
\label{appendix:online_error}
Average online error is another metric for studying loss of plasticity, but it can misdiagnose the phenomenon.
Even if a neural network maintains a consistent error at the end of a task, its online error can increase due to an increase in its error at the beginning of a task.
But the error at the beginning of a task is not controllable, because it is due to a non-stationarity in the experience.
Thus, we focus on the batch error at task end alone.

\begin{figure}[h!]
  \centering
  \includegraphics[width=0.49\linewidth]{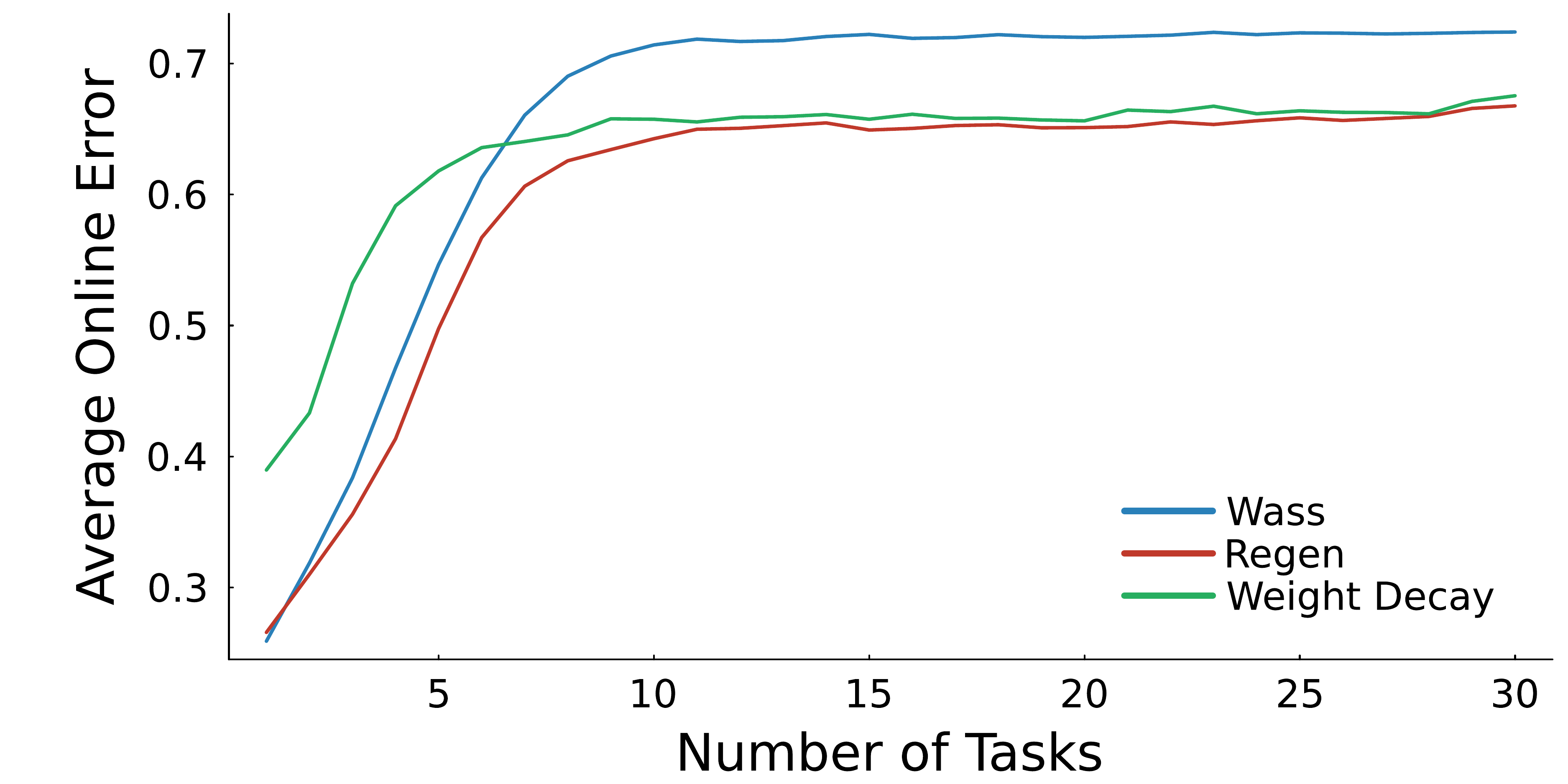}
  \includegraphics[width=0.49\linewidth]{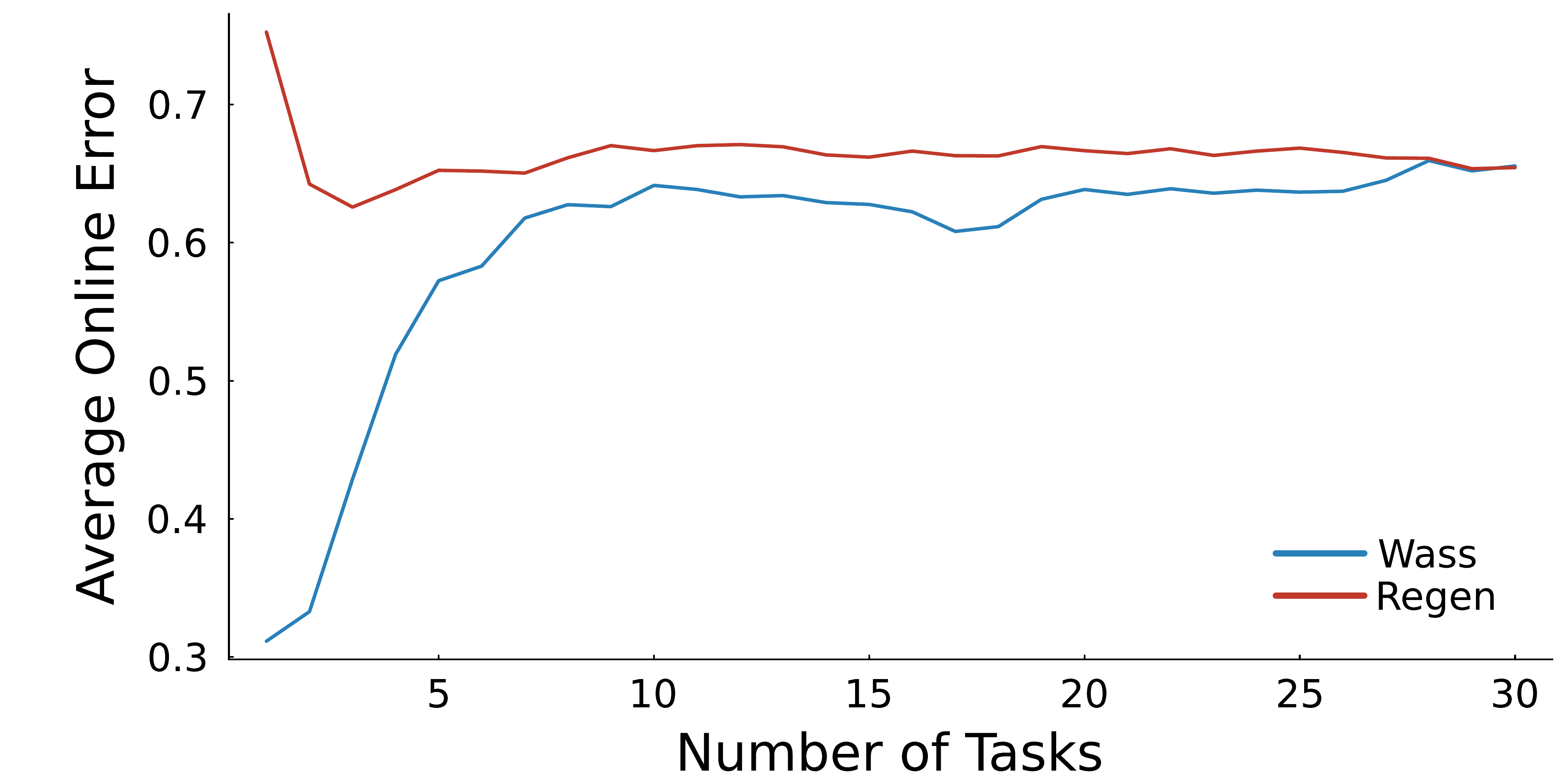}
  \caption{
    Regularization prevents loss of plasticity, in the sense that the error at the end of the task is constant. The average online error increases, because the error at the start increases.
    }
  \label{fig:mnistshuffle_avgonline}
\end{figure}

\subsection{Further Discussion and Results on Hessian Approximation}
\label{appendix:hessian_approx}
We use a stochastic projection matrix to reduce the dimensionality of the MNIST images to $36$, then use a neural network with 3 hidden layers with 32 neurons. While the scale of this problem is small, its results with respect to plasticity remain strikingly similar to the larger scale problems in the main experiments.

The Fisher approximation differs from the empirical fisher approximation because it requires sampling from the predictive distribution induced by the classifier, and we use only 1 sample per datapoint. Sampling additional times would be more effective but less efficient. The Gauss-Newton approximation is $ \mathbf{H} \approx J_{f}^T H_{z} J_{f}$, where $J_{f}$ is the Jacobian of the neural network output and $H_{z}$ is the Hessian of the loss function with respect to the prediction. we cannot interchange the inner and outerproduct because of the middle Hessian matrix. Thus, calculating the svd cannot be made efficient.

\begin{figure}[h!]
  \centering
  \includegraphics[width=0.99\linewidth]{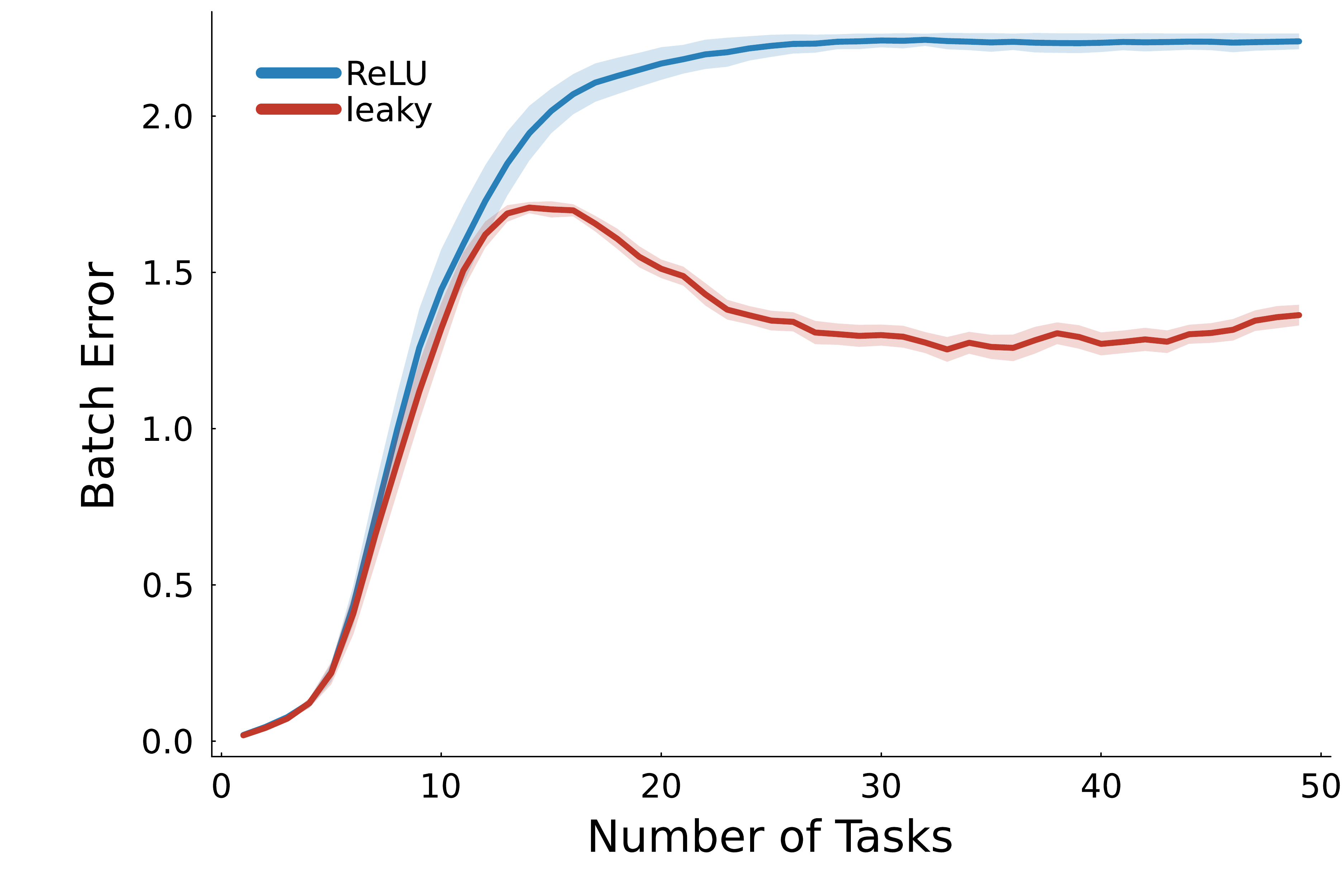}
  \caption{
    Neural networks suffer from loss of plasticity with both activations on low-dimensional projected MNIST.
    }
  \label{fig:approxhessian_errors}
\end{figure}

\begin{figure}[h!]
  \centering
  \includegraphics[width=0.49\linewidth]{plots/polished_icml/confounding_approxhessian}
  \includegraphics[width=0.49\linewidth]{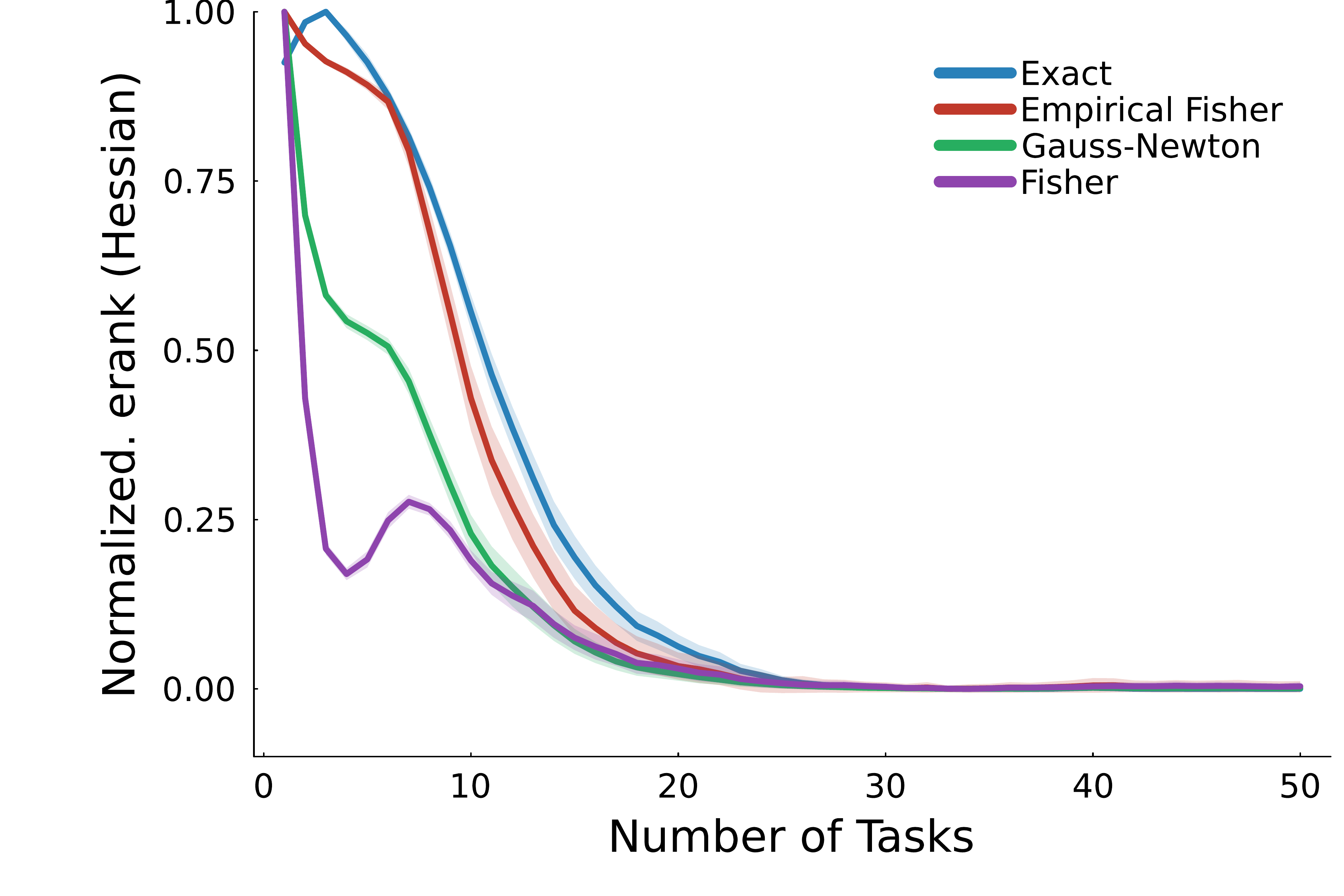}
  \caption{
    With \texttt{ReLU}, the empirical Fisher approximation accurately approximates the normalized rank throughout continual learning.
    }
  \label{fig:approxhessian_relu}
\end{figure}

\begin{figure}[h!]
  \centering
  \includegraphics[width=0.49\linewidth]{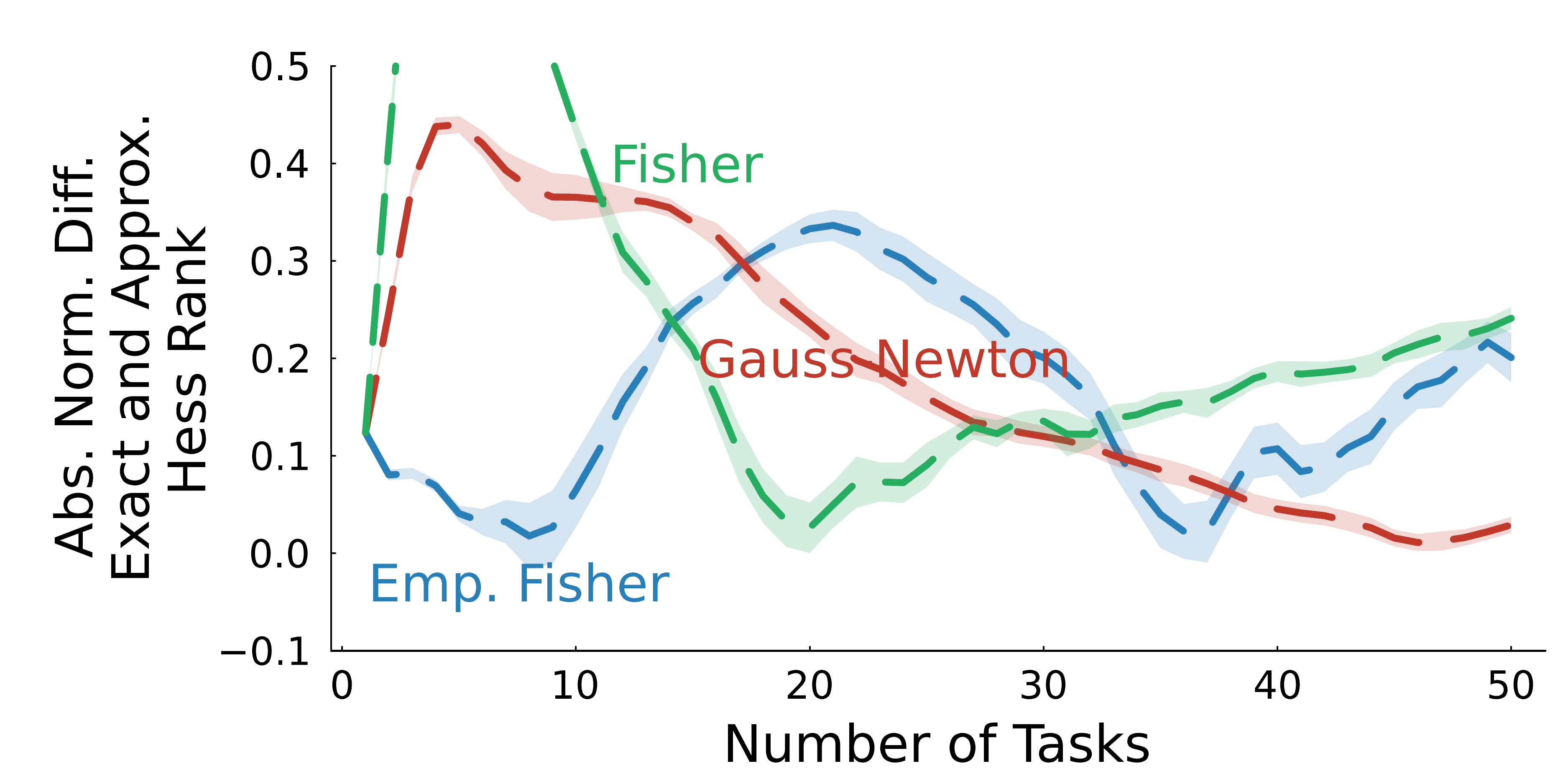}
  \includegraphics[width=0.49\linewidth]{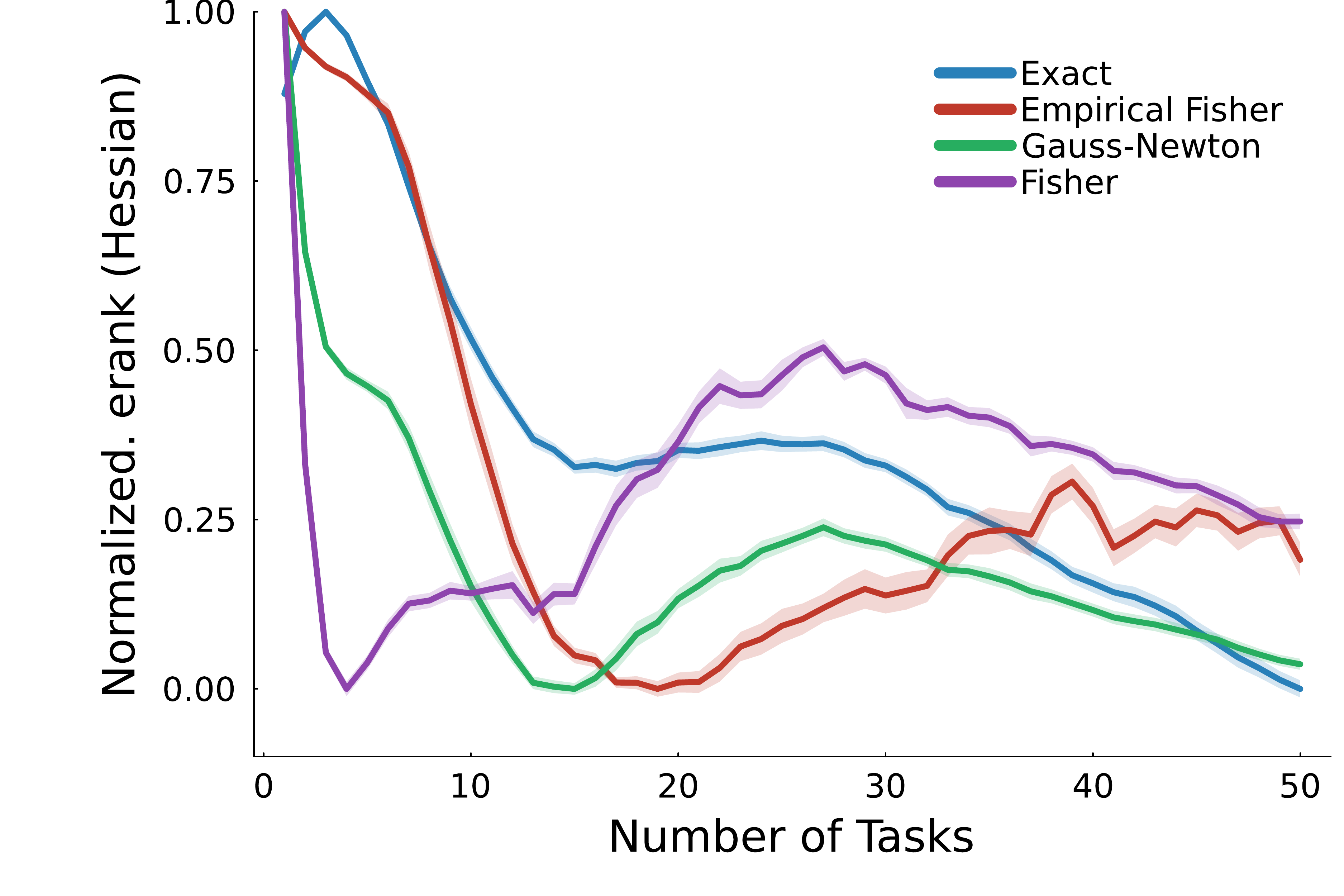}
  \caption{
    With \texttt{leaky-ReLU}, the empirical Fisher approximation accurately approximates the normalized rank
    when plasticity is being lost, which is sufficient as an indicator for loss of plasticity.
  }
  \label{fig:approxhessian_leaky}
\end{figure}

\newpage
\subsection{Results on All Activation Functions}
\label{appendix:full_act}

\begin{figure}[h!]
  \centering
  \includegraphics[width=0.99\linewidth]{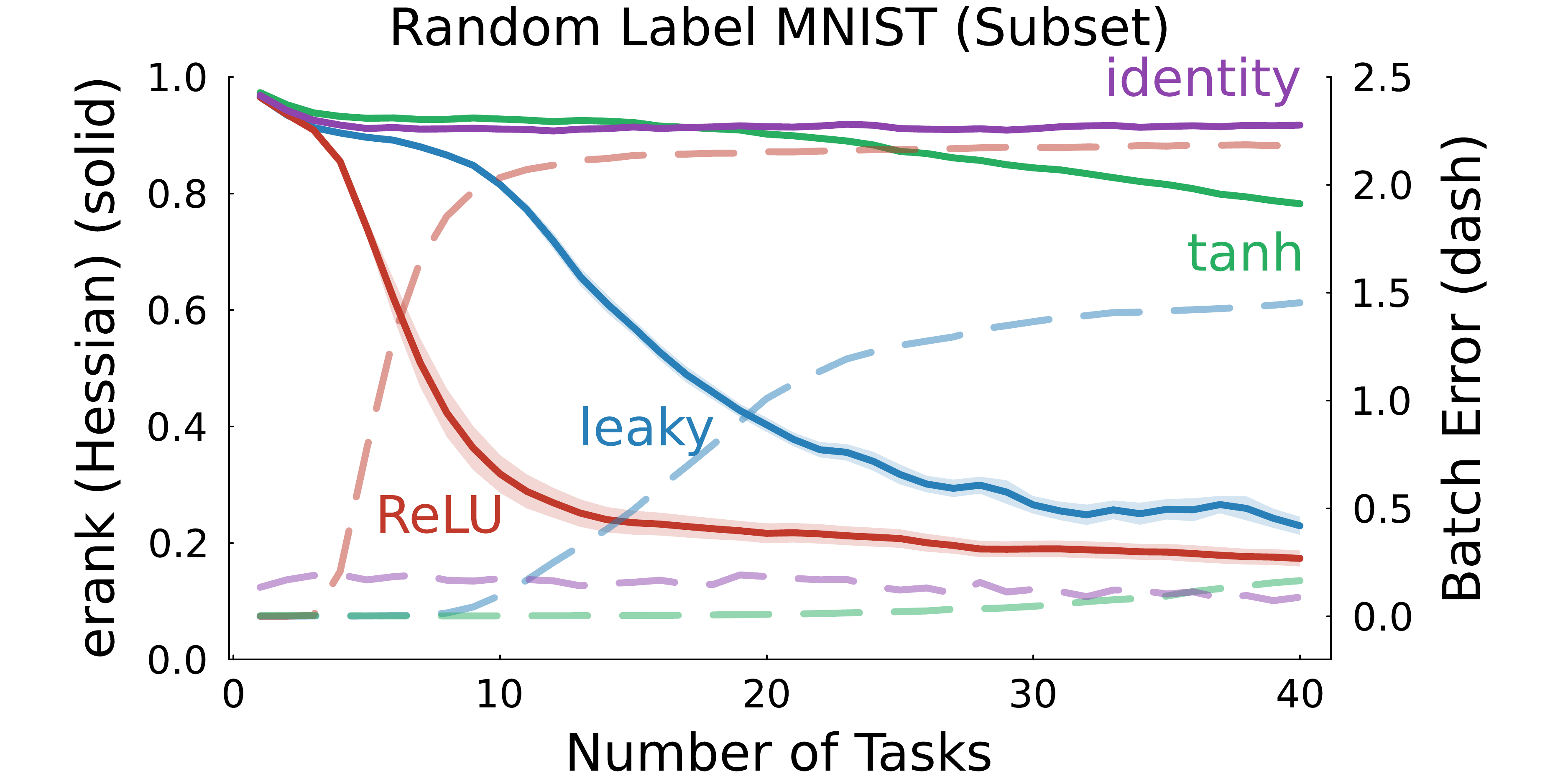}
  \caption{
    If we run the experiment in Section~\ref{sec:counter} for more tasks, \texttt{tanh} eventually loses plasticity and the Hessian rank accurately predicts this whereas the feature rank remains constant.
    }
  \label{fig:confounding_hessian}
\end{figure}

\begin{figure}[h!]
  \centering
  \includegraphics[width=0.99\linewidth]{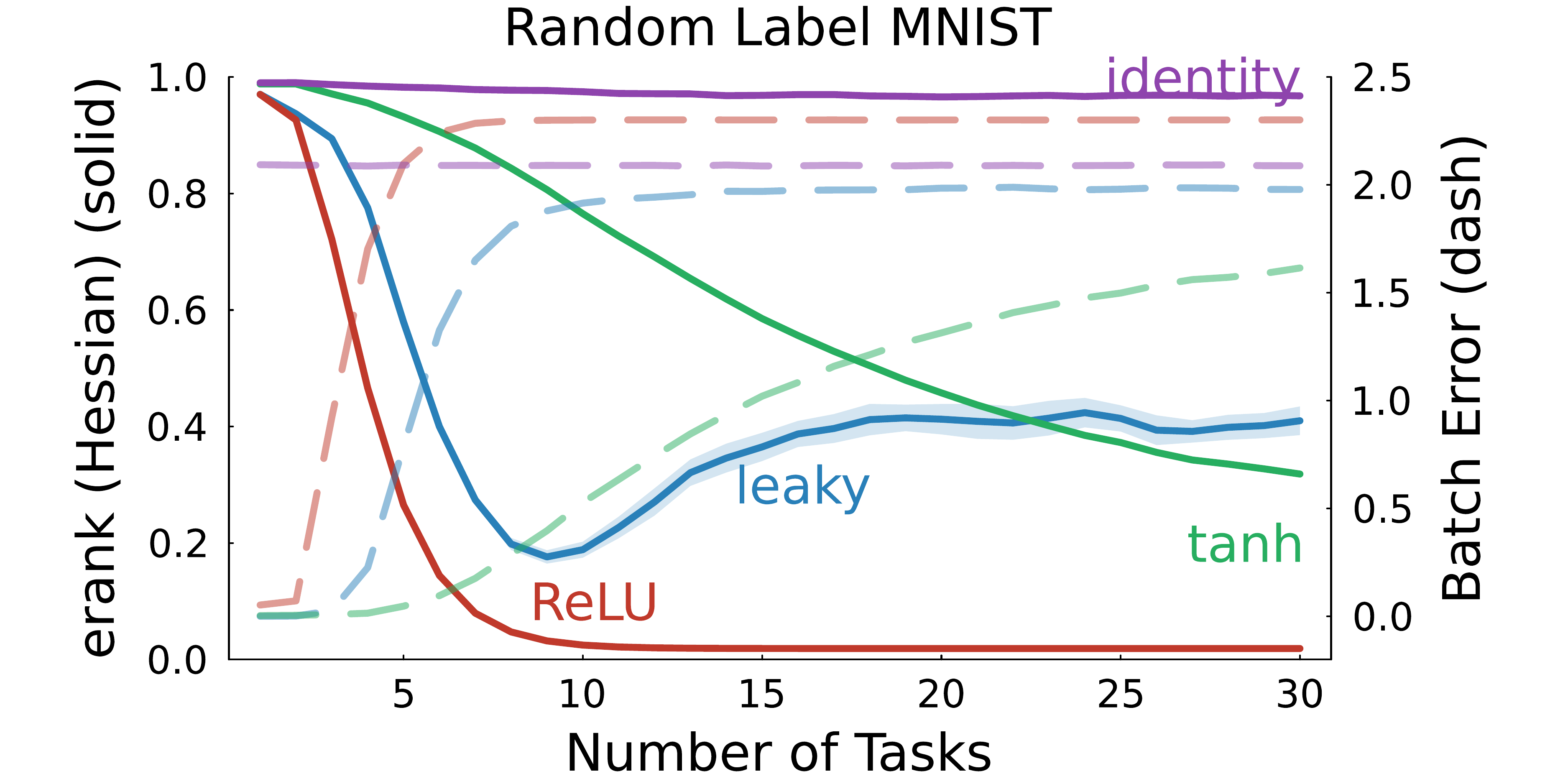}
  \caption{
    On the full dataset, the deep linear network does not have enough capacity to learn and it's error remains high but constant. All non-linear activation functions lose plasticity, which the Hessian rank correctly explains.
    }
  \label{fig:full_hessian}
\end{figure}

\subsection{Parameter Regularization Preserves Plasticity But Does Not Always Control Feature Rank}
\label{appendix:more_feature_rank}

\begin{figure}[h!]
  \centering
  \includegraphics[width=0.49\linewidth]{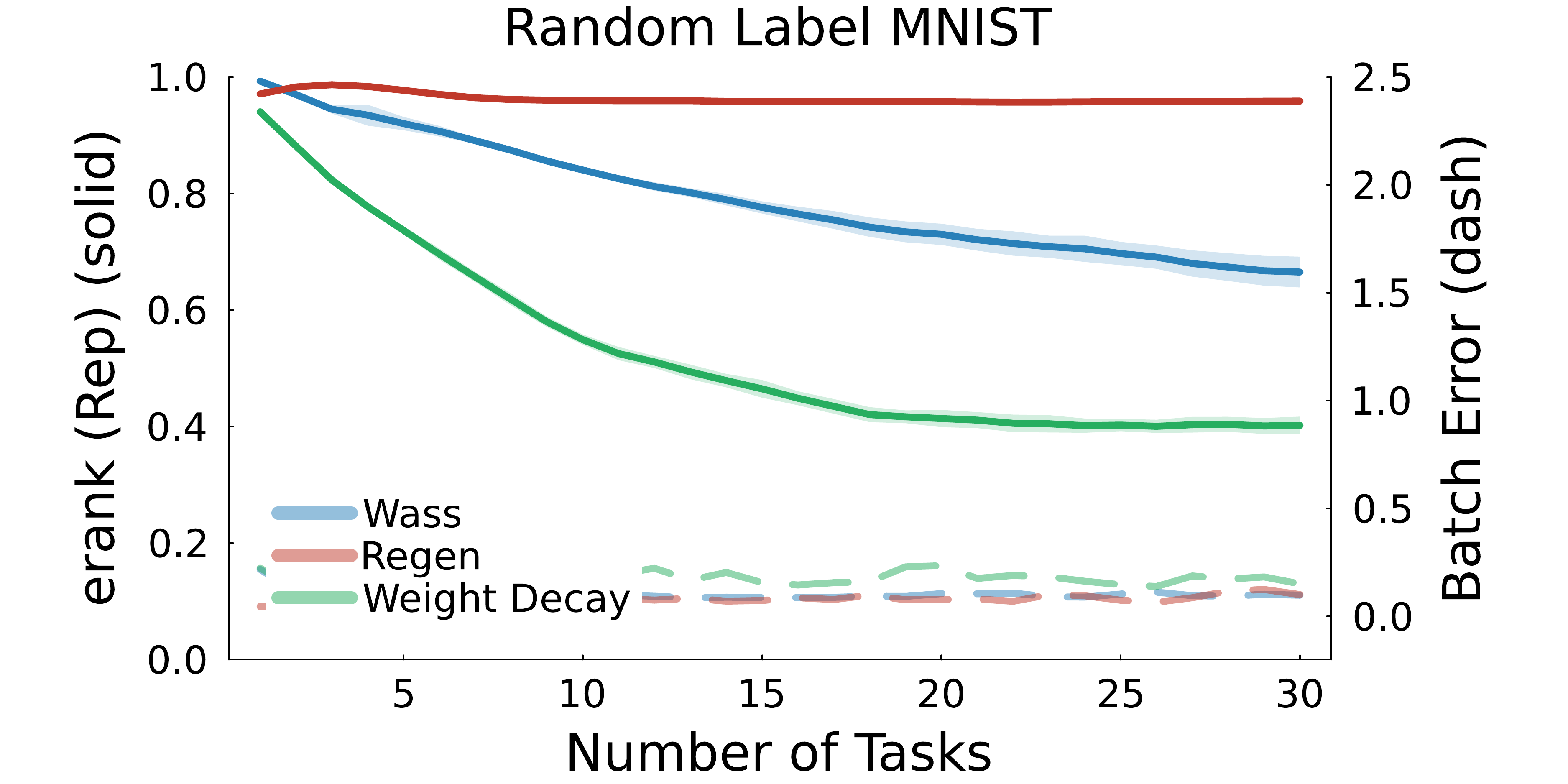}
  \includegraphics[width=0.49\linewidth]{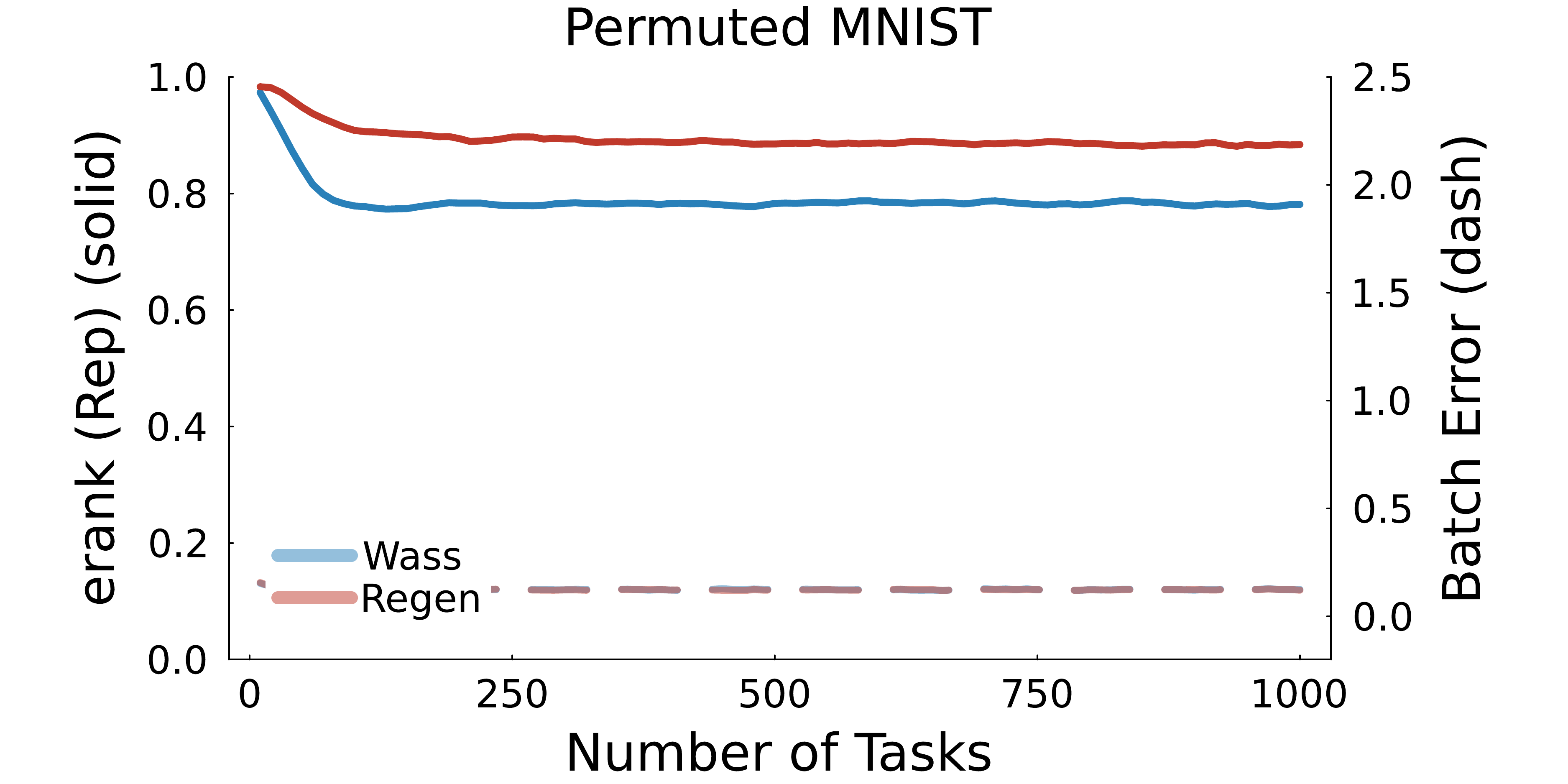}\\
  \includegraphics[width=0.49\linewidth]{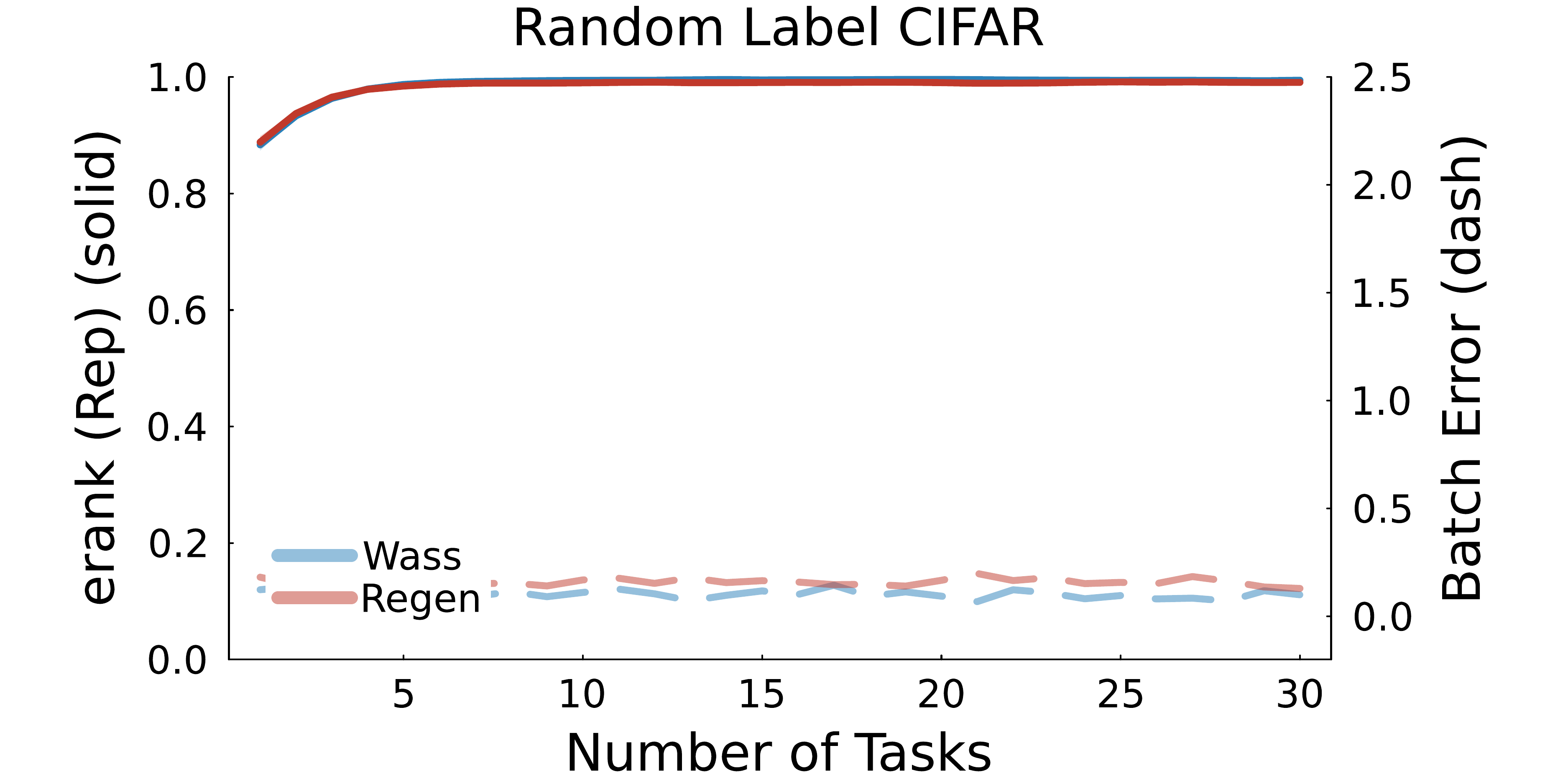}
  \includegraphics[width=0.49\linewidth]{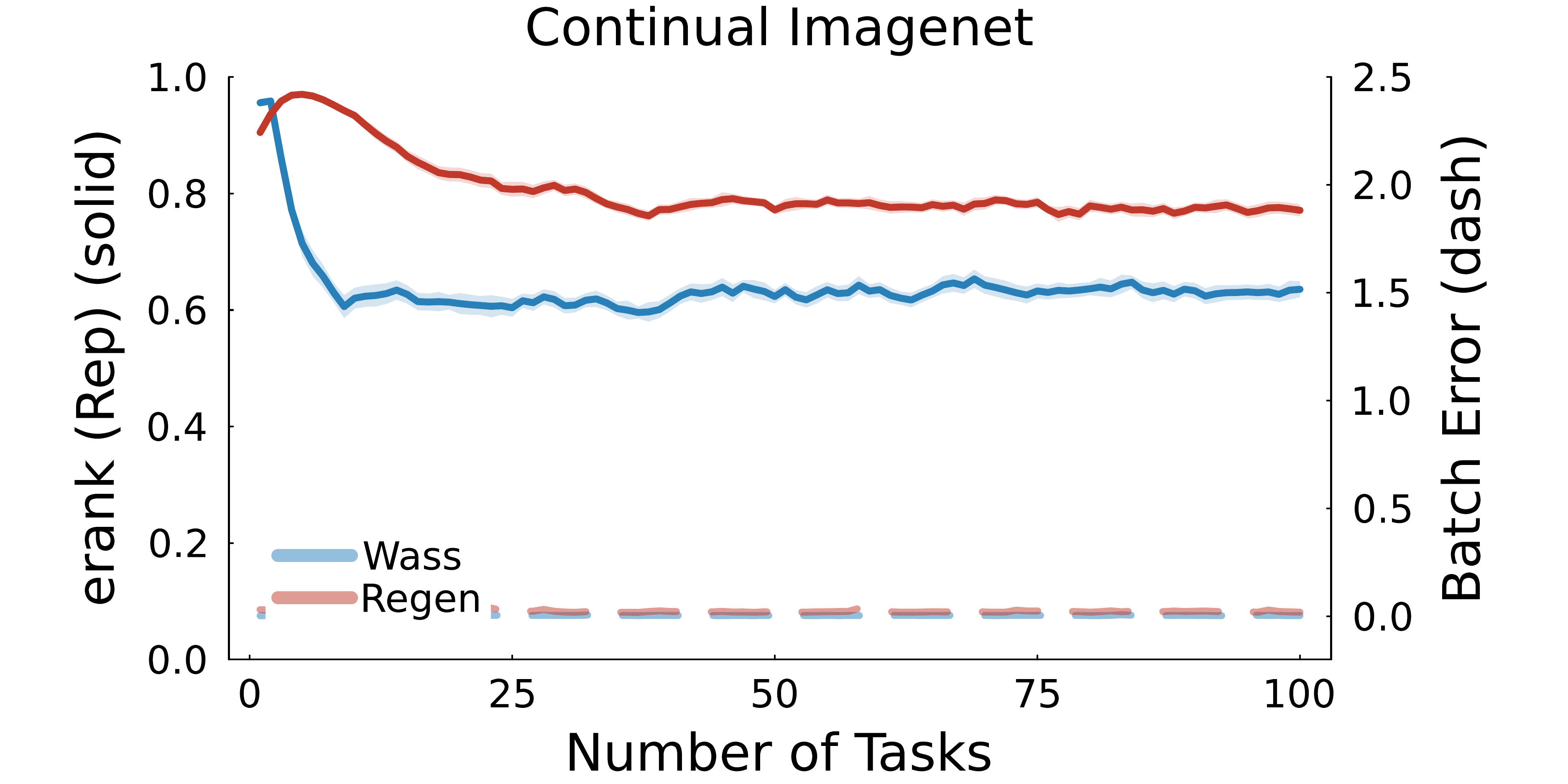}
  \caption{
With regularization, the feature rank sometimes still decreases. The decrease is problem dependent, and only in the CIFAR problem does the feature rank increase.
    }
  \label{fig:reg_and_feature_rank}
\end{figure}

\newpage
\subsection{Distances from Initialization With and Without Regularization}
\label{appendix:dist}

\begin{figure}[h!]
  \centering
  \includegraphics[width=0.49\linewidth]{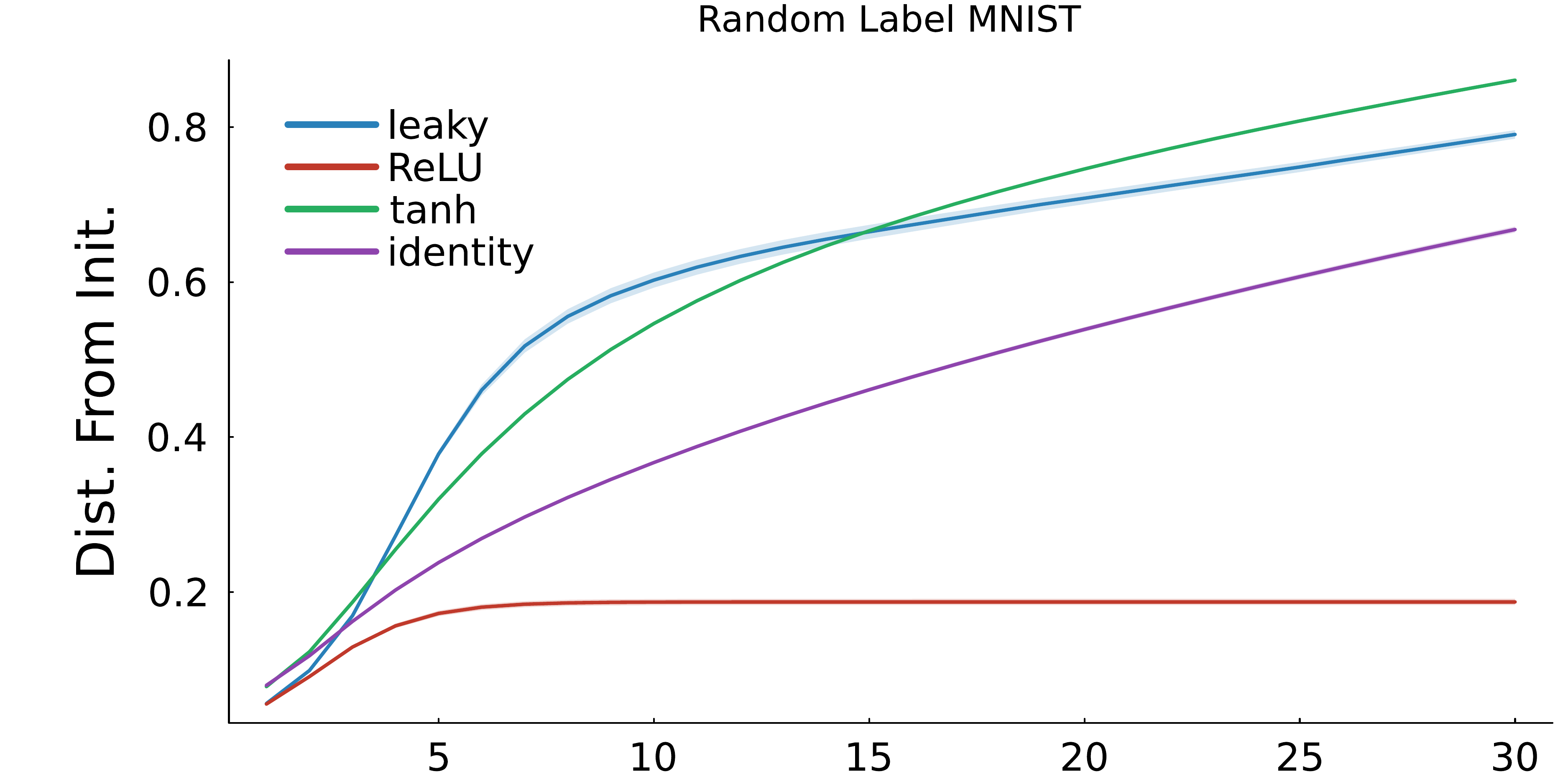}
  \includegraphics[width=0.49\linewidth]{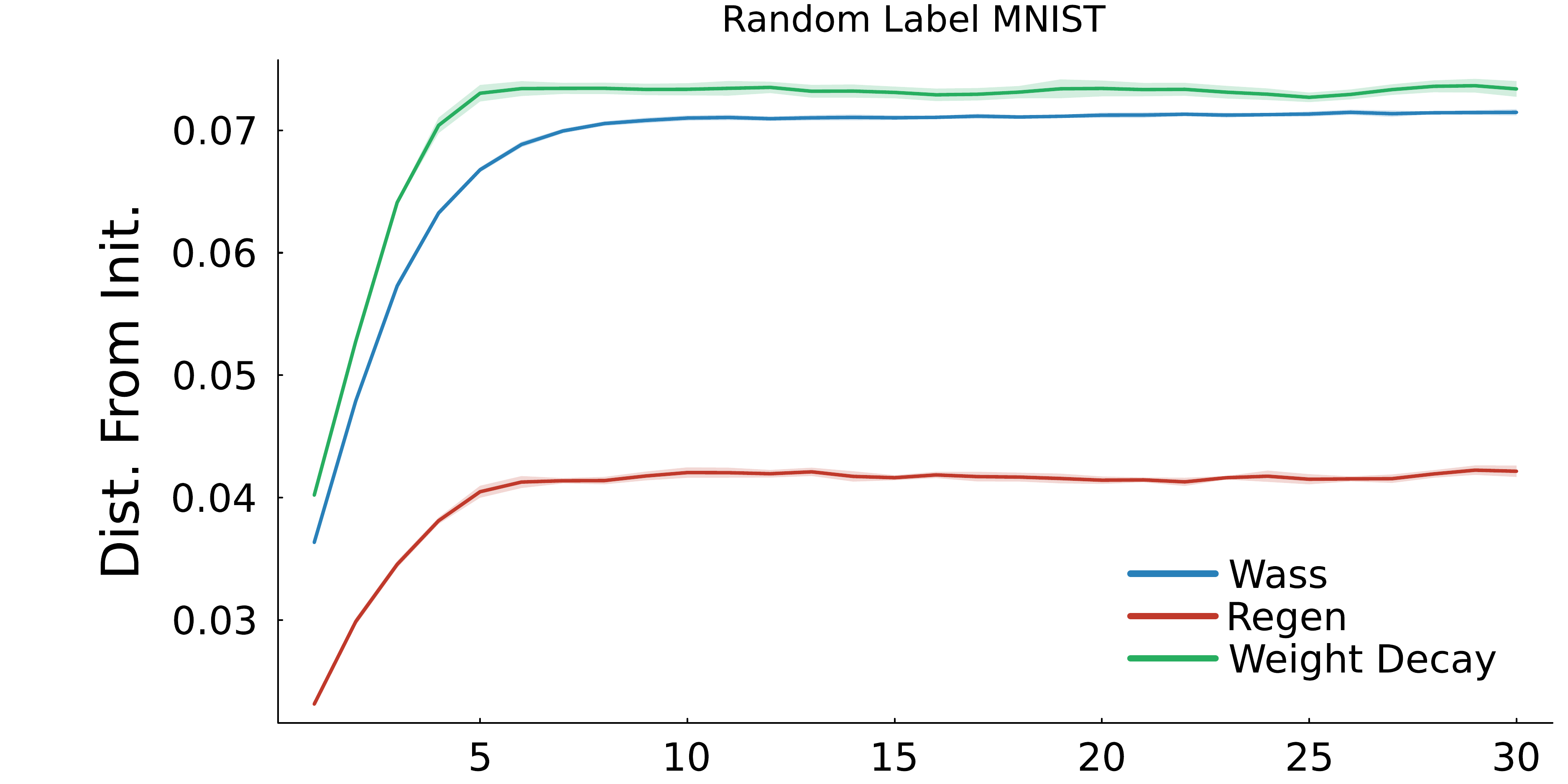}\\
  \includegraphics[width=0.49\linewidth]{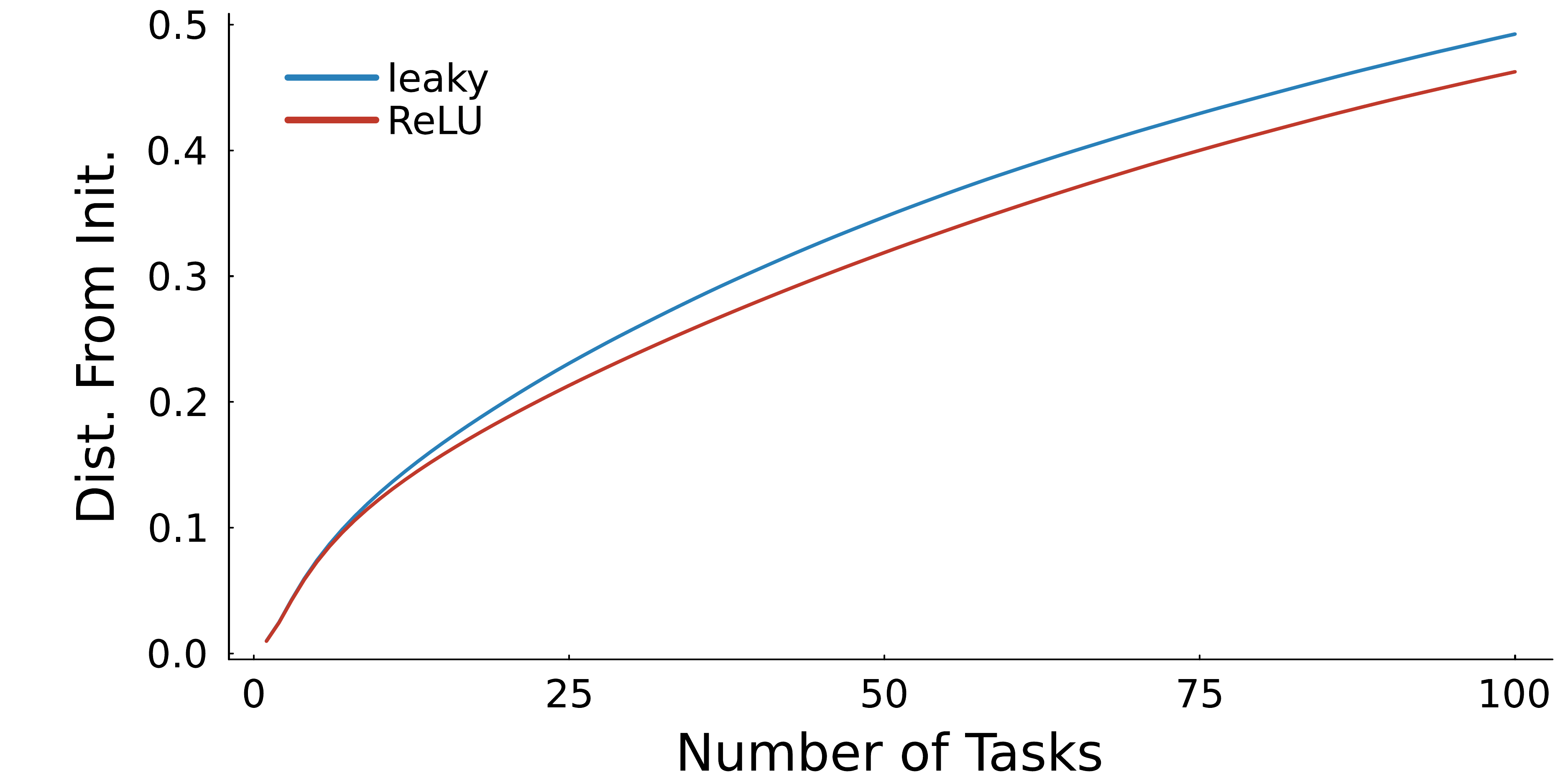}
  \includegraphics[width=0.49\linewidth]{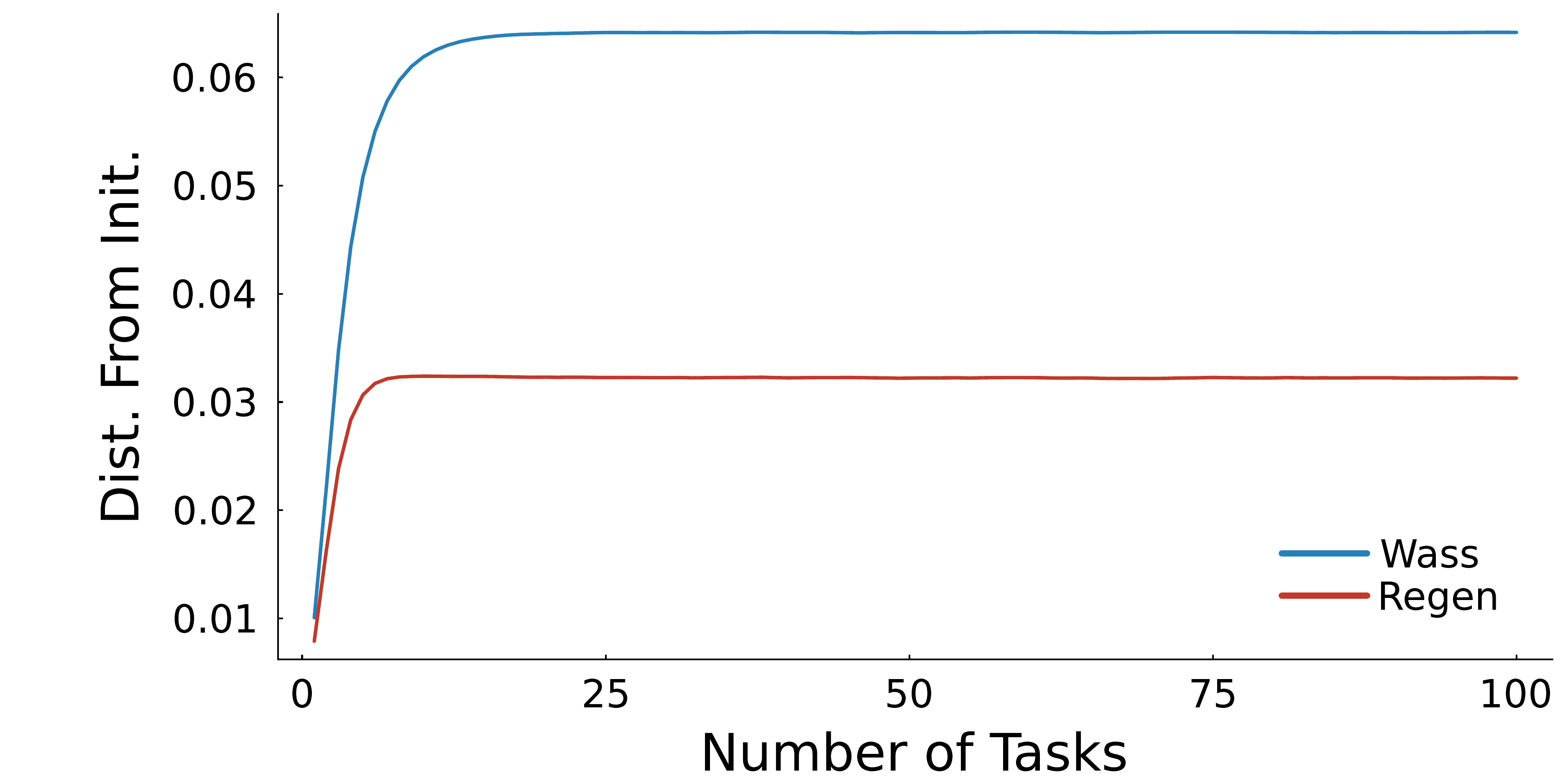}\\
  \includegraphics[width=0.49\linewidth]{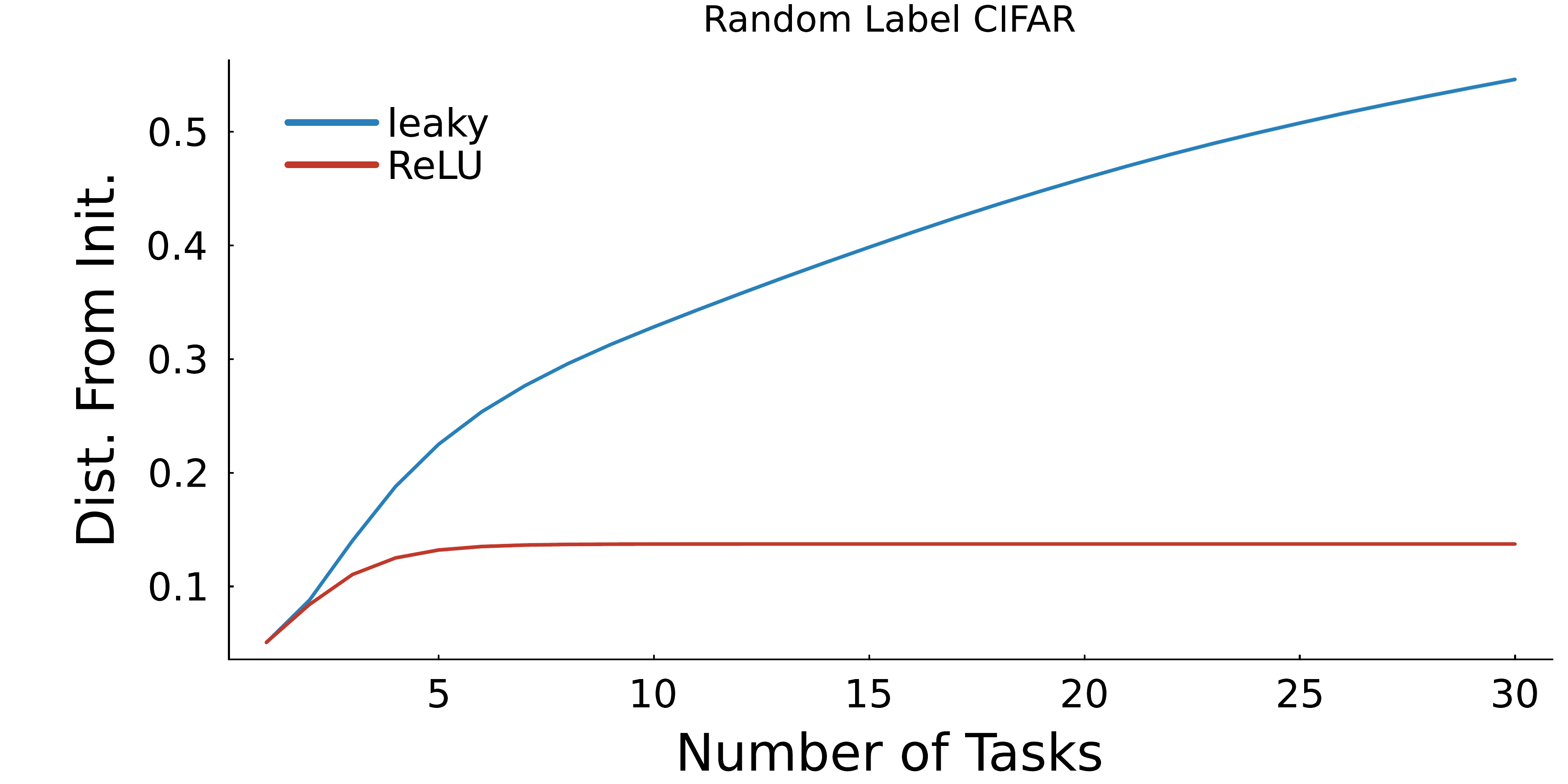}
  \includegraphics[width=0.49\linewidth]{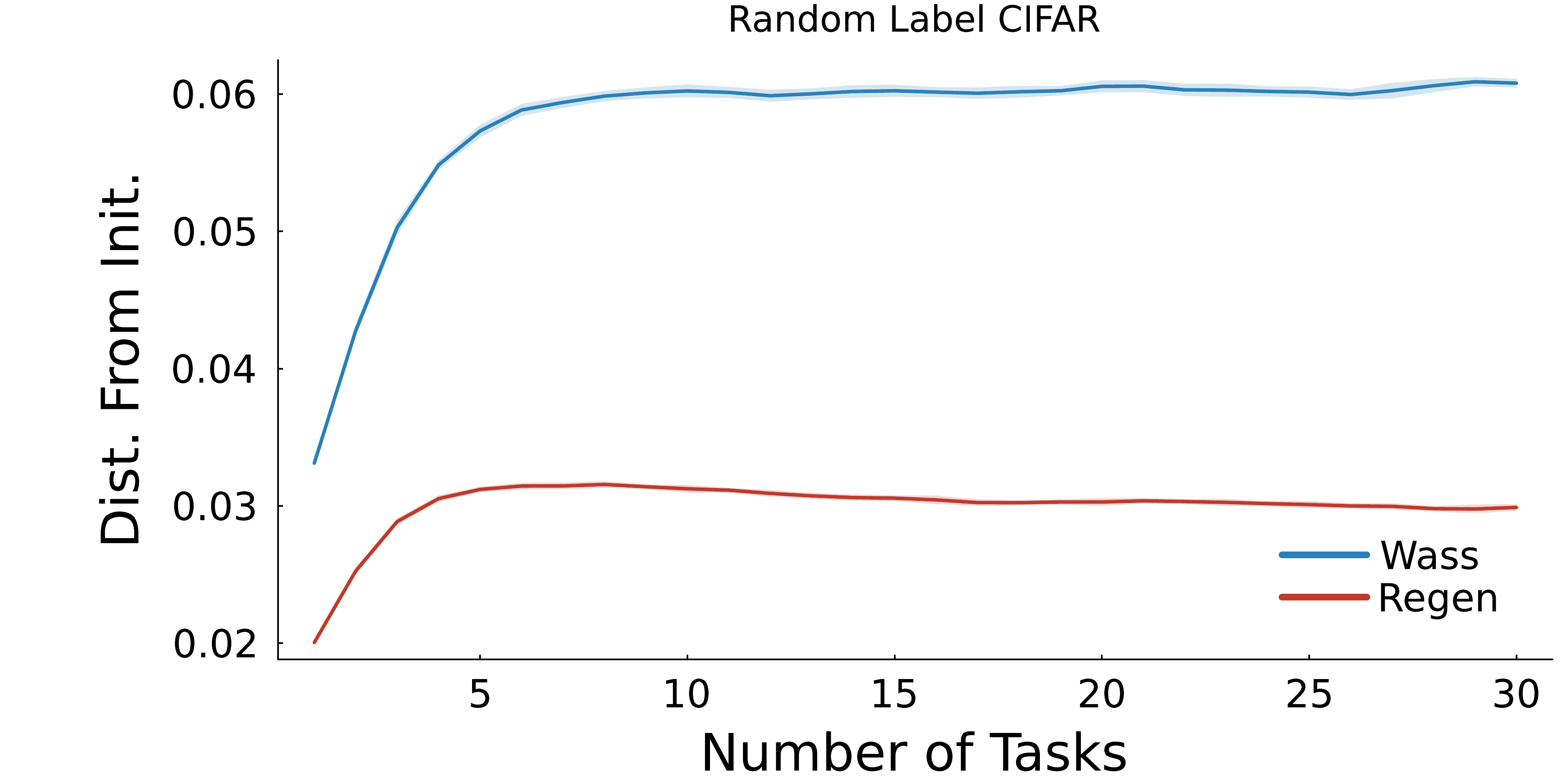}\\
  \includegraphics[width=0.49\linewidth]{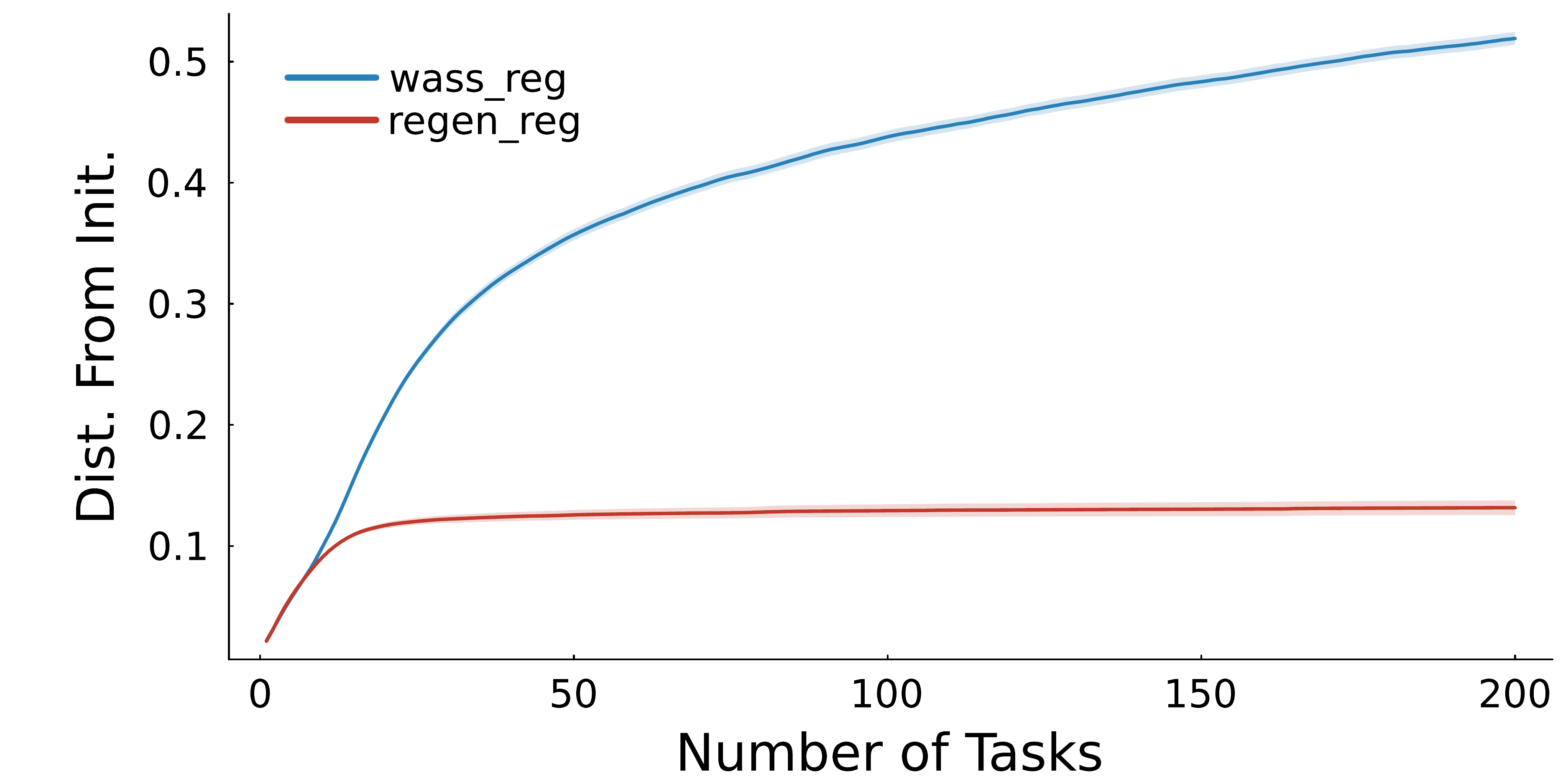}
  \includegraphics[width=0.49\linewidth]{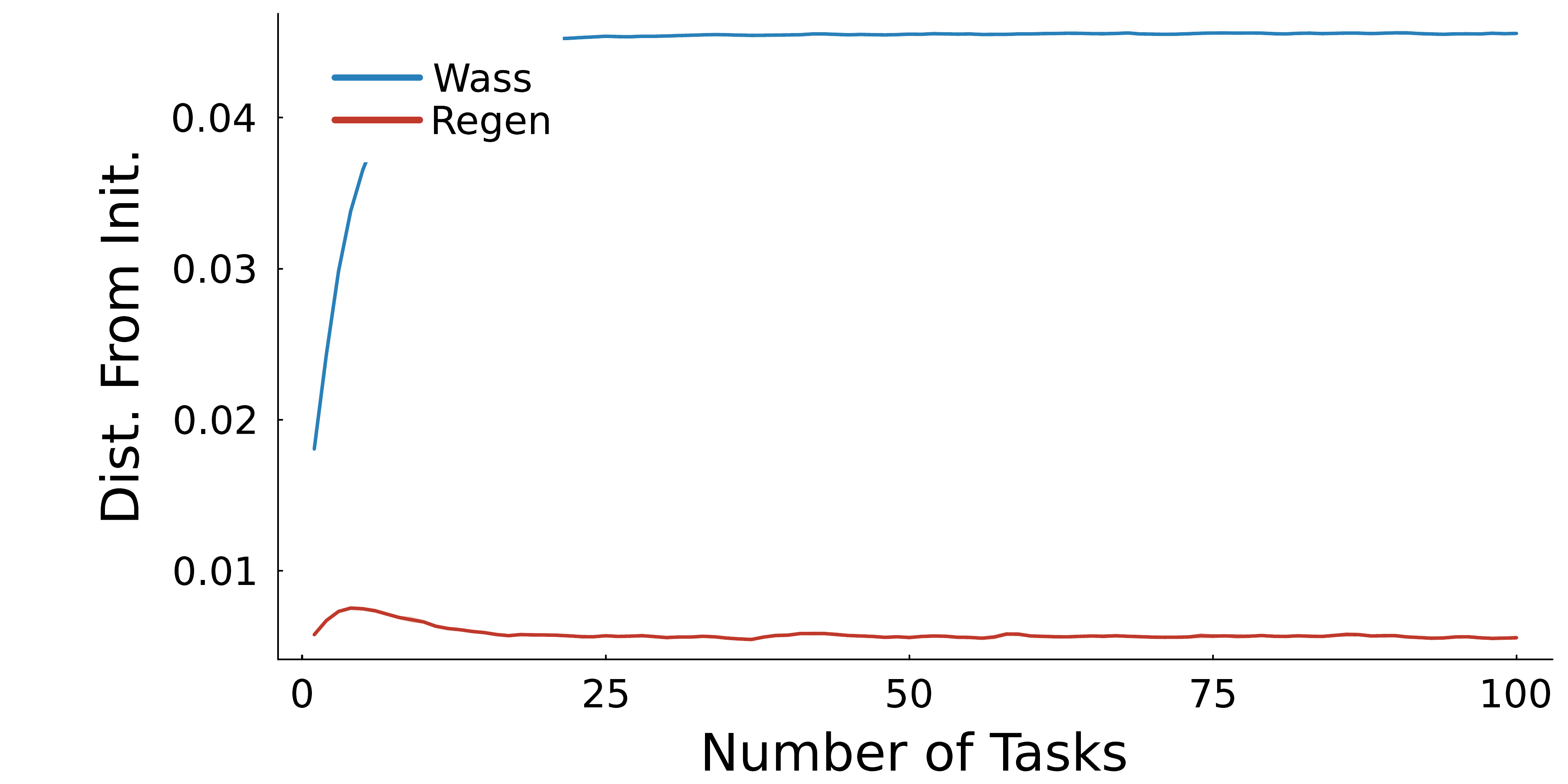}\\
  \caption{
    Regularization generally prevents neural network parameters from deviating from initialization compared to the unregularized setting. But, the Wasserstein regularizer minimizes a distributional distance that allows it to travel further from its initialization. Weight decay, although further from the initialization, is closer to the initialization distribution which can also lead to loss of plasticity \citep{zilly21}.
    }
  \label{fig:mnist_dist}
\end{figure}
\clearpage

\subsection{Regularizer Hyperparameter Sensitivity}
\label{appendix:reg_hyperparam}

The plots below show the batch error at the end of a task for different regularization strengths. Compared to weight decay and regenerative regularization, the Wasserstein regularizer is able to reach and maintain a lower error across most problems and activation functions.

\begin{figure}[h!]
  \centering
  \includegraphics[width=0.99\linewidth]{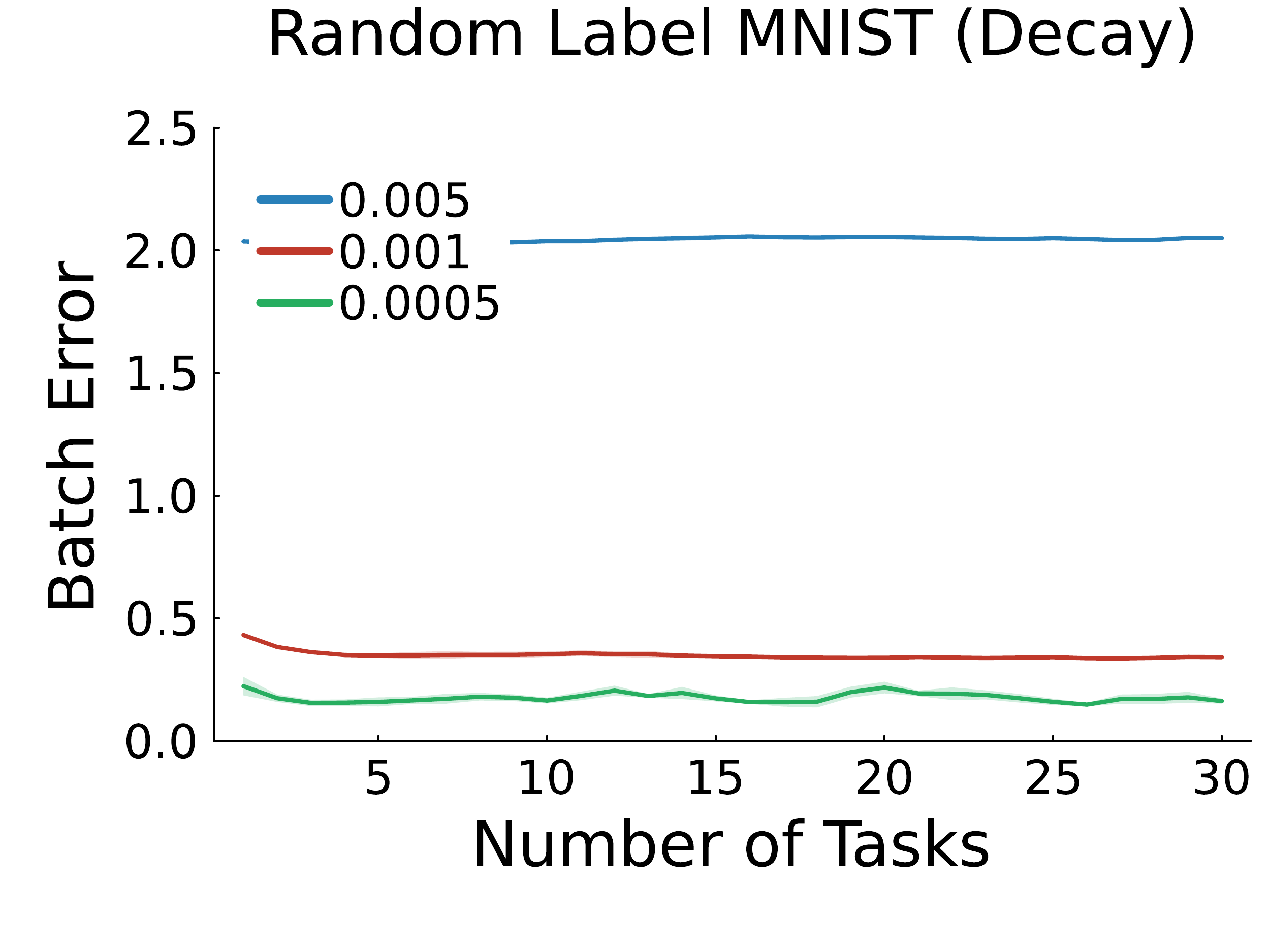}\\
  \includegraphics[width=0.49\linewidth]{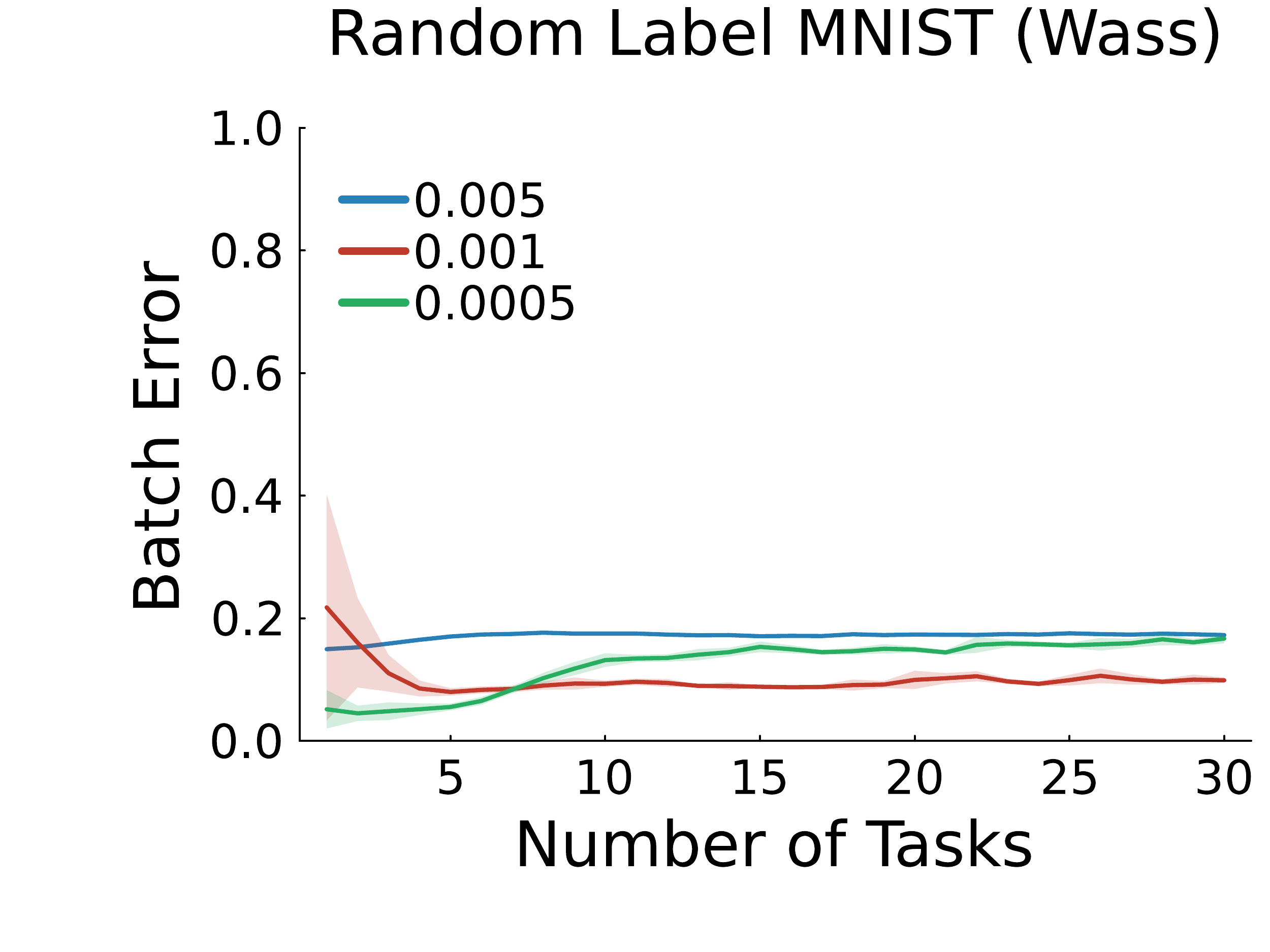}
  \includegraphics[width=0.49\linewidth]{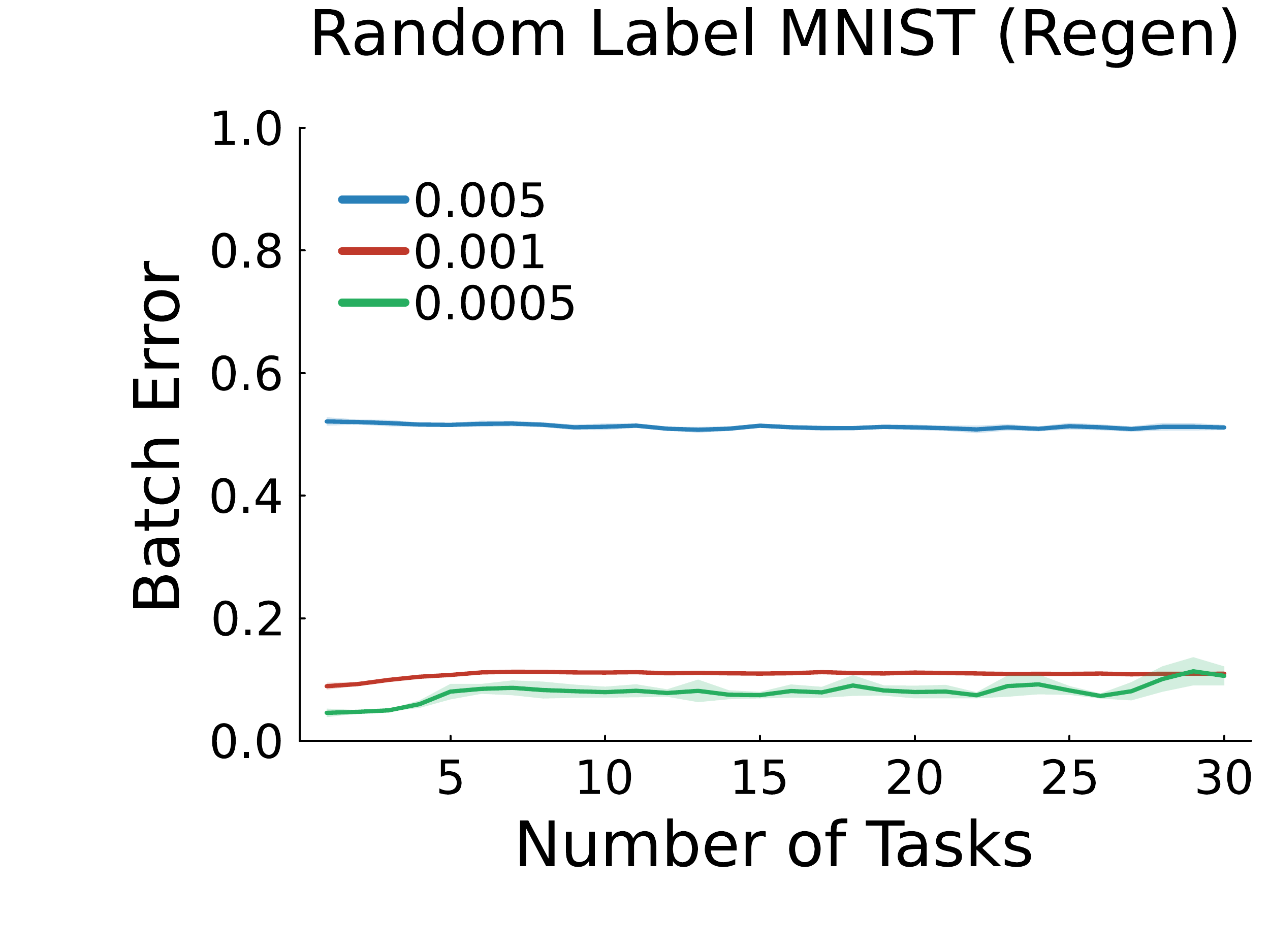}
  \caption{
    Learning curves on Random Label MNIST with different regularizers and different regularization strengths. The Wasserstein regularizer is less sensitive to the regularization strength
    }
  \label{fig:mnistshuffle_abl}
\end{figure}

\begin{figure}[h!]
  \centering
  \includegraphics[width=0.49\linewidth]{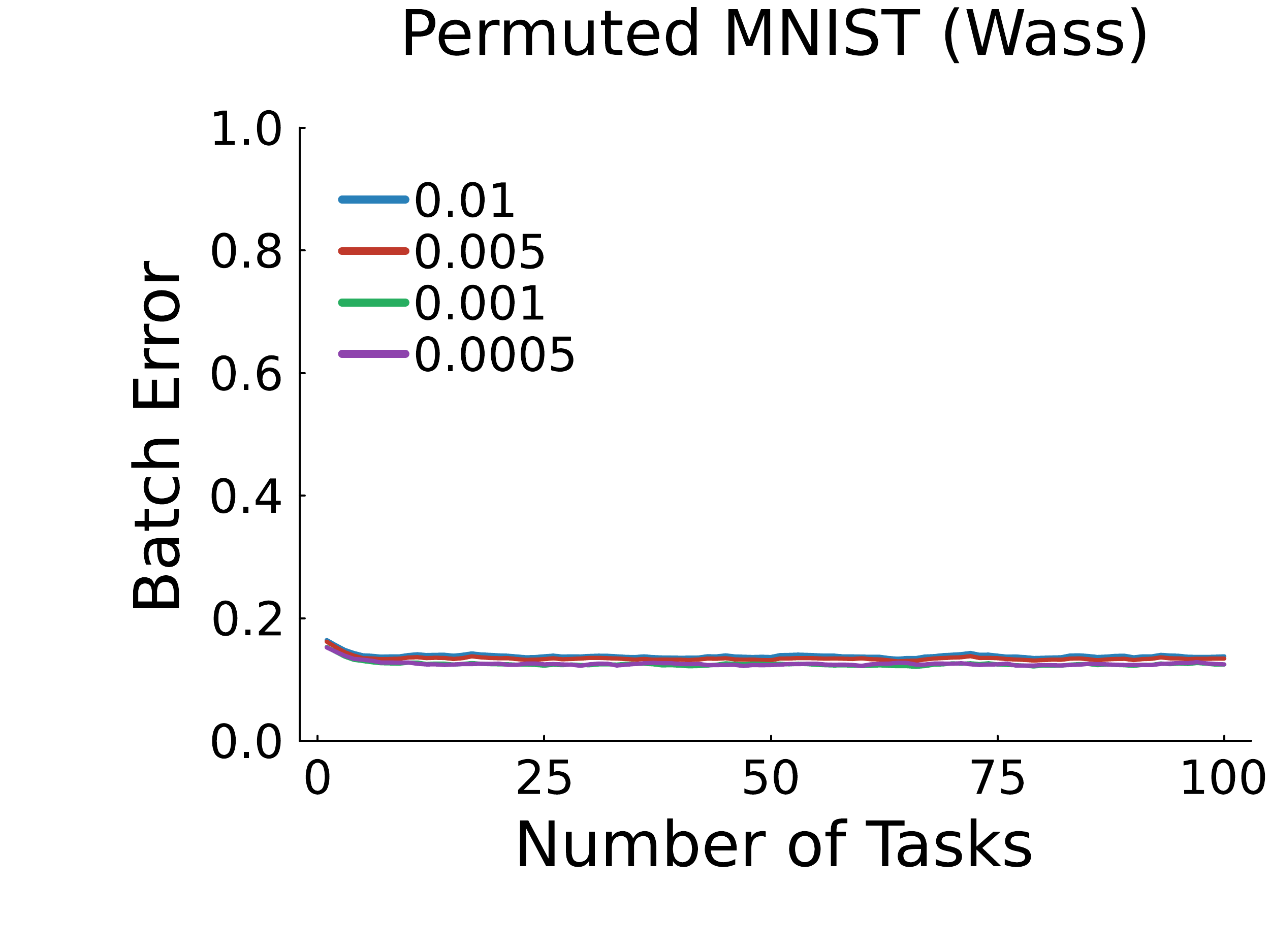}
  \includegraphics[width=0.49\linewidth]{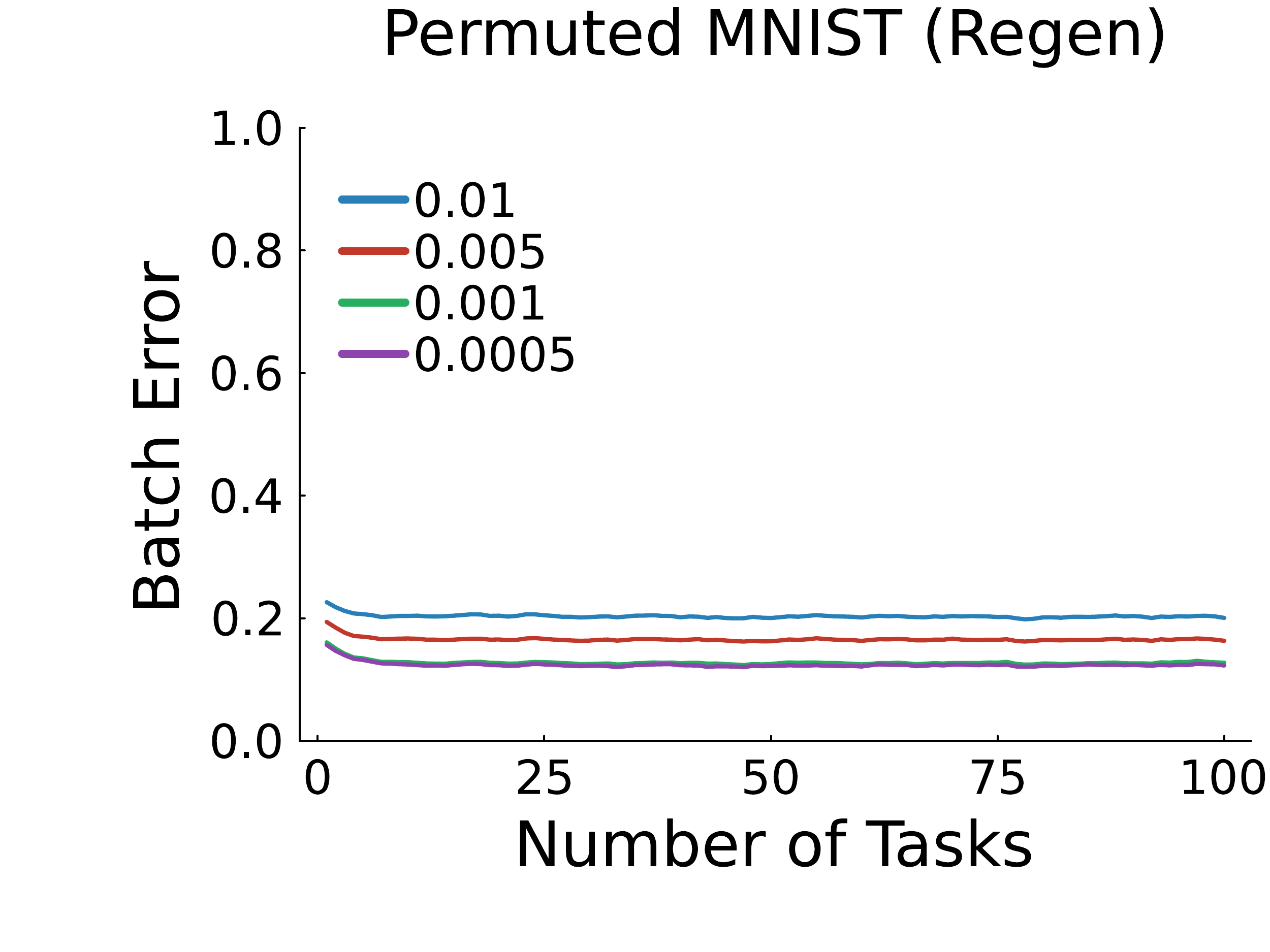}
  \caption{
    Learning curves on Permuted MNIST with different regularizers and different regularization strengths. The Wasserstein regularizer is less sensitive to the regularization strength
    }
  \label{fig:mnistperm_abl}
\end{figure}

\begin{figure}[h!]
  \centering
  \includegraphics[width=0.49\linewidth]{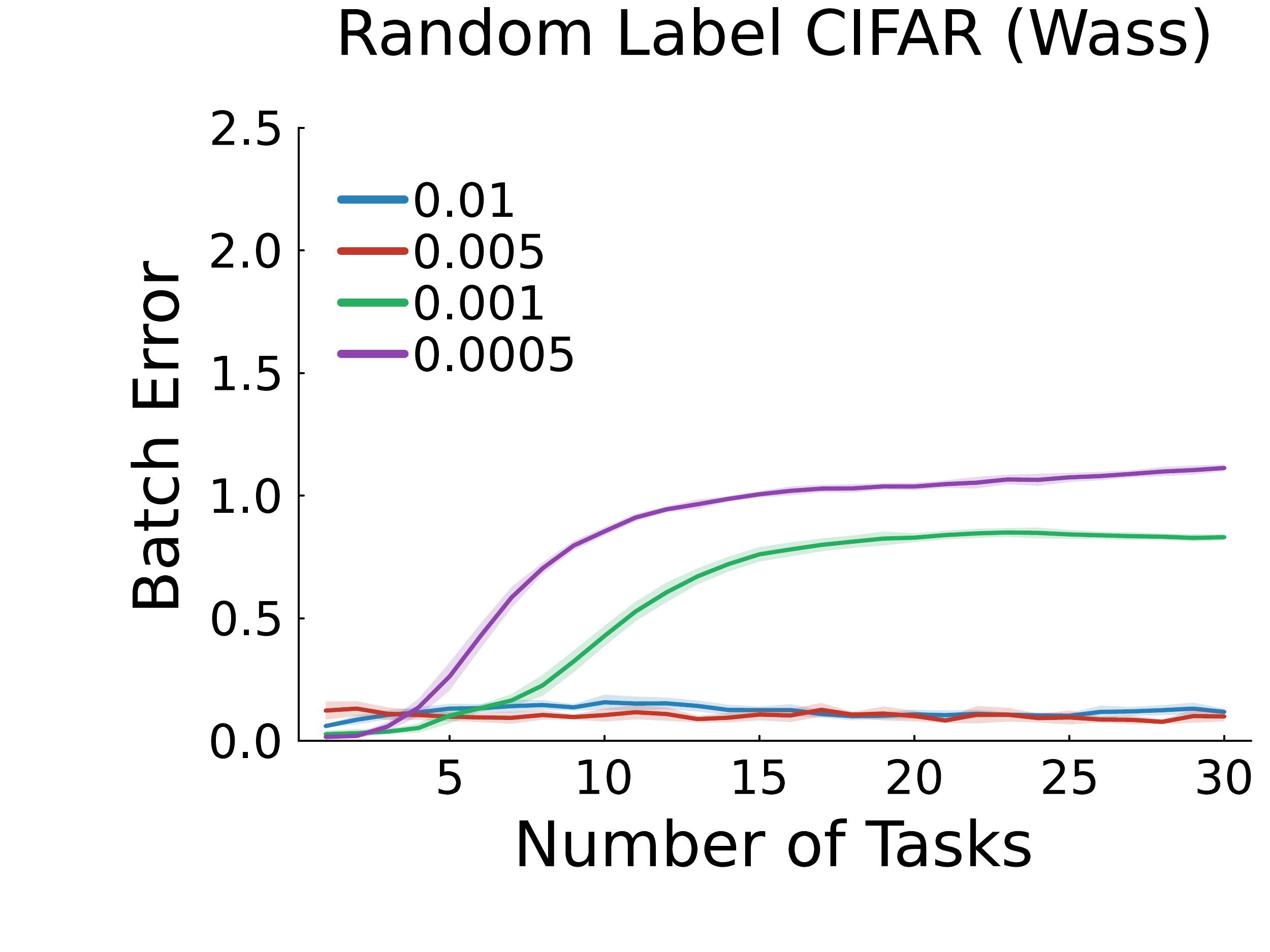}
  \includegraphics[width=0.49\linewidth]{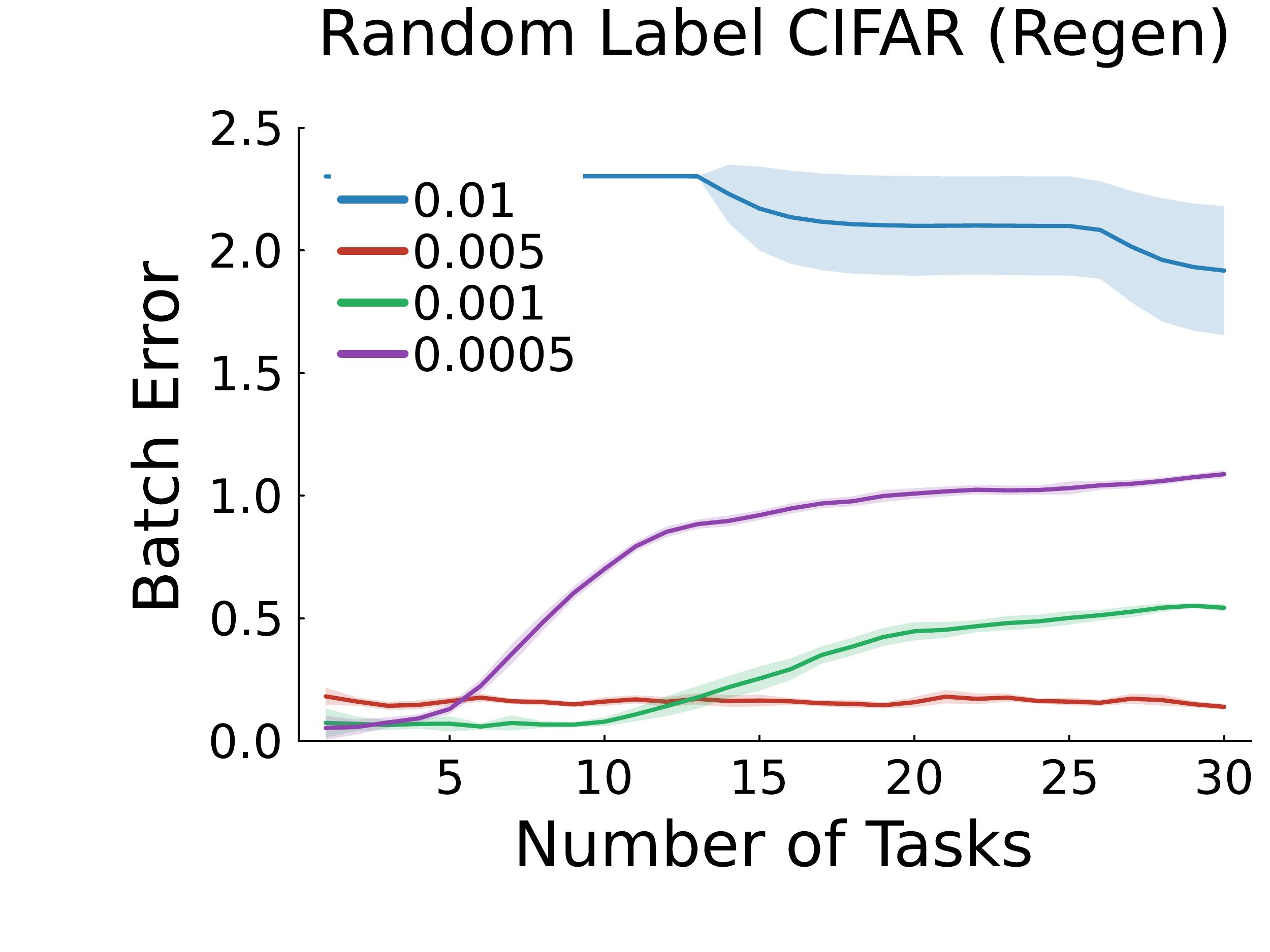}
  \caption{
    Learning curves on Random Label CIFAR with different regularizers and different regularization strengths. The Wasserstein regularizer is less sensitive to the regularization strength
    }
  \label{fig:cifar_abl}
\end{figure}

\begin{figure}[h!]
  \centering
  \includegraphics[width=0.49\linewidth]{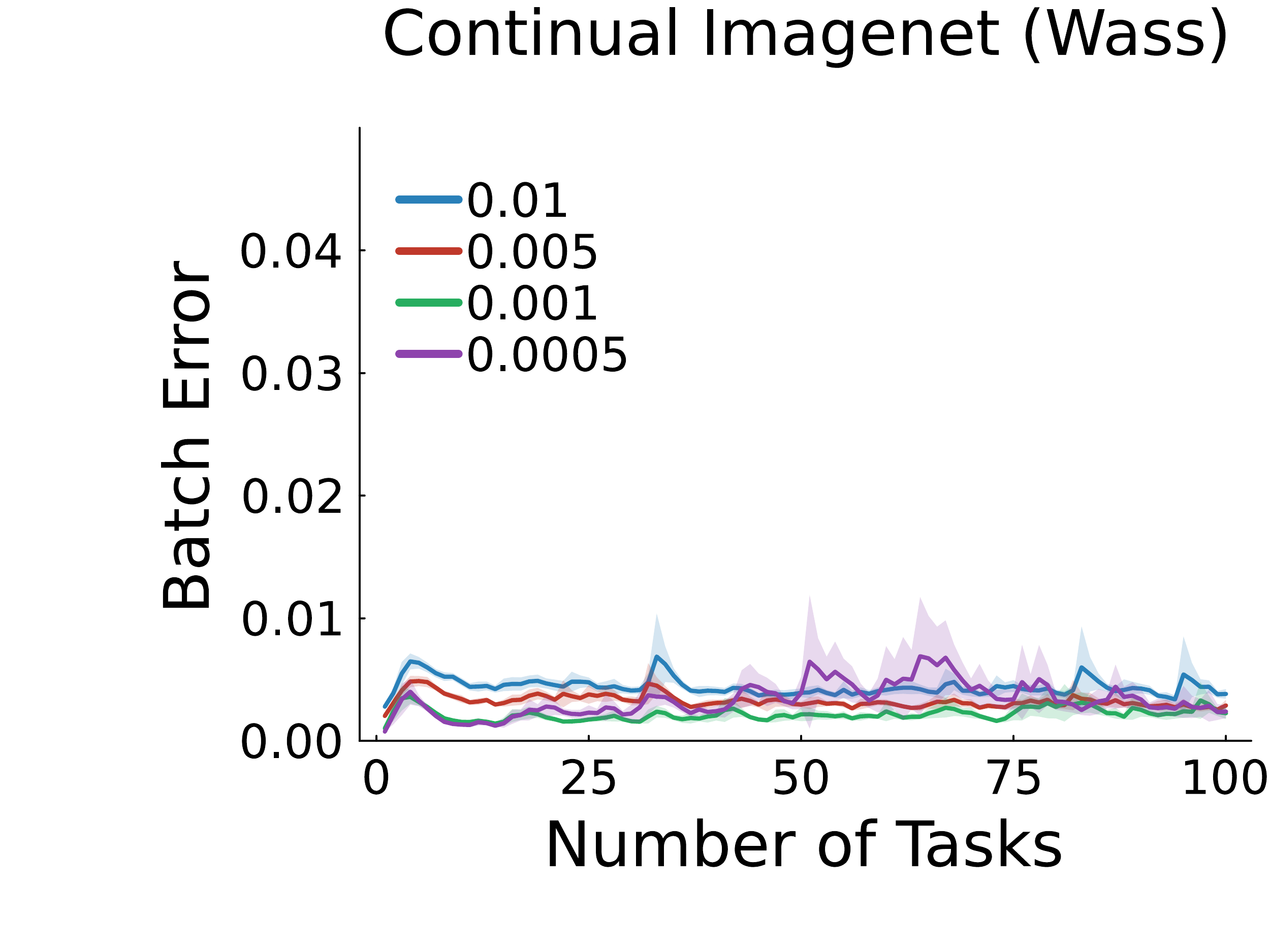}
  \includegraphics[width=0.49\linewidth]{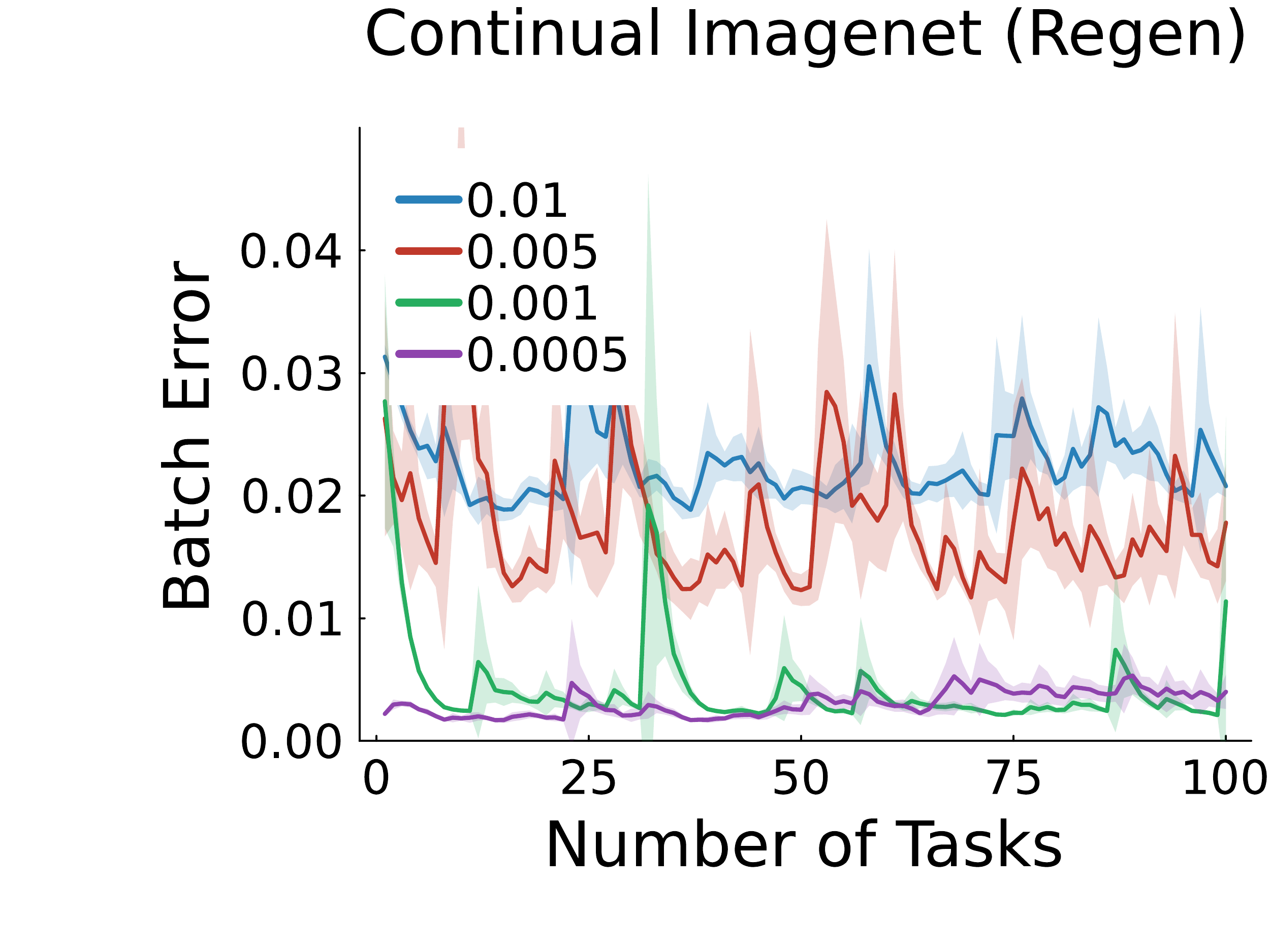}
  \caption{
    Learning curves on Continual Imagenet with different regularizers and different regularization strengths. The Wasserstein regularizer is less sensitive to the regularization strength
    }
  \label{fig:imagenet_abl}
\end{figure}

\clearpage
\subsection{Inter-task Online Learning Curves Without Regularization}
\label{appendix:intertask_noreg}

\begin{figure}[h!]
  \centering
  \includegraphics[width=0.99\linewidth]{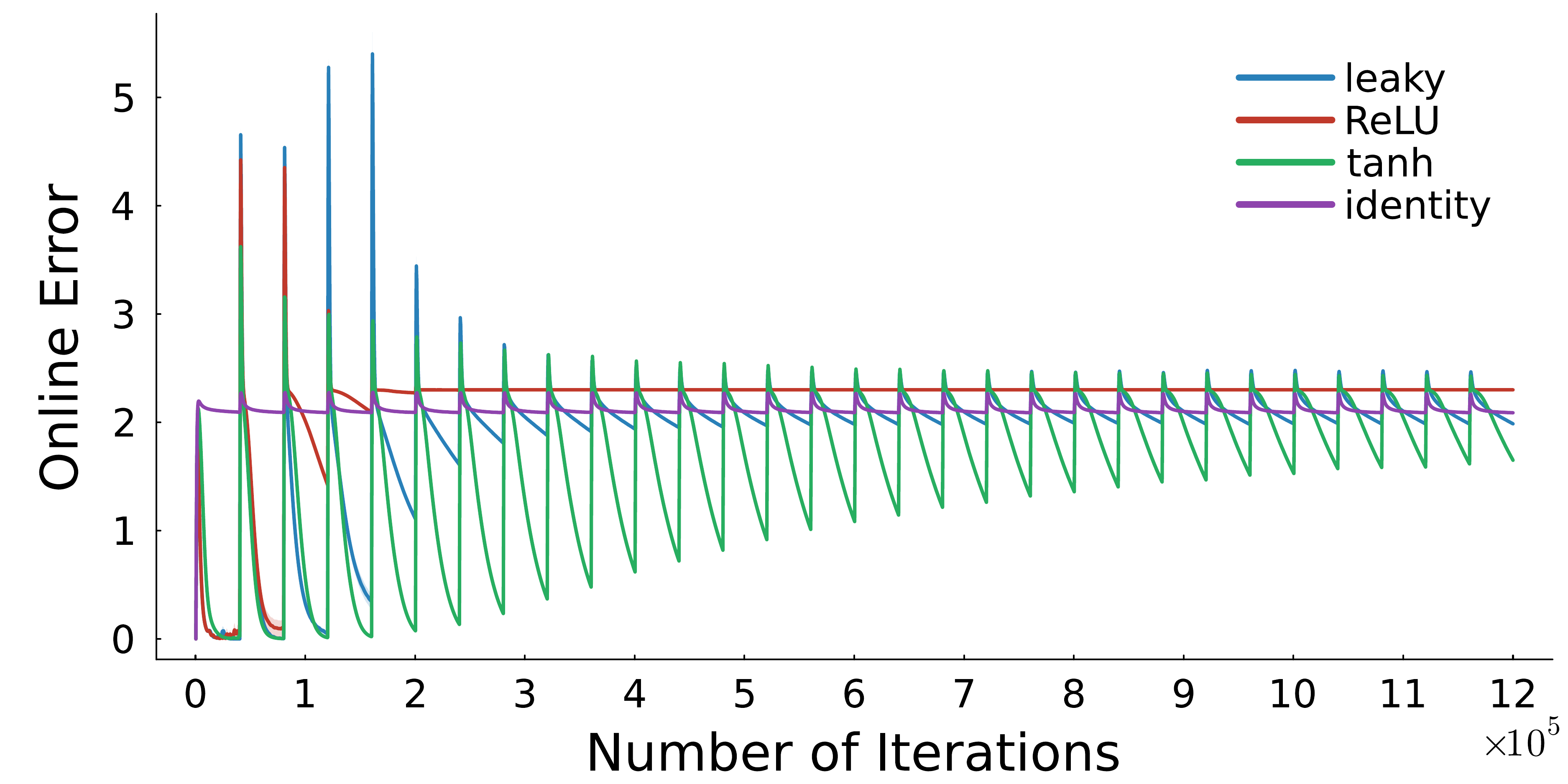}
  \includegraphics[width=0.99\linewidth]{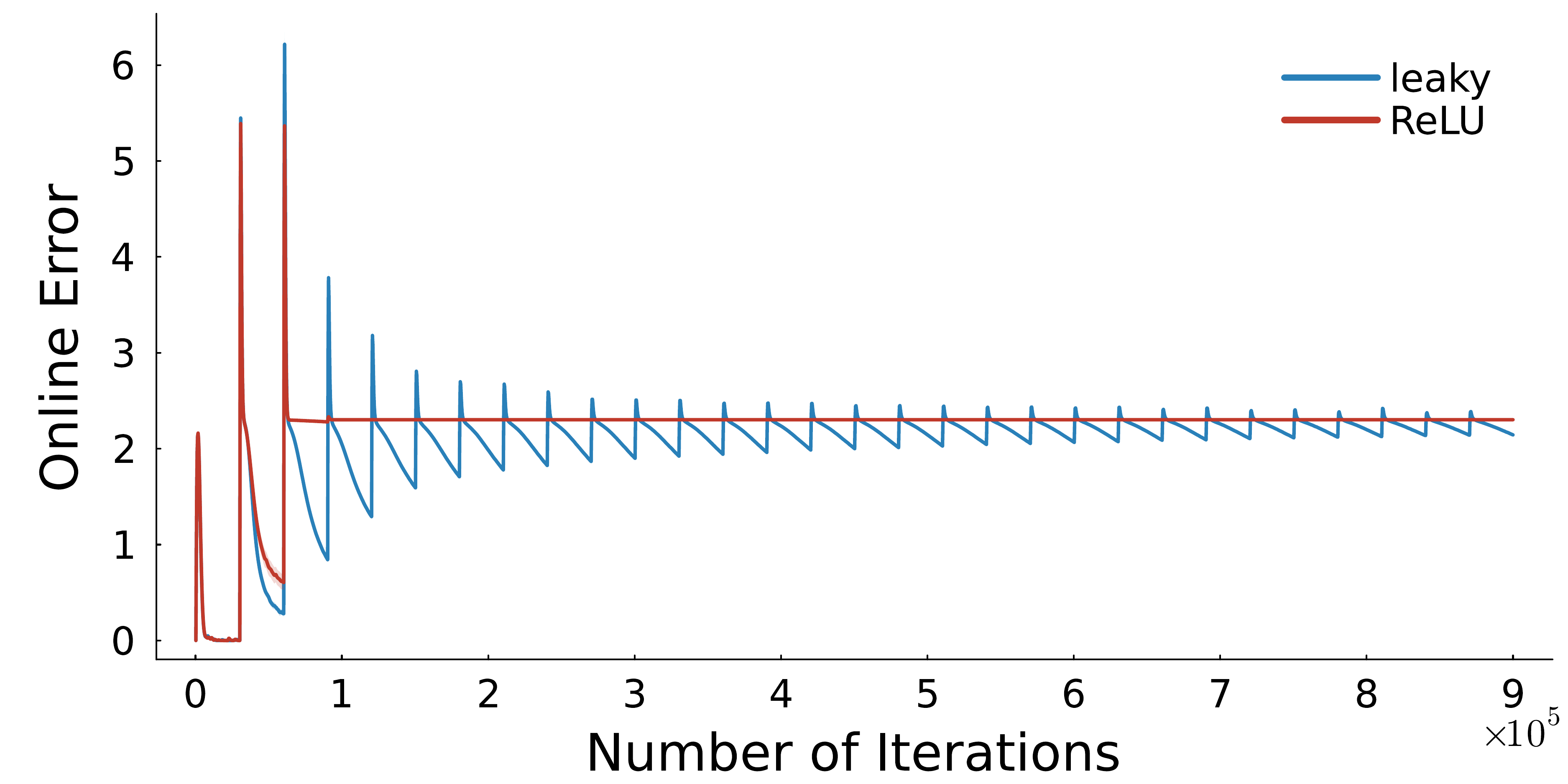}\\
  \caption{
    Intertask online learning curves on Random Label MNIST and Random Label CIFAR, without regularization.
    }
  \label{fig:intertask}
\end{figure}

\newpage
\subsection{Inter-task Online Learning Curves With Regularization}
\label{appendix:intertask}

\begin{figure}[h!]
  \centering
  \includegraphics[width=0.99\linewidth]{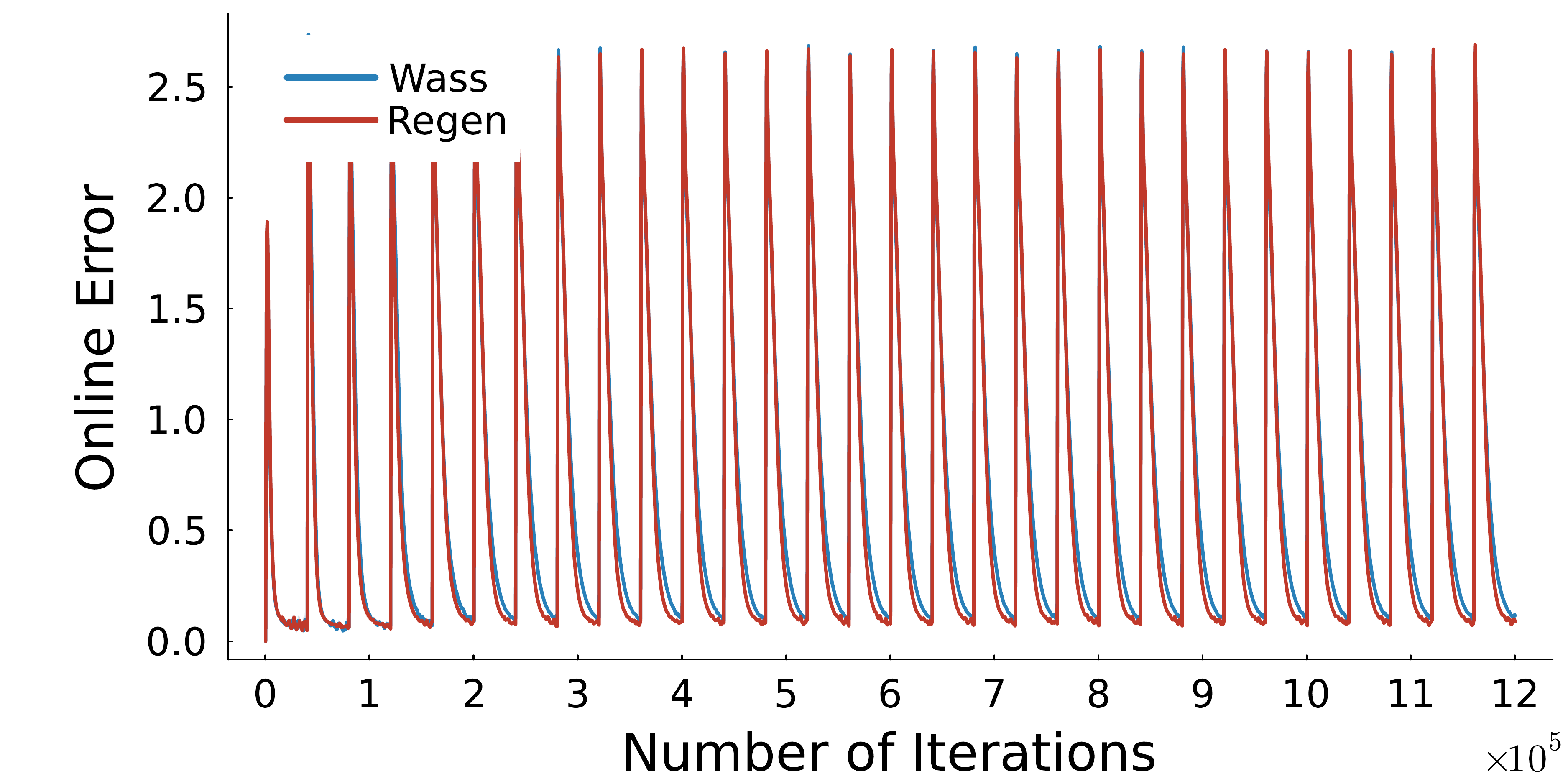}
  \includegraphics[width=0.99\linewidth]{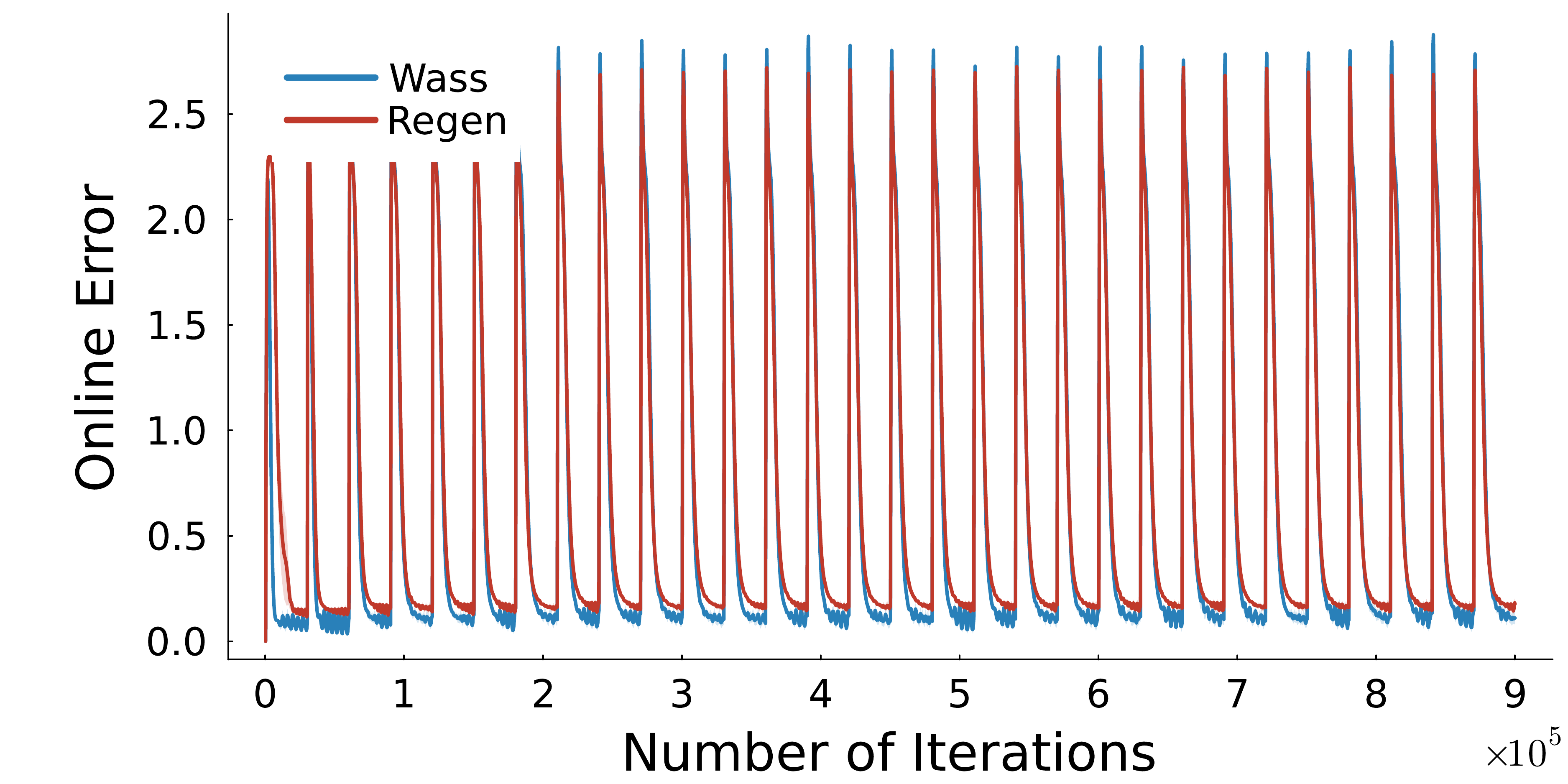}\\
  \caption{
    Intertask online learning curves on Random Label MNIST and Random Label CIFAR, with different regularizers.
    }
  \label{fig:intertask}
\end{figure}

\clearpage
\subsection{Update Budget Effect on Plasticity }
\label{appendix:epochs}

\begin{figure}[h!]
  \centering
  \includegraphics[width=0.99\linewidth]{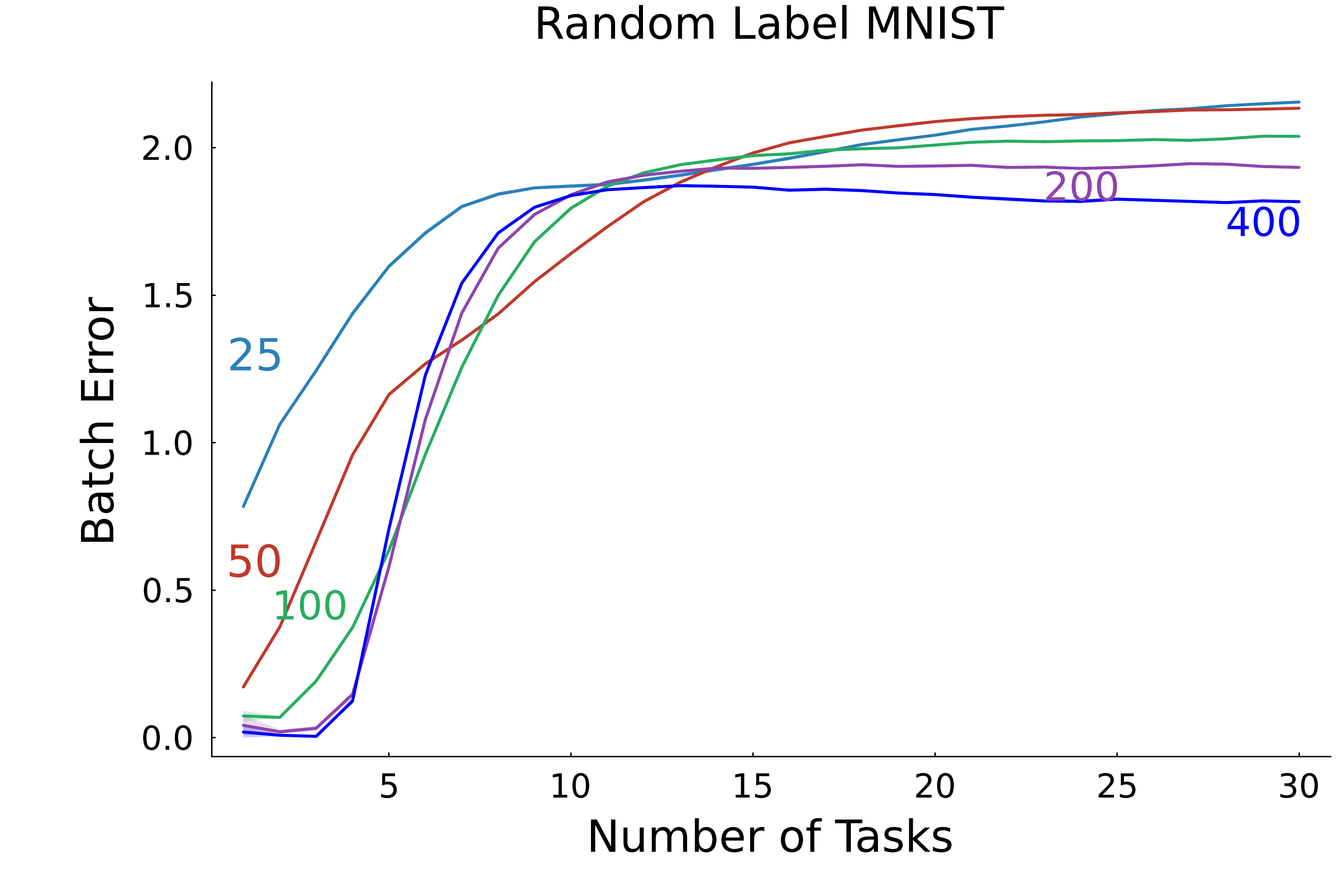}
    \caption{Ablating the number of epochs per task on Random Label MNIST. Loss of plasticity occurs when the number of epochs is small (25), despite not overfitting to the first few task. Loss of plasticity eventually also occurs when the number of epochs is large (400), but reduces the final error plateau.}
  \label{fig:mnist_updates}
\end{figure}

By varying the number of epochs in a task, the neural network is able to learn more on a task, perhaps allowing the neural network to escape from loss of plasticity.
Unfortunately, the results in Figure~\ref{fig:mnist_updates} shows that increasing the number of epochs only marginally delays the onset of loss of plasticity.
Plasticity loss still occurs, but reduction in curvature is a consistent predictor of this phenomenon.

\end{document}